\documentclass[10pt,twocolumn,letterpaper]{article}

\usepackage[pagenumbers]{cvpr}

\renewcommand{\paragraph}[1]{\vspace{.5em}\noindent\textbf{#1}}

  \definecolor{cvprblue}{rgb}{0.21,0.49,0.74}
  \usepackage[pagebackref,breaklinks,colorlinks,allcolors=cvprblue]{hyperref}

\usepackage{amsmath,amsfonts,bm}

\newcommand{\captiona}{{(a)}\xspace}
\newcommand{\captionb}{{(b)}\xspace}
\newcommand{\captionc}{{(c)}\xspace}
\newcommand{\captiond}{{(d)}\xspace}

\newcommand{\refcaptiona}{{\textcolor{cvprblue}{a}}\xspace}
\newcommand{\refcaptionb}{{\textcolor{cvprblue}{b}}\xspace}
\newcommand{\refcaptionc}{{\textcolor{cvprblue}{c}}\xspace}
\newcommand{\refcaptiond}{{\textcolor{cvprblue}{d}}\xspace}

\def\eqref#1{equation~\ref{#1}}

\def\1{\bm{1}}

\def\rvb{{\mathbf{b}}}

\def\rvf{{\mathbf{f}}}

\def\rvu{{\mathbf{i}}}

\def\rvk{{\mathbf{k}}}

\def\rvn{{\mathbf{n}}}

\def\rvq{{\mathbf{q}}}

\def\rvt{{\mathbf{t}}}
\def\rvu{{\mathbf{u}}}

\def\rvx{{\mathbf{x}}}

\def\rmR{{\mathbf{R}}}

\def\vzero{{\bm{0}}}

\def\vb{{\bm{b}}}

\def\evd{{d}}

\def\mA{{\bm{A}}}

\def\mD{{\bm{D}}}

\def\mI{{\bm{I}}}

\def\mK{{\bm{K}}}

\def\mS{{\bm{S}}}

\def\mW{{\bm{W}}}

\DeclareMathAlphabet{\mathsfit}{\encodingdefault}{\sfdefault}{m}{sl}
\SetMathAlphabet{\mathsfit}{bold}{\encodingdefault}{\sfdefault}{bx}{n}

\def\sM{{\mathbb{M}}}

\def\sQ{{\mathbb{Q}}}
\def\sR{{\mathbb{R}}}

\DeclareMathOperator*{\argmin}{arg\,min}

\usepackage{lipsum}
\usepackage[mode=text, ]{siunitx}
\usepackage[dvipsnames]{xcolor}  %
\usepackage[cas-rgb*]{cncolours}

\usepackage{multirow}
\usepackage{makecell}  %
\usepackage{colortbl}

\usepackage{titlecaps}  %
\usepackage[capitalize]{cleveref}

\usepackage{pifont}  %
\usepackage{dsfont}  %

\usepackage{tikz}

\hypersetup{
  citecolor=cvprblue,
}
\def\wo{\emph{w/o}\xspace}  %
\def\with{\emph{w/}\xspace}  %
\def\Wo{\emph{W/o}\xspace}  %
\def\see{\emph{see}\xspace}  %

\newcommand{\qelegend}[1]{\raisebox{-.7ex}{\includegraphics[height=1.9\fontcharht\font`B]{assets/qualitative-examples/#1}}\xspace}
\newcommand{\recalllegend}{\raisebox{-.5ex}{\includegraphics[height=1.6\fontcharht\font`B]{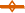}}\xspace}

\newcommand{\queryicon}{\raisebox{-.3ex}{\includegraphics[height=1.2\fontcharht\font`B]{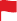}}}
\newcommand{\mapicon}{\raisebox{-.3ex}{\includegraphics[height=1.2\fontcharht\font`B]{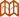}}}
\newcommand{\pointcorrespondenceicon}{\raisebox{-.3ex}{\includegraphics[height=1.2\fontcharht\font`B]{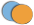}}}
\newcommand{\planecorrespondenceicon}{\raisebox{-.3ex}{\includegraphics[height=1.4\fontcharht\font`B]{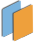}}}

\colorlet{tableRankFirst}{朱红!35}
\colorlet{tableRankSecond}{琥珀黄!35!white}
\colorlet{tableRankThird}{芽绿!40}

\newcommand{\TRankFirst}[1]{\cellcolor{tableRankFirst}\textbf{#1}}
\newcommand{\TRankSecond}[1]{\cellcolor{tableRankSecond}\emph{#1}}
\newcommand{\TRankThird}[1]{\cellcolor{tableRankThird}#1}

\definecolor{tikzColorMethodGT}{HTML}{33a02c}
\definecolor{tikzColorMethodCoarse}{HTML}{ad9519}
\definecolor{tikzColorMethodOurs}{HTML}{1f78b4}
\definecolor{tikzColorMethodOursNoRefine}{HTML}{1f78b4}
\definecolor{tikzColorMethodGeoTr}{HTML}{cab2d6}
\definecolor{tikzColorMethodMatchAny}{HTML}{a6cee3}
\definecolor{tikzColorMethodNOPESAC}{HTML}{fb9a99}
\definecolor{tikzColorMethodMast3r}{HTML}{ff7f00}
\definecolor{warpOnto}{HTML}{1247dc}
\definecolor{matchinlier}{HTML}{57e389}
\definecolor{matchoutlier}{HTML}{ffa348}

\newcommand{\CASE}[1]{{\MakeUppercase{#1}}}
\newcommand{\enparen}[1]{\CASE{(}#1\CASE{)}}

\crefname{section}{Sec.
}{Secs.}
\Crefname{section}{Section}{Sections}
\Crefname{table}{Table}{Tables}
\crefname{table}{Tab.}{Tabs.}
\Crefname{equation}{Equation}{Equations}
\crefname{equation}{Eq.}{Eqs.}
\Crefname{appendix}{Appendix}{Appendices}
\crefname{appendix}{Appendix}{Appendices}

\makeatletter
\@ifpackageloaded{CJKutf8}{
  \newcommand{\textcn}[1]{\begin{CJK*}{UTF8}{gbsn}#1\end{CJK*}}
}{
  \newcommand{\textcn}[1]{#1}
}

\@ifpackageloaded{todonotes}{
  \if@todonotes@disabled
    
    \newcommand{\change}[2]{#1}
    
    \newcommand{\TODO}[1]{}
    \newcommand{\outline}[1]{}
  \else

    \newcounter{uncertain}

    \newcommand{\change}[2]{\st{#1}\todo[linecolor=鹤顶红,backgroundcolor=鹤顶红!25,bordercolor=鹤顶红]{\textcn{#2}}}
    \newcommand{\outline}[1]{\todo[inline, linecolor=orange,backgroundcolor=orange!25,bordercolor=orange, author=Outline, size=\large]{\textcn{#1}}}

    \newcounter{yhqcomment}
  \fi
}{
  
  \newcommand{\change}[2]{#2}
  
  \newcommand{\outline}[1]{}
}
\makeatother

\def\repo{https://github.com/3dv-casia/PlanaReLoc}
\def\repourl{\href{\repo}{\repo}}

\def\hanqiaopage{https://timber-ye.github.io/home.html}

\def\methodname{PlanaReLoc\xspace}
\def\mapname{3D planar map\xspace}

\def\postrefine{post-refinement\xspace}
\def\matchingtwothreeD{2D\textendash 3D\xspace}

\def\AbbrCoarseInit{\emph{Coarse Init.}\xspace}
\def\AbbrImgToPC{I2P\xspace}
\def\AbbrGeoTransformer{GeoTr.\xspace}

\def\AbbrMatchAnything{MatchAny.\xspace}
\def\AbbrPoseRefine{refine.\xspace}

\def\AbbrTruncatedMap{Map trunc.\xspace}

\def\gtpose{the ground truth\xspace}
\def\geotransformer{GeoTransformer\xspace}
\def\freereg{FreeReg\xspace}
\def\ffreereg{Free-FreeReg\xspace}
\def\meshloc{MeshLoc\xspace}
\def\mastr{MASt3R\xspace}
\def\matchanything{MatchAnything\xspace}
\def\nopesac{NOPE-SAC\xspace}
\def\planetr{PlaneTR\xspace}
\def\planerectr{PlaneRecTR\xspace}
\def\zeroplane{ZeroPlane\xspace}
\def\planathreer{Plana3R\xspace}
\def\seqRANSAC{RANSAC\xspace}

\def\mogevtwo{MoGe-2\xspace}

\def\scannet{ScanNet\xspace}
\def\twelvescenes{12Scenes\xspace}
\def\sevenscenes{7Scenes\xspace}

\def\Rot{\rmR}
\def\Pos{\rvt}
\def\RotError{\Delta \Rot}
\def\PosError{\Delta \Pos}

\def\AccFine{\enparen{\qty{0.2}{\meter},\,\qty{10}{\degree}}}
\def\AccMed{\enparen{\qty{0.5}{\meter},\,\qty{15}{\degree}}}
\def\AccCoarse{\enparen{\qty{1.0}{\meter},\,\qty{30}{\degree}}}

\def\AbbrMatchPrecision{\text{Prec.}}
\def\AbbrMatchRecall{\text{Rec.}}
\def\AbbrMatchFScore{\text{F$_\text{1}$}}

\def\Tna{-}
\def\Tcheckmark{\ding{51}}
\def\TNoRankColor{gray!15}
\def\Thookrightarrow{{\footnotesize $\hookrightarrow$}\,}

\def\TCoarseInit{Coarse\\init.}
\def\TTruncatedMap{Map\\ trunc.}
\def\TTextured{Map ap-\\pearance}
\def\TRotErr{$\RotError$ \enparen{\unit{\degree}} $\downarrow$}
\def\TPosErr{$\PosError$ \enparen{\unit{m}} $\downarrow$}
\def\TAccThres{Pose Recall \enparen{\unit{\percent}} $\uparrow$}
\def\TMethods{}
\def\TRotErrMean{\small Mean}
\def\TRotErrMed{\small Med.}
\def\TPosErrMean{\TRotErrMean}
\def\TPosErrMed{\TRotErrMed}
\def\TAccFine{\footnotesize \AccFine}
\def\TAccMed{\footnotesize \AccMed}
\def\TAccCoarse{\footnotesize \AccCoarse}
\def\TTime{Time\\\enparen{\unit{\second\per\text{iter}}}}
\def\Tsim{~$\sim$~}

\def\Timgtopc{\AbbrImgToPC}
\def\Tcoarseinit{\AbbrCoarseInit}
\def\TgeotransformerF{\geotransformer}
\def\TgeotransformerT{\geotransformer-T}
\def\Tfreereg{\freereg}
\def\Tffreereg{\ffreereg}
\def\Tmeshloc{\meshloc}

\def\Tsplg{SP + LG}
\def\Tloftr{LoFTR}
\def\Tmastr{\mastr}
\def\Tmatchanything{\matchanything}
\def\Tnopesac{\nopesac}
\def\Tplanathreer{\planathreer}

\def\TAbbrgeotransformerT{\AbbrGeoTransformer-T}

\def\TAbbrmatchanything{\AbbrMatchAnything}

\def\Tours{Ours}
\def\Toursnorefine{\Wo \postrefine}
\def\TAbbroursnorefine{\Wo \AbbrPoseRefine}
\def\Toursfull{\emph{Full} proposed}
\def\TAbbroursfull{\emph{Full}}

\def\TFeatureType{Feature\\type}
\def\TFPoint{\ttfamily Point}
\def\TFPlane{\ttfamily Plane}
\def\TcoloredFeature#1{{#1\textcolor{橘橙}{$^\bullet$}}}
\def\TMatchScanNet{\scannet}
\def\TMatchTwelveScenes{\twelvescenes}
\def\TMatchPrecision{\AbbrMatchPrecision$\uparrow$}
\def\TMatchRecall{\AbbrMatchRecall$\uparrow$}
\def\TMatchFScore{\AbbrMatchFScore\,$\uparrow$}
\def\TMatchAP{AP$\uparrow$}
\def\TMatchTP{\#\,TP}
\def\TMatchTotal{\#\,GT}

\def\TAblationMatchFScore{\TMatchFScore}
\def\TAblationPoseErr{Med.
  Err.
  $\downarrow$}
\def\TAblationRotErr{\footnotesize $\Delta \Rot$\enparen{\unit{\degree}}}
\def\TAblationPosErr{\footnotesize $\Delta \Pos$\enparen{\unit{m}}}
\def\TAblationPoseAccHigh{Pose \\ Recall\,$\uparrow$}

\def\TAblationOurs{\textbf{Ours}(\wo refine.)\xspace}
\def\TAblationNoScenewiseEncoding{\Thookrightarrow\wo scn. enc.\xspace}
\def\TAblationNoPiecewiseEncoding{\Thookrightarrow\wo obj. enc.\xspace}
\def\TAblationNoPositionalEncoding{\Thookrightarrow\wo pos. emb.\xspace}
\def\TAblationNoSolverRANSAC{\Thookrightarrow \wo robust est.\xspace}
\def\TAblationNoSolverScaleOpt{\Thookrightarrow \wo scale opt.\xspace}

\def\TTimeMilliSecond{Time\\\enparen{\unit{\milli\second\per\text{iter}}}}
\def\TAblationMatchFScorePct{F$_\text{1}$\enparen{\unit{\percent}}\,$\uparrow$}
\def\TAblationPlaneTR{\planetr}
\def\TAblationPlaneRecTR{\planerectr}
\def\TAblationZeroPlane{\zeroplane}
\def\TAblationPLANATHREER{\planathreer}
\def\TAblationOursDetector{\emph{\mogevtwo\cite{wangMoGe2AccurateMonocular2025}{\scriptsize +\seqRANSAC}}\,}
\def\TAblationGTDepthSeqRANSAC{GT.%
  Depth\emph{\scriptsize +\seqRANSAC}}
\def\TAblationGTDepthGTMask{GT.Depth+GT.Mask}

\def\TAblationMatchingPerformance{Plane Matching~\enparen{\unit{\percent}}\,$\uparrow$}
\def\TAblationMatchPrecisionPctNoArrow{\AbbrMatchPrecision}
\def\TAblationMatchRecallPctNoArrow{\AbbrMatchRecall}
\def\TAblationMatchFScorePctNoArrow{\AbbrMatchFScore}
\def\TAblationEncoderResNet{\enparen{1} ResNet50}
\def\TAblationEncoderDINONoNeck{\enparen{2} DINOv2}
\def\TAblationEncoderDINOWithNeck{\Thookrightarrow \enparen{3} \with Conv.
  head}
\def\TAblationEncoderOurs{Ours}

\def\TRebuttalMethods{Methods}
\def\TRebuttalMapType{Map\\ Type}
\def\TRebuttalMapSize{Map\\ Size\,$\downarrow$}
\def\TRebuttalMappingTime{Mapping\\ Time\,$\downarrow$}
\def\TRebuttalVisualAppearance{Visual\\ Cues}
\def\TRebuttalRetrieval{Pose\\ Prior}
\def\TRebuttalPoseNet{{\small PoseNet17~\cite{kendallGeometricLossFunctions2017}}}
\def\TRebuttalhLoc{{\small HLoc(SP+SG)~\cite{sarlinCoarseFineRobust2019}}}
\def\TRebuttalGoMatch{{\small GoMatch(SP)~\cite{zhouGeometryEnoughMatching2022}}}
\def\TRebuttalACE{{\small ACE~\cite{brachmannAcceleratedCoordinateEncoding2023}}}
\def\TRebuttalRelocr{{\small Reloc3r~\cite{dongReloc3rLargeScaleTraining2025}}}
\def\TRebuttalRelocrMapSize{\qty{1.6}{\giga\byte}}  %
\def\TRebuttalSTDLoc{{\small STDLoc~\cite{huangSparseDenseCamera2025}}}
\def\TRebuttalOurs{{\small \textbf{Ours}}}

\def\TRebuttalMapTypeNetwork{Network}
\def\TRebuttalMapTypeSfMPoints{SfM points}
\def\TRebuttalMapTypePoseImages{Ref. images}
\def\TRebuttalMapTypeGS{3DGS}
\def\TRebuttalMapTypePlanes{Planes}

\def\FAbbAccHigh{T$_{\text{3}}$}
\def\FAbbAccMed{T$_{\text{2}}$}
\def\FAbbAccCoarse{T$_{\text{1}}$}

\def\scenemap{\sM}
\def\querymap{\sQ}
\def\planarprimitive{\Pi}
\def\point{\rvx}
\def\planarprimitivequery{\planarprimitive^{q}}
\def\planarprimitivemap{\planarprimitive^{m}}
\def\planarprimitiveshape{\Omega}
\def\mapplanenum{{N_m}}
\def\queryplanenum{{N_q}}

\def\planenormal{\rvn}
\def\planeoffset{\evd}
\def\planeparameter{\bm{\pi}}
\def\planeparametermap{\planeparameter^{m}}
\def\planeparameterquery{\planeparameter^{q}}

\def\queryimage{\mI^{q}}
\def\queryintrinsics{\mK}
\def\querypixel{\rvu}
\def\projection{\rho}
\def\unprojection{\projection^{-1}}

\def\embedding{\rvf}

\def\rope{\text{\ttfamily RoPE}}
\def\roperot{\text{\ttfamily R}}
\def\assignment{\mA}
\def\similarity{\mS}
\def\matchability{\sigma}
\DeclareMathOperator*{\softmax}{\mathrm{Softmax}}
\DeclareMathOperator*{\sigmoid}{\mathrm{Sigmoid}}
\DeclareMathOperator*{\linear}{\mathrm{Linear}}

\def\correspondence{\mathcal{M}}
\def\inliercorrespondence{\widehat{\correspondence}}
\def\mnn{\text{\ttfamily MNN}}

\def\matchthreshold{\tau}
\def\gtmatchthreshold{\tau^*}
\def\unmatchableindexquery{\mathcal{U}^q}
\def\unmatchableindexmap{\mathcal{U}^m}
\def\lossmatch{\mathcal{L}_\mathrm{match}}

\def\querypose{\mathbf{P}}

\def\initRot{\Rot_0}
\def\initPos{\Pos_0}
\def\scale{s}
\def\weightforpos{\omega}

\def\refinedpose{\mathbf{T}_\mathrm{tr}}
\def\offsetseed{\delta}
\def\offsetseedname{offset seed}
\def\offsetseedsname{offset seeds\xspace}
\def\rendereddepth{\mD}
\def\depthsegment{\mathfrak{D}}
\def\perprimitiveresidual{r}
\def\depthcost{E_{\mathrm{depth}}}

\def\diagonal{\text{\ttfamily diag}}

\def\paperID{}
\def\confName{CVPR}
\def\confYear{2026}

\title{\methodname : Camera Relocalization in 3D Planar Primitives \\ via Region-Based Structure Matching}

\author{
{\hypersetup{urlcolor=black}\href{\hanqiaopage}{Hanqiao Ye$^{1,2}$}}%
\and Yuzhou Liu$^{1,2}$ %
\and Yangdong Liu$^{2}$\footnotemark[1] %
\and Shuhan Shen$^{1,2}$\footnotemark[1] %
\and \vspace{-1.2em}\\
$^{1}$School of Artificial Intelligence, University of Chinese Academy of Sciences \\
$^{2}$Institute of Automation, Chinese Academy of Sciences
\vspace{-.2em}\\
{\tt\small\hypersetup{urlcolor=black}  \{\href{mailto:yehanqiao2022@ia.ac.cn}{yehanqiao2022}, \href{mailto:liuyuzhou2021@ia.ac.cn}{liuyuzhou2021}, \href{mailto:yangdong.liu@ia.ac.cn}{yandong.liu}\}@ia.ac.cn; \href{mailto:shshen@nlpr.ia.ac.cn}{shshen@nlpr.ia.ac.cn}}
}

\begin{document}
\twocolumn[{%
      \renewcommand\twocolumn[1][]{#1}%
      \maketitle
      \vspace{-1.2cm}
      \begin{center}
        \centering
        \captionsetup{type=figure}
        \includegraphics[width=\linewidth]{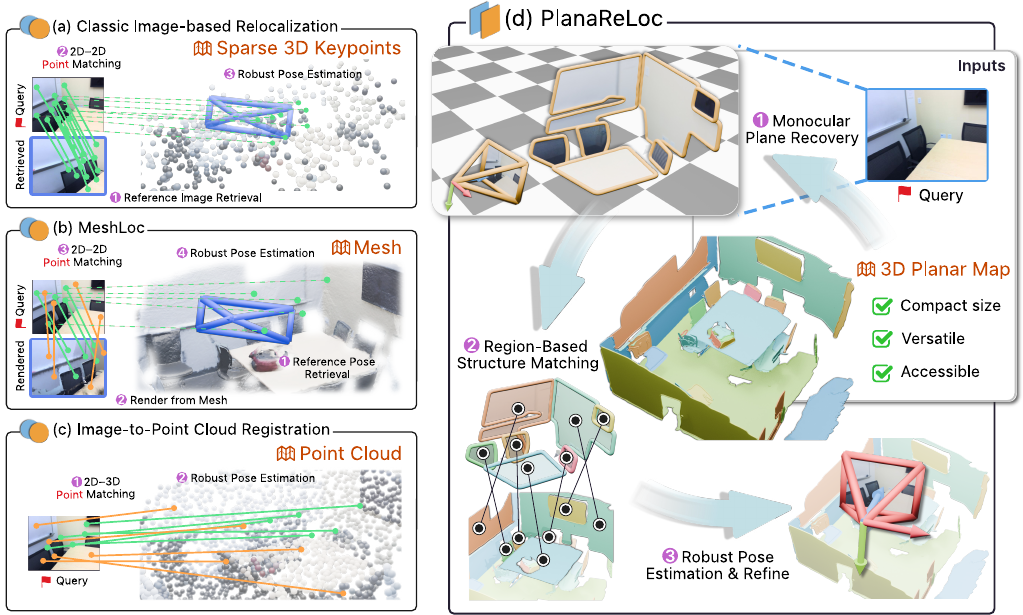}
        \captionof{figure}{\textbf{Overview.}
          \enparen{a-c}
          Existing structure-based camera relocalization approaches that establish \pointcorrespondenceicon\ \emph{point} correspondences on top of various \mapicon\ \emph{map} representations.
          \enparen{d}
          We propose a plane-centric paradigm that establishes cross-modal \planecorrespondenceicon\ \emph{plane} correspondences against the compact \mapicon\ \emph{plane-based 3D maps}, enabling lightweight and efficient 6-DoF camera relocalization in structured indoor environments.
        }
        \label{fig:teaser}
      \end{center}%
    }]
\footnotetext[1]{Corresponding authors.}
\begin{abstract}
  While structure-based relocalizers have long strived for point correspondences when establishing or regressing query-map associations, in this paper, we pioneer the use of planar primitives and 3D planar maps for lightweight 6-DoF camera relocalization in structured environments.
  Planar primitives, beyond being fundamental entities in projective geometry, also serve as region-based representations that encapsulate both structural and semantic richness.
  This motivates us to introduce \textbf{\mbox{\methodname}}, a streamlined ``plane-centric'' paradigm where a deep matcher associates planar primitives across the query image and the map within a learned unified embedding space, after which the 6-DoF pose is solved and refined under a robust framework.
  Through comprehensive experiments on the \scannet and \twelvescenes datasets across hundreds of scenes, our method demonstrates the superiority of planar primitives in facilitating reliable cross-modal structural correspondences and achieving effective camera relocalization without requiring realistically textured/colored maps, pose priors, or per-scene training.
  The code and data are available at \repourl.
\end{abstract}
\vspace{-3.5em}

\section{Introduction}
  \label{sec:introduction}
  Camera relocalization, the task of estimating the 6-DoF camera pose from a query image \wrt a known 3D environment, underpins real-time applications such as augmented reality~\enparen{AR}~\cite{zhaiSplatLoc3DGaussian2024a,brejchaLandscapeARLargeScale2020} and robot navigation~\cite{liActLocLearningLocalize2025,wangNeRFIBVSVisualServo2023}.

  One prevalent family of approaches, known as the structure-based methods, establishes \emph{point} correspondences between the query image and a pre-built scene representation which anchors landmarks in the coordinate space, and then solves for the camera pose within a robust estimation framework such as PnP-RANSAC~\cite{gaoCompleteSolutionClassification2003,fischlerRandomSampleConsensus1981}.
  Classic structure-based systems~\cite{humenbergerRobustImageRetrievalbased2022,sarlinCoarseFineRobust2019,camposecoHybridSceneCompression2019a}, as shown in~\cref{fig:teaser}\refcaptiona, rely on Structure-from-Motion~\enparen{SfM} techniques~\cite{schonbergerStructurefrommotionRevisited2016} to \change{construct}{triangulate} sparse 3D keypoints from posed reference images, where each point is associated with a visual descriptor for local feature matching.
  While leading to top accuracy, the SfM maps are costly to build and maintain, and image retrieval~\cite{arandjelovicNetVLADCNNArchitecture2016,torii247Place2015} or intricate search strategies~\cite{sattlerEfficientEffectivePrioritized2016,liuEfficientGlobal2d3d2017,sattlerImprovingImagebasedLocalization2012} are often required to narrow down matching candidates.
  Meanwhile, as illustrated in~\cref{fig:teaser}\refcaptionb, MeshLoc~\cite{panekMeshLocMeshbasedVisual2022} shows that modern point features can match real photos against non-photorealistic renderings of \emph{textured} meshes, thereby eliminating the need to store visual descriptors in the map representation.
  That said, the performance \change{deteriorates significantly}{degrades considerably} as the fidelity of scene appearance and geometry decreases~\cite{panekVisualLocalizationUsing2023,abeNormalLocVisualLocalization2025}.
  There also exist prior arts that utilize bearing vectors to match 2D pixels with sparse keypoints without relying on visual descriptors~\cite{campbellSolvingBlindPerspectivenpoint2020,zhouGeometryEnoughMatching2022,wangDGCGNNLeveragingGeometry2024,zhangA2GNNAngleAnnularGNN2025}.
  However, when applied to point clouds captured by depth sensors or LiDAR scans~\enparen{\cref{fig:teaser}\refcaptionc}, image-to-point cloud registration methods~\cite{li2D3DMATR2D3DMatching2023,wangFreeRegImagetoPointCloud2024,muDiff$^2$I2PDifferentiableImagetoPoint2025} struggle to generalize~\cite{anMinCDPnPLearning2D3D2025} and often fail to robustly predict cross-modal pixel\textendash point correspondences across the entire scene.

  Among various geometric entities that go beyond points, \emph{planar primitives} offer notable simplicity and compactness in representing physical surfaces.
  Consequently, 3D maps composed of planar primitives, \ie, \emph{3D planar maps}, are notably lean and well-suited for real-world applications such as AR~\cite{arkitPlacingContentDetected2025,arcoreFundamentalConceptsEnvironmental2025} and robotics~\cite{zhangCornerVINSAccurateLocalization2025,bavleSituationalGraphsRobot2022,shiPlaneMatchPatchCoplanarity2018}.
  This has further given rise to extensive research on constructing such maps from diverse sources, including not only multi-view reconstruction~\cite{xiePlanarReconRealtime3D2022,watsonAirPlanesAccuratePlane2024,heAlphaTabletsGenericPlane2024,ye2025neuralplane,tanPlanarSplattingAccuratePlanar2025}, but also raw point clouds~\cite{monszpartRAPterRebuildingManmade2015a,yuFindingGoodConfigurations2022}, as well as other modalities such as scene layouts~\cite{zhengStructured3DLargePhotorealistic2020,cruzZillowIndoorDataset2021}.
  Therefore, in this paper, we depart from prior structure-based methods that focus on point correspondences and instead capitalize on the ubiquity of planar surfaces in indoor environments, investigating 3D planar maps as a compact, versatile, and accessible form of scene representation toward 6-DoF camera relocalization.

  As illustrated in \cref{fig:teaser}\refcaptiond, we build upon the traditional feature-matching pipeline and propose a novel \emph{plane-centric} paradigm that establishes plane correspondences against \emph{untextured} 3D planar maps.
  We first extract plane segments from the query image and estimate their parameters by exploiting general-purpose monocular models.
  Then, by aggregating region-of-interest features and modeling interactions between cross-modal plane embeddings, we show that planar primitives enable \emph{direct} structure matching, eliminating the need for matching on virtual renderings.
  Finally, we introduce a robust framework with \postrefine that estimates the 6-DoF camera pose by leveraging established plane matches and their parameters.

  We summarize our \textbf{contributions} as follows:
  \begin{itemize}
    \item Given the region-based representation and favorable geometric properties, we place a premium on \emph{planar primitives} and investigate the use of \emph{3D planar maps} for leaner camera relocalization in structured environments.
    \item We propose \emph{\methodname}, a novel plane-centric paradigm for relocalization that matches cross-modal planar regions and estimates the 6-DoF poses, eliminating the need for realistic map textures, pose priors or per-scene training.
    \item As shown by experiments, the proposed pipeline, even with minimal specialized design, demonstrates the superiority of planar primitives in supporting reliable cross-modal matching and effective camera relocalization.
  \end{itemize}

\section{Related Work}
  \label{sec:related_work}

  \paragraph{Structure-Based Camera Relocalizers}
    typically establish query-map associations via feature matching~\cite{camposecoHybridSceneCompression2019a,sarlinCoarseFineRobust2019,humenbergerRobustImageRetrievalbased2022,liuEfficientGlobal2d3d2017,tairaInLocIndoorVisual2018,schonbergerSemanticVisualLocalization2018} or coordinate regression~\cite{shottonSceneCoordinateRegression2013,brachmannDSACDifferentiableRANSAC2016,brachmannAcceleratedCoordinateEncoding2023,dongVisualLocalizationFewShot2022,jiangRSCoReRevisitingScene2025}, followed by robust pose estimation~\cite{gaoCompleteSolutionClassification2003,fischlerRandomSampleConsensus1981,larssonPoseLibMinimalSolvers2020}.
    Most of them focus on establishing \emph{point} correspondences, while some also explore \emph{line segments} as complementary primitives~\cite{liu3DLineMapping2023,liuLightweightStructuredLine2024,hrubyEfficientSolutionPointline2024,pautratGlueStickRobustImage2023,ramalingamPoseEstimationUsing2011}.
    Beyond ad hoc maps constructed from posed reference images, more general scene representations have been explored for this task, including textured meshes~\cite{panekMeshLocMeshbasedVisual2022}, NeRF~\cite{yen-chenINeRFInvertingNeural2021,moreauCROSSFIRECameraRelocalization2023,zhouNeRFectMatchExploring2024}, and 3DGS~\cite{huangSparseDenseCamera2025,pietrantoniGaussianSplattingFeature2025,wang3DGaussianSplatting2025}.
    While \emph{visual appearance} heavily dominates the association process in these methods,
    some alternatives attempt to directly register images to point clouds without visual cues~\cite{li2D3DMATR2D3DMatching2023,wangFreeRegImagetoPointCloud2024,muDiff$^2$I2PDifferentiableImagetoPoint2025,anMinCDPnPLearning2D3D2025}, yet they struggle to produce stable cross-modal correspondences across the entire scene.
    Other methods explore higher-level maps like floorplans~\cite{howard-jenkinsLalalocLatentLayout2021,gardSPVLocSemanticPanoramic2024,chenF$^3$LocFusionFiltering2024, liuPolyRoomRoomawareTransformer2024} and LoD models~\cite{abeNormalLocVisualLocalization2025,juelinzhuLoDlocVisualLocalization2024, liuBWFormerBuildingWireframe2025}, which, however, often rely on pose priors or exhaustive render-and-compare strategies.

    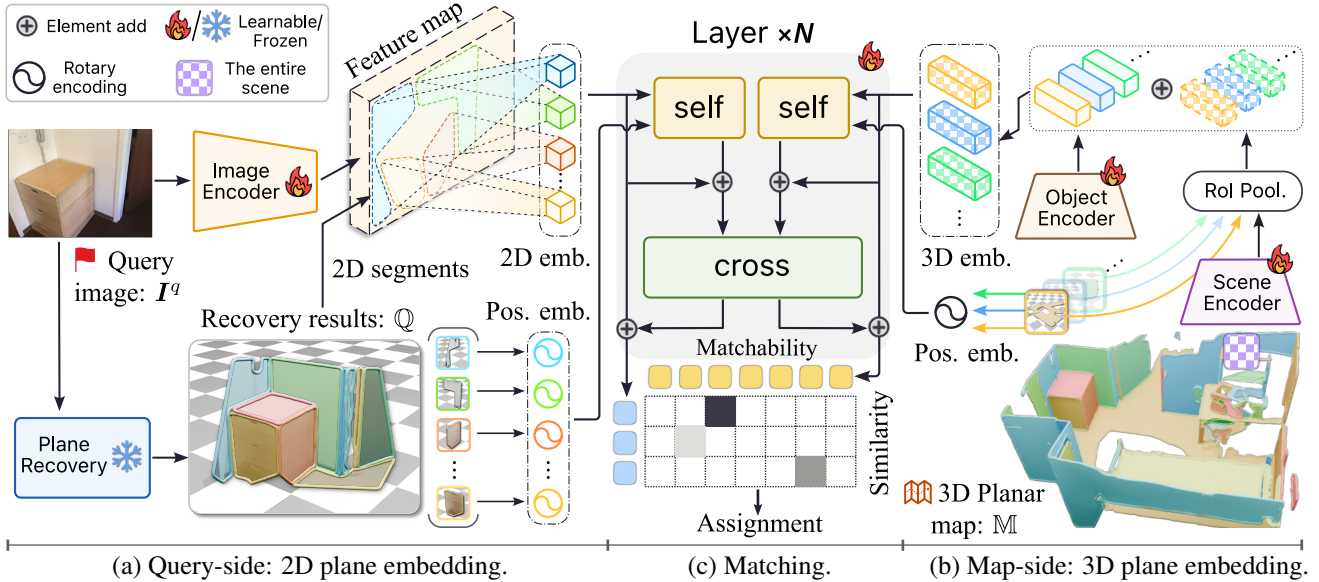
\begin{figure*}[!t]
      \centering

      \begin{tikzpicture}
        \node[anchor=south west,inner sep=0] (image) at (-2.5,0) {
          \includegraphics{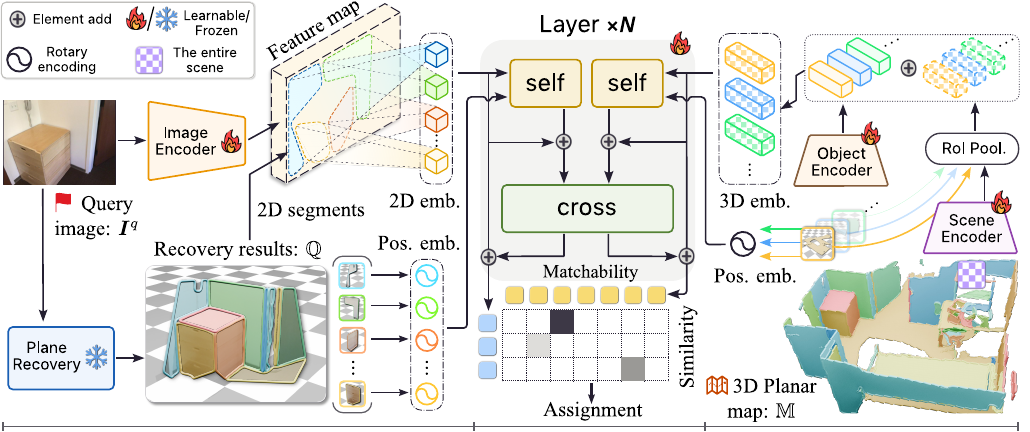}
        };

        {
        \pgfmathsetmacro{\y}{-0.03}
        \pgfmathsetmacro{\xtwod}{0.1}
        \pgfmathsetmacro{\xmatching}{0.5}
        \pgfmathsetmacro{\xthreed}{0.82}
        \begin{scope}[x={(image.south east)},y={(image.north west)}]
          \node[fill=white, fill opacity=0.7, text opacity=1, inner sep=2pt] at (\xtwod, \y) {\captiona Query-side: 2D plane embedding.};
          \node[fill=white, fill opacity=0.7, text opacity=1, inner sep=2pt] at (\xmatching, \y) {\captionc Matching.};
          \node[fill=white, fill opacity=0.7, text opacity=1, inner sep=2pt] at (\xthreed, \y) {\captionb Map-side: 3D plane embedding.};
        \end{scope}
        }
      \end{tikzpicture}
      \caption{\textbf{Overview of the planar primitive embedding~\enparen{\cref{sec:front-end}} and matching~\enparen{\cref{sec:plane-matching}} pipeline between the \queryicon\ query and the \mapicon\ map.}
        \captiona The query image is first reconstructed into a set of 3D planar primitives via a frozen monocular plane recovery module, with each primitive further encoded into a plane embedding by aggregating visual features within its corresponding 2D segment.
        \captionb Each map primitive is encoded by an object encoder and a scene encoder, capturing both its shape and pose features.
        \captionc The two sets of embeddings are fed into a stack of transformer layers to produce a soft assignment matrix, from which plane correspondences are inferred.
      }
      \label{fig:matching-pipeline}
      \vspace{-0.5em}
    \end{figure*}

  \paragraph{Plane-Based 3D Representation}
    centers on \emph{planar primitives} as its fundamental building blocks, benefiting from their compact form and prevalence in structured environments.
    The classic task of monocular plane recovery shows that precise and semantically well-aligned planar primitives can be inferred by jointly predicting 2D segmentations and their corresponding plane parameters~\cite{liuPlaneNetPiecewisePlanar2018,liuPlaneRCNN3DPlane2019,yuSingleimagePiecewisePlanar2019}.
    Such compactness is also demonstrated in organizing room-scale multiview reconstructions~\cite{liuPlaneMVS3DPlane2022,xiePlanarReconRealtime3D2022,
      chenPlanarNeRFOnlineLearning2025,tanPlanarSplattingAccuratePlanar2025,ye2025neuralplane} or LiDAR scans~\cite{nanPolyFitPolygonalSurface2017,monszpartRAPterRebuildingManmade2015a,sommerPlanesCornersMultiPurpose2020} into piecewise planar representations, \ie, 3D planar maps.
    More closely related, several methods estimate relative poses by exploiting plane correspondences~\cite{raposoPlanebasedOdometryUsing2013,shiPlaneMatchPatchCoplanarity2018,raposoPiecewiseplanarStereoScanSequential2017,wietrzykowskiPlaneLocProbabilisticGlobal2019}, which exhibit robustness under challenging scenarios such as extreme viewpoint changes~\cite{jinPlanarSurfaceReconstruction2021,agarwalaPlaneFormersSparseView2022,tanNOPESACNeuralOneplane2023,shiPlaneRecTRUnifiedQuery2025,liuPLANA3RZeroshotMetric2025}.
    Motivated by these works, we further explore planar primitives for camera relocalization.

\section{Method}
  \label{sec:method}

  Our goal is to study how planes can enable a lean relocalization pipeline.
  In light of this, we introduce \methodname, which estimates the 6-DoF pose of a query image $\queryimage$ \wrt a scene mapped as a collection of piecewise planar surfaces $\scenemap=\{\planarprimitivemap_i\}_{i=1}^{\mapplanenum}$.
  Each map primitive $\planarprimitivemap_i$ is defined by its plane parameters $\planeparametermap$ and a bounding shape $\planarprimitiveshape^m$.

  Our three-stage pipeline starts by %
  establishing plane correspondences between the query and the map~\enparen{%
    \cref{sec:front-end,sec:plane-matching}%
  },
  then it performs robust pose estimation~\enparen{\cref{sec:pose-estimation}},
  and finally concludes with \postrefine~\enparen{\cref{sec:pose-refinement}}.

  \subsection{Front-End: Planar Primitive Embedding}
    \label{sec:front-end}

    To bridge the modality gap between the query image $\queryimage$ and the \mapname $\scenemap$,
    our front-end module first rebuilds $\queryimage$ into a set of planar primitives,
    and then projects all primitives, from both $\queryimage$ and $\scenemap$, into their respective embedding spaces for subsequent matching, as depicted in \cref{fig:matching-pipeline}.

    \paragraph{Monocular Plane Recovery.}
      Recovering 3D planes from a single query image is a joint task \change{of}{involving} class-agnostic instance segmentation and metric-scale plane parameter estimation, which has been significantly advanced by end-to-end learning frameworks~\cite{%
        liuPlaneNetPiecewisePlanar2018,
        yangRecovering3DPlanes2018,
        liuPlaneRCNN3DPlane2019,
        yuSingleimagePiecewisePlanar2019,
        tanPlaneTRStructureGuidedTransformers2021,
        shiPlaneRecTRUnifiedQuery2023,
        liuInthewild3DPlane2025%
      }.
      While these off-the-shelf models are trained on large-scale datasets and can distinguish two adjacent yet coplanar semantic entities, \eg, a closed door and its surrounding wall, we opt for a purely geometric module, \ie, sequentially fitting planes on the predicted geometry, based on the observation that the strong geometric priors provided by vision foundation models~\cite{%
        wangDUSt3RGeometric3D2024,
        keRepurposingDiffusionbasedImage2024,
        bochkovskiy2025depth,
        wangMoGeUnlockingAccurate2025,
        wangMoGe2AccurateMonocular2025%
      }
      are sufficient to achieve good performance%
      .
      The output, denoted as $\querymap=\{\planarprimitivequery_i \mid i=1,\cdots,\queryplanenum\}$, comprises a collection of query primitives, each with predicted \emph{metric} plane parameters $\planeparameterquery$ and a binary 2D segment mask $\planarprimitiveshape^q\in \mathds{1}^{H\times W}$ representing the shape.
      Here we clarify that the metric scale, though error-prone, caters to the crucial initial ``guess'' for pose estimation, which will be elaborated in \cref{sec:pose-estimation}.

    \paragraph{2D Plane Embeddings.}
      Encoding \change{a 2D instance-level}{an instance-level 2D} representation is essentially aggregating patch-level visual features within its Region of Interest~\enparen{RoI}.
      RoI-Align~\cite{heMaskRCNN2017} followed by a fully connected layer as in \cite{liMatchingAnythingSegmenting2024} is a common practice.
      However, akin to prior works~\cite{%
        shlapentokh-rothmanRegionBasedRepresentationsRevisited2024,
        zhangMultiviewSceneGraph2024,
        khoslaRELOCATESimpleTrainingfree2025%
      }, we find that a simple average pooling performs well.
      As shown in \cref{fig:matching-pipeline}\refcaptiona, the query image is first patchified by a pretrained encoder and reshaped into a feature map.
      Then, we resize the 2D segments $\{\planarprimitiveshape^q_i\}$ accordingly and apply average pooling within each segment to aggregate features, yielding query-side 2D plane embeddings $\{\embedding^q_i\in\sR^{c}\}_{i=1}^{\queryplanenum}$.

    \paragraph{3D Plane Embeddings.%
      }
      The input map is \emph{untextured}, \ie, a structure-only scene representation.
      Therefore, a map primitive $\planarprimitivemap$ is fully described by its shape $\planarprimitiveshape^m$ and plane parameters $\planeparametermap$.
      This calls for two separate 3D encoders, namely, an object encoder and a scene encoder, to respectively capture the shape and spatial pose characteristics of each $\planarprimitivemap$, as illustrated in \cref{fig:matching-pipeline}\refcaptionb.
      The object encoder takes \emph{batched} map primitives $\mathbf{\Pi}^m\in\sR^{\mapplanenum\times L\times 3}$ as input for shape embeddings, where each primitive is centralized and represented as a uniformly sampled point cloud of length $L$.
      Meanwhile, the scene encoder processes the \emph{entire} scene to produce \emph{point-wise} features,
      from which we aggregate RoI features for each $\planarprimitivemap$, \ie, features of points belonging to $\planarprimitivemap$, into a pose-aware spatial embedding via max pooling.

      Embeddings from both encoders are fused via an $\alpha$-weighted sum, with $\alpha$ a learnable parameter.
      The resulting embeddings $\{\embedding^m_i\in\sR^{c}\}_{i=1}^{\mapplanenum}$ are expected to encode both the shape and spatial pose for each map primitive.

  \subsection{Matching Planar Primitives Like Points}
    \label{sec:plane-matching}
    Constructing the contrastive loss~\cite{%
      chopraLearningSimilarityMetric2005,
      vandenoordRepresentationLearningContrastive2019,
      khoslaSupervisedContrastiveLearning2020,
      chenSimpleFrameworkContrastive2020%
    }
    is a common approach to learn a discriminative embedding space for matching, particularly for instance-level and cross-modal features~\cite{%
      radfordLearningTransferableVisual2021,
      sarkarSGAligner3DScene2023,
      liMatchingAnythingSegmenting2024,
      miaoSceneGraphLocCrossModalCoarse2024,
      sarkarCrossOver3DScene2025%
    }.
    However, we observe that planar primitives, despite their region-based form and latent semantics, are essentially class-agnostic geometric entities that exhibit recurring patterns, potentially leading to detrimental hard negatives.%
    Instead, we maximize the log-likelihood of assignment matrices, adopting a scheme similar to those used in matching points~\cite{%
      sarlinSuperGlueLearningFeature2020,
      sunLoFTRDetectorFreeLocal2021,
      qinGeoTransformerFastRobust2023,
      lindenbergerLightGlueLocalFeature2023,
      pautratGlueStickRobustImage2023%
    }.

    \paragraph{Architecture.}
      Embeddings from both sides, $\{\embedding^q_i\}$ and $\{\embedding^m_i\}$, are fed into a stack of $N$ identical transformer layers~\cite{vaswaniAttentionAllYou2017} that process the two sets without distinction, as illustrated in \cref{fig:matching-pipeline}\refcaptionc.
      Each layer is a succession of one self- and one cross-attention unit, which together refine the representation of each primitive in the context of all the others.

    \paragraph{Positional Embedding.}
      To reinforce the sense of relative pose when reasoning over different entities, we retrofit the self-attention score between two unimodal planes as
      $a_{ij} = \rvq_i^\top\rope(\planenormal_j - \planenormal_i)\,\rvk_j$,
      where $\rvq_i$ and $\rvk_j$ denote the query and key vectors projected from the plane embeddings $\embedding_i$ and $\embedding_j$, respectively.
      The rotary encoding~\cite{suRoFormerEnhancedTransformer2023} $\rope(\cdot)$ constructs a $c\times c$ positional embedding from two plane normals $\planenormal_i$ and $\planenormal_j$, representing the relative rotation between these two planes while remaining equivariant \wrt the camera pose.
      Following~\cite{%
        liLearnableFourierFeatures2021,
        lindenbergerLightGlueLocalFeature2023%
      }, the $c$-dimensional embedding space is partitioned into $c/2$ 2D subspaces, each being rotated by an angle computed as the inner product with a learnable basis vector $\rvb_k\in\sR^3$, where $k\in[1, c/2]$:
      {
      \begin{equation}
        \rope(\planenormal) = \left(
        \begin{smallmatrix}
          \roperot(\rvb_1^\top\planenormal) & & \\
          & \ddots & \\
          & & \roperot(\rvb_{c/2}^\top\planenormal)
        \end{smallmatrix}
        \right),\,
        \roperot(\theta) = \left(
        \begin{smallmatrix}
          \cos\theta & -\sin\theta \\
          \sin\theta & \cos\theta
        \end{smallmatrix}
        \right).
      \end{equation}
      }

    \paragraph{Correspondences.}
      Following~\cite{lindenbergerLightGlueLocalFeature2023}, the soft assignment matrix $\assignment\in\sR^{\queryplanenum\times \mapplanenum}$ is formulated as the combination of both matchability scores and similarity:
      \begin{equation}
        \assignment_{ij} = \matchability^q_i\matchability^m_j\,\softmax_{k\in[1,\queryplanenum]}(\similarity_{kj})_i\,\softmax_{k\in[1,\mapplanenum]}(\similarity_{ik})_j.
      \end{equation}

      The matchability score $\matchability_i$ for plane $i$, signifying its likelihood of contributing to a correspondence, is predicted by a learnable sigmoid-activated linear layer:{
      \small
      $
        \matchability_i = \sigmoid(\linear(\embedding_i))\,\in [0, 1].
      $
      }
      The pairwise similarity $\similarity_{ij}$ between the query embedding $\embedding^q_i$, {\small $i\in[1,\queryplanenum]$} and the map embedding $\embedding^m_j$, {\small $j\in[1,\mapplanenum]$} is computed as:
      {
      \small
      $
        \similarity_{ij} = \linear(\embedding^q_i) \cdot \linear(\embedding^m_j).
      $
      }

      Finally, a pair of cross-modal primitives $(\planarprimitivequery_i, \planarprimitivemap_j)$ constitutes a correspondence if~\enparen{1} both planes are predicted as matchable and
      \enparen{2} the similarity score $\similarity_{ij}$ stands out in both the $i$-th row and $j$-th column.
      More formally, correspondence predictions are selected from $\assignment$ by enforcing a confidence threshold $\matchthreshold$ and the Mutual Nearest Neighbor~\enparen{MNN} criterion:
      $
        \correspondence = \{(i, j)\,|\,\forall (i, j) \in \mnn(\assignment), \assignment_{ij} > \matchthreshold\}.
      $

    \paragraph{Supervision.}
      We precompute the projections of map primitives into the training queries using ground-truth camera poses.
      This enables on-the-fly generation of matching labels $\correspondence^*$ during training.
      Specifically, for each recovered query primitive $\planarprimitivequery_i$, its ground-truth correspondence is defined as the map primitive whose 2D projection $\planarprimitiveshape^{m\rightarrow q}_j$ has the highest Intersection-over-Union~\enparen{IoU} overlap with the query segment $\planarprimitiveshape^q_i$.
      Note that, for the sake of simplicity and due to the inherent ambiguity and uncertainty in defining plane shapes, we avoid the use of the bipartite matching strategy as in~\cite{%
        chengMaskedattentionMaskTransformer2022,
        shiPlaneRecTRUnifiedQuery2023,
        liuInthewild3DPlane2025%
      }, allowing for one-to-many plane correspondences, \ie, one map primitive may corresponds to multiple query primitives.
      This is often the case when planes detected in the query are over-segmented due to occlusions and surface discontinuities.
      Moreover, planar primitives with a mask IoU lower than $\gtmatchthreshold$ are labeled as unmatchable and indexed by $\unmatchableindexquery\subseteq [1, \queryplanenum]$ and $\unmatchableindexmap\subseteq [1, \mapplanenum]$.

      The training objective is to minimize the negative log-likelihood of the assignment and unmatchable predictions:
      {\small
      \begin{equation}
        \begin{split}
          \lossmatch = - \bigg( & \frac{1}{|\correspondence^*|}\sum_{(i,j)\in \correspondence^*} \log \assignment_{ij} + \frac{1}{2|\unmatchableindexquery|}\sum_{i\in \unmatchableindexquery} \log ( \\
                                & 1 - \matchability^q_i) + \frac{1}{2|\unmatchableindexmap|}\sum_{j\in \unmatchableindexmap} \log (1 - \matchability^m_j)\bigg)\enspace.
        \end{split}
        \label{equ:loss}
      \end{equation}
      }%
      To speed up training, we impose $\lossmatch$ at each of the $N$ layers to deeply supervise the overall learning as in~\cite{lindenbergerLightGlueLocalFeature2023}.

  \subsection{Pose Estimation from Plane Correspondences}
    \label{sec:pose-estimation}

    In addition to the region-based representation that delivers effective feature aggregation for cross-modal \matchingtwothreeD matching, the planar primitives also serve as fundamental parametric entities in projective geometry~\cite{hartleyProjectiveGeometryTransformations2003}, allowing for straightforward pose estimation from correspondences.

    \paragraph{Problem Formulation.}
      We define the camera pose $\querypose:=[\Rot\,|\,\Pos]\in \mathrm{SE}(3)$ as a rigid transformation composed of a rotation $\Rot\in \mathrm{SO}(3)$ and a translation $\Pos\in \mathbb{R}^3$, mapping coordinates from the camera space to the map space.
      Estimating $\querypose$ can be formulated as a registration problem over the set of putative plane correspondences $\{(\planarprimitivequery_i, \planarprimitivemap_j)\mid(i, j)\in\correspondence\}$, where each plane is associated with parameters $\planeparameter=[\planenormal^\top, \planeoffset\,]^\top$ defined in its respective coordinate space.
      However, it should be noted that the predicted correspondences may contain outliers, and the monocular front-end is inevitably noisy, leading to inaccurate plane parameters $\{\planeparameterquery_i\}$ for the query primitives.

    \paragraph{Robust Estimation.}
      First, in 3D projective space, points and planes form a dual pair~\cite{hartleyProjectiveGeometryTransformations2003,raposoPiecewiseplanarStereoScanSequential2017}.
      Given the point transformation $\point^m = \querypose\point^q$, a plane transforms as:
      \begin{equation}
        \planeparametermap \sim \underbrace{
        \begin{pmatrix}
          \Rot        & \Pos \\
          \vzero^\top & 1
        \end{pmatrix}^{-\top}
        }_{\querypose^{-\top}}\planeparameterquery \sim  \begin{pmatrix}
          \Rot           & \vzero \\
          -\Pos^\top\Rot & 1
        \end{pmatrix}\planeparameterquery,
        \label{equ:plane-transform}
      \end{equation}
      where $\sim$ denotes equality up to a non-zero scale.
      \Cref{equ:plane-transform} indicates that the plane normal is rotated by $\Rot$ independently of $\Pos$, whereas the plane offset depends on both $\Rot$ and $\Pos$. This yields:
      \begin{align}
        \planenormal^m & = \Rot\planenormal^q,                                                                                                  \\
        \planeoffset^m & = \planeoffset^q - \Pos^\top\Rot\planenormal^q = \planeoffset^q - \Pos^\top\planenormal^m.\label{equ:pose-translation}
      \end{align}

      Based on the above relations and \cite{hornClosedformSolutionAbsolute1987}, we first derive a minimal solver that uniquely determines the camera rotation $\Rot$ from two pairs of plane correspondences with non-parallel normals.
      Next, we apply \seqRANSAC~\cite{fischlerRandomSampleConsensus1981} to randomly sample minimal sets of such two pairs of correspondences, generate rotation hypotheses, and select the hypothesis with the most inliers.
      The largest inlier set is denoted as $\inliercorrespondence$, from which we estimate the initial camera rotation $\initRot$ using the Kabsch algorithm~\cite{kabschSolutionBestRotation1976}.

      The solution for translation $\Pos$ in \cref{equ:pose-translation}, however, requires at least three non-parallel pairs of correspondences.
      Given the correspondences in $\inliercorrespondence$, we estimate the initial translation $\initPos$ alongside the scale factor $\scale$ in the following weighted least squares problem, which compensates for the metric ambiguity in the monocular front-end:
      \begin{equation}
        \initPos, \scale^* = \argmin_{\Pos, s} \sum_{(i,j)\in \inliercorrespondence} \weightforpos_i(\Pos^\top\planenormal^m_j-\planeoffset^m_j+\scale\planeoffset^q_i)^2.
        \label{equ: estimating-camera-translation}
      \end{equation}
      The weight $\weightforpos_i$, indicating the reliability of $\planarprimitivequery_i$, is measured by the size of its 2D segment $\planarprimitiveshape_i$ based on the intuition that larger planes are typically better recovered and matched.

      \begin{figure}[t]
        \centering
        \includegraphics[width=\linewidth]{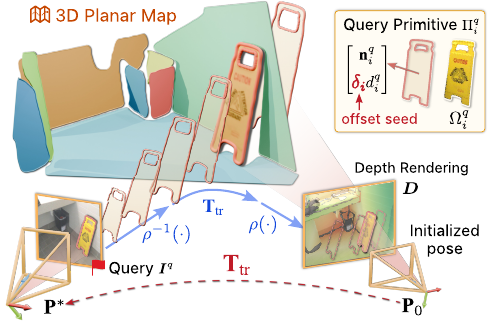}
        \caption{\textbf{Pose refinement via per-primitive depth alignment.}
          Given the optimization variables\textemdash the \offsetseedname\,$\offsetseed_i$ and $\refinedpose$\textemdash the query primitive $\planarprimitivequery_i $ \textcolor{warpOnto}{is warped onto} the depth rendering $\rendereddepth$ via \cref{equ:primitive-warping}.
          Then, the depth alignment error is computed in \cref{equ:per-primitive-depth-alignment-error}.
        }
        \label{fig:per-primitive-depth-alignment}
        \vspace{-.5em}
      \end{figure}

      \begin{table*}[h!t]
        \centering
        \caption{
          \textbf{Relocalization results on \scannet.}
          For each method, a ``\Tcheckmark'' denotes the use of auxiliary map truncation, coarse pose initialization, or realistic map appearance.
          We report rotation and translation errors, along with pose recalls\textemdash the ratio of successfully localized queries across thresholds.
          Top-3 results are highlighted as the {\setlength{\fboxsep}{1pt}\colorbox{tableRankFirst}{\textbf{first}}}, {\setlength{\fboxsep}{1pt}\colorbox{tableRankSecond}{\emph{second}}}, and {\setlength{\fboxsep}{1pt}\colorbox{tableRankThird}{third}}.
          We also provide the average runtime per query.
        }
        \setlength{\tabcolsep}{5.5pt} %
        \resizebox{\textwidth}{!}{%
          
\begin{tabular}{p{3pt}lccccccccccc}
  \Xhline{3\arrayrulewidth}                                                                                                                                                                                                                                                                                                                                                                                                                                                                                                                                                                                                                                                                                                  \\[-0.9em]
                                                       &                                                                                                 & \multirow{2}{*}{\rotatebox[origin=c]{0}{\makecell[t]{\TTruncatedMap}}} & \multirow{2}{*}{\rotatebox[origin=c]{0}{\makecell[t]{\TCoarseInit}}} & \multirow{2}{*}{\rotatebox[origin=c]{0}{\makecell[t]{\TTextured}}} & \multicolumn{2}{c}{\TRotErr}  & \multicolumn{2}{c}{\TPosErr} & \multicolumn{3}{c}{\TAccThres} & \multirow{2}{*}{\rotatebox[origin=c]{0}{\makecell[t]{\TTime}}}                                                                                                                                                                               \\[-0.05em]
  \cmidrule(r){6-7} \cmidrule(r){8-9} \cmidrule(r){10-12}                                                                                                                                                                                                                                                                                                                                                                                                                                                                                                                                                                                                                                                                    \\[-1.3em]
                                                       & \TMethods                                                                                       &                                                                        &                                                                      &                                                                    & \TRotErrMean                  & \TRotErrMed                  & \TPosErrMean                   & \TPosErrMed                                                    & \TAccFine                     & \TAccMed                      & \TAccCoarse                   &                                                                             \\[-0.05em]
  \Xhline{.3\arrayrulewidth}                                                                                                                                                                                                                                                                                                                                                                                                                                                                                                                                                                                                                                                                                                 \\[-0.9em]
  \rowcolor{\TNoRankColor}                             & \Tcoarseinit                                                                                    & \Tcheckmark                                                            &                                                                      &                                                                    & \tablenum{32.7}               & \tablenum{28.7}              & \tablenum{1.00}                & \tablenum{0.94}                                                & \tablenum{0.4}                & \tablenum{5.3}                & \tablenum{33.9}               & \Tna                                                                        \\[0.05em]
  \Xhline{.3\arrayrulewidth}                                                                                                                                                                                                                                                                                                                                                                                                                                                                                                                                                                                                                                                                                                 \\[-0.9em]
  \multirow{4}{*}{\rotatebox[origin=t]{90}{\Timgtopc}} & \TgeotransformerF~\cite{qinGeoTransformerFastRobust2023}                                        &                                                                        &                                                                      &                                                                    & \tablenum{53.7}               & \tablenum{42.1}              & \tablenum{1.93}                & \tablenum{1.80}                                                & \tablenum{17.0}               & \tablenum{26.4}               & \tablenum{29.0}               & \Tsim\tablenum[round-mode = places, round-precision=1]{0.428735}            \\[0.05em]
                                                       & \TgeotransformerT~\cite{qinGeoTransformerFastRobust2023}                                        & \Tcheckmark                                                            &                                                                      &                                                                    & \tablenum{45.2}               & \tablenum{26.5}              & \tablenum{1.42}                & \tablenum{1.06}                                                & \tablenum{24.6}               & \tablenum{38.8}               & \tablenum{42.9}               & \Tsim\tablenum[round-mode = places, round-precision=1]{0.297699}            \\[0.05em]
                                                       & \Tfreereg~\cite{wangFreeRegImagetoPointCloud2024}                                               &                                                                        & \Tcheckmark                                                          &                                                                    & \tablenum{36.9}               & \tablenum{29.4}              & \tablenum{1.06}                & \tablenum{0.96}                                                & \tablenum{0.8}                & \tablenum{6.4}                & \tablenum{33.1}               & \Tsim\tablenum[round-mode = places, round-precision=1]{14.187742}           \\[0.05em]
                                                       & \Tffreereg~\cite{wangFreeRegImagetoPointCloud2024}                                              & \Tcheckmark                                                            &                                                                      &                                                                    & \tablenum{40.7}               & \tablenum{27.2}              & \tablenum{2.14}                & \tablenum{1.41}                                                & \tablenum{13.7}               & \tablenum{26.3}               & \tablenum{36.2}               & \Tsim\tablenum[round-mode = places, round-precision=1]{11.143295}           \\[0.05em]
  \Xhline{.3\arrayrulewidth}                                                                                                                                                                                                                                                                                                                                                                                                                                                                                                                                                                                                                                                                                                 \\[-0.9em]
  \multirow{6}{*}{\rotatebox[origin=t]{90}{\Tmeshloc}} & \Tsplg~\cite{detoneSuperPointSelfsupervisedInterest2018a,lindenbergerLightGlueLocalFeature2023} &                                                                        & \Tcheckmark                                                          & \Tcheckmark                                                        & \tablenum{58.9}               & \tablenum{43.3}              & \tablenum{1.38}                & \tablenum{1.19}                                                & \tablenum{11.7}               & \tablenum{19.5}               & \tablenum{32.0}               & \Tsim\tablenum[round-mode = places, round-precision=1]{0.307796}            \\[0.05em]
                                                       & \Tloftr~\cite{sunLoFTRDetectorFreeLocal2021}                                                    &                                                                        & \Tcheckmark                                                          & \Tcheckmark                                                        & \tablenum{44.4}               & \tablenum{14.2}              & \TRankThird{\tablenum{0.86}}   & \tablenum{0.51}                                                & \tablenum{33.5}               & \tablenum{46.6}               & \TRankThird{\tablenum{58.0}}  & \Tsim\tablenum[round-mode = places, round-precision=1]{0.35266}             \\[0.05em]
                                                       & \Tmastr~\cite{leroyGroundingImageMatching2024}                                                  &                                                                        & \Tcheckmark                                                          & \Tcheckmark                                                        & \tablenum{46.0}               & \TRankThird{\tablenum{12.2}} & \tablenum{1.02}                & \TRankThird{\tablenum{0.43}}                                   & \TRankThird{\tablenum{35.4}}  & \TRankThird{\tablenum{49.5}}  & \tablenum{57.6}               & \Tsim\tablenum[round-mode = places, round-precision=1]{0.683168}            \\[0.05em]
                                                       & \Tmatchanything~\cite{heMatchAnythingUniversalCrossModality2025}                                &                                                                        & \Tcheckmark                                                          &                                                                    & \tablenum{35.7}               & \tablenum{19.9}              & \tablenum{1.23}                & \tablenum{0.74}                                                & \tablenum{20.0}               & \tablenum{35.7}               & \tablenum{52.1}               & \Tsim\tablenum[round-mode = places, round-precision=1]{0.925184}            \\[0.05em]
                                                       & \Tnopesac~\cite{tanNOPESACNeuralOneplane2023}                                                   &                                                                        & \Tcheckmark                                                          & \Tcheckmark                                                        & \tablenum{28.7}               & \tablenum{15.9}              & \tablenum{0.90}                & \tablenum{0.77}                                                & \tablenum{3.3}                & \tablenum{21.2}               & \tablenum{54.6}               & \Tsim\tablenum[round-mode = places, round-precision=1]{0.368222}            \\[0.05em]
                                                       & \Tplanathreer~\cite{liuPLANA3RZeroshotMetric2025}                                               &                                                                        & \Tcheckmark                                                          & \Tcheckmark                                                        & \TRankThird{\tablenum{26.8}}  & \tablenum{12.9}              & \tablenum{0.92}                & \tablenum{0.52}                                                & \tablenum{17.9}               & \tablenum{37.6}               & \tablenum{57.1}               & \Tsim\tablenum[round-mode = places, round-precision=1]{0.423516}            \\[0.05em]
  \Xhline{.3\arrayrulewidth}\noalign{\vskip 3pt}\Xhline{.3\arrayrulewidth}                                                                                                                                                                                                                                                                                                                                                                                                                                                                                                                                                                                                                                                   \\[-0.9em]
  \multirow{2}{*}{\rotatebox[origin=t]{90}{\Tours}}    & \Toursnorefine                                                                                  &                                                                        &                                                                      &                                                                    & \TRankSecond{\tablenum{17.3}} & \TRankSecond{\tablenum{3.9}} & \TRankSecond{\tablenum{0.65}}  & \TRankSecond{\tablenum{0.27}}                                  & \TRankSecond{\tablenum{37.1}} & \TRankSecond{\tablenum{69.8}} & \TRankSecond{\tablenum{79.8}} & \Tsim{\bfseries\tablenum[round-mode = places, round-precision=1]{0.059854}} \\[0.05em]
                                                       & \Toursfull                                                                                      &                                                                        &                                                                      &                                                                    & \TRankFirst{\tablenum{17.2}}  & \TRankFirst{\tablenum{3.8}}  & \TRankFirst{\tablenum{0.60}}   & \TRankFirst{\tablenum{0.20}}                                   & \TRankFirst{\tablenum{48.5}}  & \TRankFirst{\tablenum{73.1}}  & \TRankFirst{\tablenum{81.8}}  & \Tsim\tablenum[round-mode = places, round-precision=1]{0.544342}            \\[0.05em]
  \Xhline{3\arrayrulewidth}                                                                                                                                                                                                                                                                                                                                                                                                                                                                                                                                                                                                                                                                                                  \\[-0.9em]
\end{tabular}

        }
        \label{tab:scannetv2-pose-results}
        \vspace{-.75em}
      \end{table*}

  \subsection{Primitive-Based Pose Refinement}
    \label{sec:pose-refinement}
    In the visual localization literature, an initialized camera pose can be further refined through various render-and-compare strategies, depending on the \change{exact}{specific} map representation, such as the NeRF/3DGS-based~\cite{%
      liuGSCPREfficientCamera2025,
      chenRefinementAbsolutePose2024,
      zhaoPNeRFLocVisualLocalization2024,
      linBARFBundleadjustingNeural2021%
    } and LoD/Floorplan-based~\cite{%
      juelinzhuLoDlocVisualLocalization2024,
      zhuLoDlocV2Aerial2025,
      howard-jenkinsLalalocLatentLayout2021,
      howard-jenkinsLalalocGlobalFloor2022,
      chenF$^3$LocFusionFiltering2024,
      graderSuperchargingFloorplanLocalization2025a%
    } approaches.
    In this work, we draw inspiration from per-primitive photometric alignment proposed by~\cite{mazurSuperPrimitiveSceneReconstruction2024}, and show how planar primitives can be exploited for effective pose refinement.

    \paragraph{Problem Formulation.}
      As illustrated in \cref{fig:per-primitive-depth-alignment}, the core idea of primitive-based pose refinement is to estimate a transformation $\refinedpose$ that refines the initial pose $\querypose_0$ towards a more accurate pose $\querypose^*$,
      while jointly optimizing the noisy plane parameters $\planeparameterquery_i$ for query primitives so that they better align with the depth map $\rendereddepth$ rendered at $\querypose_0$.
      More specifically, since the query normals $\{\planenormal^q_i\}$ are generally reliable, we keep them fixed during optimization.
      In contrast, the offsets $\{\planeoffset^q_i\}$ are more prone to errors hypothetically up to \emph{a-priori unknown scales}.
      We therefore introduce \emph{\offsetseedsname}$\,\{\offsetseed_i\}$ as optimization variables to compensate for this.

    \paragraph{Per-Primitive Depth Alignment.}
      First, given the camera intrinsics $\queryintrinsics$, the \emph{offset-seeded} depth segment of a query primitive $\planarprimitivequery_i$ is computed from its predicted plane parameters $\planeparameter^q_i$ and 2D segment $\planarprimitiveshape^q_i$, as %
      $\offsetseed_i\cdot\depthsegment_i(\planeparameter^q_i, \planarprimitiveshape^q_i; \queryintrinsics)$.
      Then, we warp $\offsetseed_i\depthsegment_i$ onto the depth rendering $\rendereddepth$ as follows:
      \begin{equation}
        \hat{\planarprimitivequery_i}[\querypixel], \hat{\rendereddepth}_i[\querypixel] = \projection\left(\refinedpose\,\unprojection(\querypixel, \offsetseed_i\depthsegment_i)\right),
        \label{equ:primitive-warping}
      \end{equation}
      where the pixel $\querypixel\in\planarprimitiveshape^q_i$ with offset-seeded depth value $\offsetseed_i\depthsegment_i[\querypixel]$ is unprojected by $\unprojection(\cdot)$, transformed by $\refinedpose$, and subsequently projected onto $\rendereddepth$ via $\projection(\cdot)$.
      Next, the depth residual is defined as the difference between the projection depth $\hat{\rendereddepth}_i[\querypixel]$ and the rendered depth at the warped pixel location $\hat{\planarprimitivequery_i}[\querypixel]$.
      Averaging residuals over all pixels $\querypixel\in\planarprimitiveshape^q_i$ yields the per-primitive depth alignment error:
      \begin{equation}
        \perprimitiveresidual(\offsetseed_i, \refinedpose; \planarprimitivequery_i, \rendereddepth) = \frac{1}{|\planarprimitiveshape^q_i|}\sum_{\querypixel\in\planarprimitiveshape^q_i}(
        \rendereddepth[\hat{\planarprimitivequery_i}[\querypixel]] - \hat{\rendereddepth}_i[\querypixel]
        )^2.
        \label{equ:per-primitive-depth-alignment-error}
      \end{equation}

      We aggregate the per-primitive depth alignment errors across all offset-seeded $\planarprimitivequery\in\querymap$, and minimize the resulting depth cost $\depthcost$ via gradient descent to jointly refine the camera pose $\querypose^*=\refinedpose^*\times\querypose_0$ and the \offsetseedsname $\{\offsetseed_i^*\}$:
      \begin{equation}
        \refinedpose^*, \{\offsetseed_i^*\} = \argmin_{\refinedpose, \{\offsetseed_i\}}\underbrace{\frac{1}{\queryplanenum}\sum_{(\planarprimitivequery_i, \offsetseed_i)} \perprimitiveresidual(\offsetseed_i, \refinedpose; \planarprimitivequery_i, \rendereddepth)}_{\depthcost}.
        \label{equ:depth-alignment-cost}
      \end{equation}
      \vspace{-2.4em} %

\section{Experiments}
  \label{sec:experiments}

  \paragraph{Datasets.}
    For our task, we curated a dataset from \scannet~\cite{daiScanNetRichlyannotated3D2017} following the split in~\cite{tanNOPESACNeuralOneplane2023}.
    Building on the scripts provided by~\cite{liuPlaneRCNN3DPlane2019,tanNOPESACNeuralOneplane2023}, we extended annotations for each query-map pair with ground-truth camera pose and \matchingtwothreeD plane matches.
    The resulting dataset contains \num{45802}\slash \num{7735} query-map pairs from \num{1210}\slash \num{303} scenes for training\slash testing.
    We also prepared \num{1023} pairs from the \twelvescenes~\cite{valentinLearningNavigateEnergy2016} dataset for out-of-the-box evaluation.
    Similarly, maps in this dataset are generated by sequentially fitting planes to the provided dense reconstructions and are further simplified for comparable compactness to that of \scannet.

  \paragraph{Baselines.}
    We adapt several existing systems as baselines to compare against our plane-centric method in achieving lean camera relocalization with no visual cues or pose priors:
    \begin{itemize}
      \item Oracle coarse initialization~\enparen{\AbbrCoarseInit}: an initial pose is coarsely estimated via heuristic rules based on the plane parameters of \num{20} map primitives, comprising all ground-truth matches and primitives nearest to the ground truth pose.
            The heuristic initialization rules ensure an average visual overlap of over \qty{30}{\percent} \wrt the ground truth poses.
      \item Image-to-point cloud registration~\enparen{\AbbrImgToPC}: the map is uniformly sampled into points at a resolution of \qty{2.5}{\centi\meter}, and the pose is estimated via either
            \enparen{1} \emph{\geotransformer}~\cite{qinGeoTransformerFastRobust2023}, which establishes 3D\textendash 3D point correspondences between the map and the metric-scale geometry of the query image recovered by~\cite{wangMoGe2AccurateMonocular2025},
            or
            \enparen{2} \emph{\freereg}~\cite{wangFreeRegImagetoPointCloud2024}, which directly establishes pixel\textendash point~\enparen{\matchingtwothreeD} correspondences.
      \item MeshLoc~\cite{panekMeshLocMeshbasedVisual2022,panekVisualLocalizationUsing2023}: we employ diverse keypoint extractors and matchers to establish pixel\textendash pixel correspondences between the query and the synthetic rendering from the map given the coarse initialization, and then lift them to pixel\textendash point correspondences for pose estimation.
    \end{itemize}

  \paragraph{Implementation Details.}
    Our method is implemented on top of the Detectron2 framework~\cite{wuDetectron22019}.
    We instantiate the plane recovery module by combining \mogevtwo~\cite{wangMoGe2AccurateMonocular2025} for monocular geometry estimation
    and
    an efficient sequential \seqRANSAC implementation by~\cite{watsonAirPlanesAccuratePlane2024} for plane fitting.
    To reduce cost, a lightweight CNN-based upsampler is employed to neatly fuse multi-scale features from the ViT~\cite{dosovitskiyImageWorth16x162021} encoder of \mogevtwo into a feature map of size $H/$\num{8} $\times$ $W/$\num{8}.
    Both the object and scene encoders for 3D embeddings are instantiated with PointNet~\cite{qiPointNetDeepLearning2017}.
    The dimensionality $c$ of 2D\slash 3D embeddings is set to \num{384}.
    The matching module consists of $N$=\,\num{4} layers, and each attention unit has \num{4} heads.
    When pose estimation degenerates due to insufficient inliers, we apply the same heuristic strategy as \AbbrCoarseInit to obtain a final output from the predicted correspondences.

    \begin{table*}[htbp]
      \centering
      \caption{
        \textbf{Matching evaluation} with IoU score\,$\geq$\,\num{0.3}.
        Point correspondences are first lifted to plane matches via majority voting.
        \TMatchTP~and \TMatchTotal~denote the total number of true positives and ground-truth correspondences, respectively.
        \TcoloredFeature{} indicates reliance on visual appearance.
      }
      \setlength{\tabcolsep}{6pt} %
      \resizebox{\textwidth}{!}{%
        
\begin{tabular}{llcccccccccccc}
  \Xhline{3\arrayrulewidth}                                                                                                                                                                                                                                                                                                                                                                                                                                                        \\[-0.9em]
                                                                   & \multirow{2}{*}{\rotatebox[origin=c]{0}{\makecell[t]{\TFeatureType}}} & \multicolumn{6}{c}{\TMatchScanNet} & \multicolumn{6}{c}{\TMatchTwelveScenes}                                                                                                                                                                                                                                                          \\[-0.05em]
  \cmidrule(r){3-8} \cmidrule(r){9-14}                                                                                                                                                                                                                                                                                                                                                                                                                                             \\[-1.3em]
  \TMethods                                                        &                                                                       & \TMatchPrecision                   & \TMatchRecall                           & \TMatchFScore                 & \TMatchAP                     & \TMatchTP   & \TMatchTotal & \TMatchPrecision              & \TMatchRecall                 & \TMatchFScore                 & \TMatchAP                     & \TMatchTP  & \TMatchTotal \\[-0.05em]
  \Xhline{.3\arrayrulewidth}                                                                                                                                                                                                                                                                                                                                                                                                                                                       \\[-0.9em]
  \TgeotransformerT~\cite{qinGeoTransformerFastRobust2023}         & \TFPoint                                                              & \tablenum{30.8}                    & \tablenum{22.8}                         & \tablenum{26.2}               & \tablenum{38.5}               & \num{13026} & \num{57253}  & \tablenum{20.3}               & \tablenum{16.8}               & \tablenum{18.4}               & \tablenum{28.0}               & \num{1565} & \num{9319}   \\[0.05em]
  \Tfreereg~\cite{wangFreeRegImagetoPointCloud2024}                & \TFPoint                                                              & \tablenum{21.7}                    & \tablenum{19.2}                         & \tablenum{20.4}               & \tablenum{34.1}               & \num{7837}  & \num{40857}  & \tablenum{20.2}               & \tablenum{14.1}               & \tablenum{16.6}               & \tablenum{21.5}               & \num{1077} & \num{7647}   \\[0.05em]
  \Xhline{.3\arrayrulewidth}                                                                                                                                                                                                                                                                                                                                                                                                                                                       \\[-0.9em]
  \Tmastr~\cite{leroyGroundingImageMatching2024}                   & \TcoloredFeature{\TFPoint}                                            & \TRankSecond{\tablenum{61.7}}      & \TRankThird{\tablenum{45.0}}            & \TRankSecond{\tablenum{52.0}} & \TRankSecond{\tablenum{84.1}} & \num{18372} & \num{40857}  & \TRankSecond{\tablenum{59.8}} & \TRankThird{\tablenum{42.9}}  & \TRankThird{\tablenum{50.0}}  & \TRankSecond{\tablenum{81.6}} & \num{3283} & \num{7647}   \\[0.05em]
  \Tmatchanything~\cite{heMatchAnythingUniversalCrossModality2025} & \TFPoint                                                              & \tablenum{42.1}                    & \TRankSecond{\tablenum{48.2}}           & \TRankThird{\tablenum{45.0}}  & \tablenum{67.7}               & \num{19698} & \num{40857}  & \TRankThird{\tablenum{51.2}}  & \TRankFirst{\tablenum{56.1}}  & \TRankSecond{\tablenum{53.5}} & \TRankThird{\tablenum{77.2}}  & \num{4289} & \num{7647}   \\[0.05em]
  \Tnopesac~\cite{tanNOPESACNeuralOneplane2023}                    & \TcoloredFeature{\TFPlane}                                            & \TRankThird{\tablenum{51.4}}       & \tablenum{35.4}                         & \tablenum{41.9}               & \TRankThird{\tablenum{79.1}}  & \num{14462} & \num{40857}  & \tablenum{43.8}               & \tablenum{22.0}               & \tablenum{29.3}               & \tablenum{71.4}               & \num{1684} & \num{7647}   \\[0.05em]
  \Xhline{.3\arrayrulewidth}\noalign{\vskip 3pt}\Xhline{.3\arrayrulewidth}                                                                                                                                                                                                                                                                                                                                                                                                         \\[-0.9em]
  \Tours                                                           & \TFPlane                                                              & \TRankFirst{\tablenum{67.6}}       & \TRankFirst{\tablenum{61.3}}            & \TRankFirst{\tablenum{64.3}}  & \TRankFirst{\tablenum{91.8}}  & \num{36893} & \num{60191}  & \TRankFirst{\tablenum{63.9}}  & \TRankSecond{\tablenum{54.2}} & \TRankFirst{\tablenum{58.6}}  & \TRankFirst{\tablenum{87.8}}  & \num{5184} & \num{9572}   \\[0.05em]
  \Xhline{3\arrayrulewidth}                                                                                                                                                                                                                                                                                                                                                                                                                                                        \\[-0.9em]
\end{tabular}

      }
      \label{tab:matching-performance}
      \vspace{-.75em}
    \end{table*}

    \begin{table}[htbp]
      \centering
      \caption{
        \textbf{Relocalization results on \twelvescenes.}
      }
      \setlength{\tabcolsep}{4pt} %
      \resizebox{\linewidth}{!}{%
        
\begin{tabular}{lccccc}
  \Xhline{3\arrayrulewidth}                                                                                                                                                                                                                                                \\[-1.0em]
                                                                                                  & \multicolumn{2}{c}{\TAblationPoseErr} & \multicolumn{3}{c}{\TAccThres}                                                                                                 \\[-0.07em]
  \cmidrule(r){2-3} \cmidrule(r){4-6}                                                                                                                                                                                                                                      \\[-1.32em]
  \TMethods                                                                                       & \TAblationRotErr                      & \TAblationPosErr               & \TAccFine                     & \TAccMed                      & \TAccCoarse                   \\[-0.07em]
  \Xhline{.3\arrayrulewidth}                                                                                                                                                                                                                                               \\[-1.0em]
  \rowcolor{\TNoRankColor}\Tcoarseinit                                                            & \tablenum{22.5}                       & \tablenum{0.47}                & \tablenum{0.5}                & \tablenum{18.0}               & \tablenum{69.8}               \\[0.03em]
  \Xhline{.3\arrayrulewidth}                                                                                                                                                                                                                                               \\[-1.0em]
  \TAbbrgeotransformerT~\cite{qinGeoTransformerFastRobust2023}                                    & \tablenum{33.2}                       & \tablenum{0.80}                & \tablenum{22.2}               & \tablenum{37.8}               & \tablenum{43.5}               \\[0.03em]
  \Tfreereg~\cite{wangFreeRegImagetoPointCloud2024}                                               & \tablenum{23.7}                       & \tablenum{0.49}                & \tablenum{1.7}                & \tablenum{17.5}               & \tablenum{64.2}               \\[0.03em]
  \Xhline{.3\arrayrulewidth}                                                                                                                                                                                                                                               \\[-1.0em]
  \Tsplg~\cite{detoneSuperPointSelfsupervisedInterest2018a,lindenbergerLightGlueLocalFeature2023} & \tablenum{43.9}                       & \tablenum{0.96}                & \tablenum{10.4}               & \tablenum{17.4}               & \tablenum{34.0}               \\[0.03em]
  \Tloftr~\cite{sunLoFTRDetectorFreeLocal2021}                                                    & \tablenum{31.4}                       & \tablenum{0.62}                & \tablenum{31.9}               & \tablenum{40.2}               & \tablenum{48.8}               \\[0.03em]
  \Tmastr~\cite{leroyGroundingImageMatching2024}                                                  & \tablenum{12.0}                       & \tablenum{0.30}                & \TRankThird{\tablenum{45.2}}  & \tablenum{51.9}               & \tablenum{59.0}               \\[0.03em]
  \TAbbrmatchanything~\cite{heMatchAnythingUniversalCrossModality2025}                            & \TRankThird{\tablenum{7.9}}           & \TRankSecond{\tablenum{0.20}}  & \TRankSecond{\tablenum{46.9}} & \TRankThird{\tablenum{63.4}}  & \TRankThird{\tablenum{77.6}}  \\[0.03em]
  \Tnopesac~\cite{tanNOPESACNeuralOneplane2023}                                                   & \tablenum{17.7}                       & \tablenum{0.54}                & \tablenum{2.9}                & \tablenum{25.9}               & \tablenum{67.2}               \\[0.03em]
  \Xhline{.3\arrayrulewidth}\noalign{\vskip 3pt}\Xhline{.3\arrayrulewidth}                                                                                                                                                                                                 \\[-1.0em]
  \TAbbroursnorefine                                                                              & \TRankSecond{\tablenum{4.8}}          & \TRankThird{\tablenum{0.28}}   & \tablenum{34.9}               & \TRankSecond{\tablenum{66.7}} & \TRankSecond{\tablenum{79.9}} \\[0.03em]
  \TAbbroursfull                                                                                  & \TRankFirst{\tablenum{4.7}}           & \TRankFirst{\tablenum{0.19}}   & \TRankFirst{\tablenum{50.6}}  & \TRankFirst{\tablenum{70.8}}  & \TRankFirst{\tablenum{80.6}}  \\[0.03em]
  \Xhline{3\arrayrulewidth}                                                                                                                                                                                                                                                \\[-1.0em]
\end{tabular}

      }
      \label{tab:12scenes-pose-results}
      \vspace{-1.3em}
    \end{table}

  \subsection{Relocalization Accuracy}
    As shown in \cref{tab:scannetv2-pose-results}, we first present the overall camera relocalization performance of all methods on \scannet.
    We report the mean\slash median rotation and translation errors, along with pose recalls across three thresholds, following~\cite{tanNOPESACNeuralOneplane2023,shiPlaneRecTRUnifiedQuery2025}.
    To obtain more meaningful results for our baselines, we apply map truncation~\enparen{\AbbrTruncatedMap}, either by restricting the map to a subset of plane primitives~\enparen{\AbbrCoarseInit} or by cropping structures far from the ground-truth pose~\enparen{\emph{\TgeotransformerT} and \emph{\ffreereg}}.
    For the MeshLoc series, \AbbrCoarseInit~is required to provide an initial pose, and most methods further rely on map appearance for colored renderings during matching.

    Setting aside the \postrefine module introduced in \cref{sec:pose-refinement}, \cref{tab:scannetv2-pose-results} shows that \methodname is the only method that \enparen{1} achieves top performance across all evaluation metrics \enparen{2} while not relying on any pose priors or map appearance.
    Notably, \postrefine further improves accuracy with affordable runtime overhead.
    In the cross-dataset experiments on \twelvescenes, as reported in \cref{tab:12scenes-pose-results}, several methods perform reasonably well, due to the more reliable pose initialization and higher map rendering fidelity.
    Meanwhile, \methodname remains competitive, given its complete independence from any auxiliary inputs.
    The consistent performance advantage of our method across datasets suggests its effectiveness and highlights the potential benefits of exploiting planar primitives for camera relocalization.

  \subsection{Matching Performance}
    Next, we \change{probe}{analyze} the matching \change{capabilities}{performance} of various approaches %
    using Precision, Recall, F-score and Average Precision~\enparen{AP}.
    A predicted plane correspondence is counted as a true positive if the \emph{recovered} query primitive is matched to its ground-truth map primitive and the mask IoU of their 2D projections is $\geq$\,\num{0.3}.
    For point-based methods, point matches are first lifted to plane-level through majority voting, \ie, each \emph{ground-truth} query primitive is assigned to the map primitive where the majority of its point matches fall into.
    Then, we calculate the IoU score of such plane match as the ratio of the majority count to the total number of point matches within the union of their 2D projections.

    The results are presented in \cref{tab:matching-performance}.
    \AbbrImgToPC methods struggle to establish correct point correspondences across modalities, especially when 3D structures cover large areas and degenerate into piecewise planar representations.
    In contrast, without depending on visual appearance, \methodname achieves competitive or even superior cross-modal \matchingtwothreeD matching performance compared to MeshLoc variants that perform 2D\textendash 2D matching.
    This suggests the advantage of planar primitives, as a form of region-based representation, in supporting purely structure-based matching.

    \begin{figure}[htbp]
      \centering
      \begin{minipage}[b]{0.45\linewidth}
        \centering
        \setlength{\tabcolsep}{2pt} %
        \resizebox{\linewidth}{!}{%
          
\begin{tabular}{lcc}
  \Xhline{3\arrayrulewidth} \\[-0.9em]
  & \multirow{2}{*}{\rotatebox[origin=c]{0}{\makecell[t]{\TAblationMatchFScore}}} & \multirow{2}{*}{\rotatebox[origin=c]{0}{\makecell[t]{\TAblationPoseAccHigh}}}\\[0.04999999999999999em]
\\[-1.2000000000000002em]
 \TMethods & &\\[0.04999999999999999em]
\Xhline{.3\arrayrulewidth} \\[-0.9em]
\TAblationOurs&\TRankFirst{\tablenum{64.3}}&\TRankFirst{\tablenum{37.1}} \\[0.15em]
\Xhline{.3\arrayrulewidth} \\[-0.9em]
\TAblationNoScenewiseEncoding&\tablenum{44.0}&\tablenum{17.2} \\[0.15em]
\TAblationNoPiecewiseEncoding&\TRankThird{\tablenum{51.7}}&\tablenum{27.8} \\[0.15em]
\TAblationNoPositionalEncoding&\TRankSecond{\tablenum{60.1}}&\TRankThird{\tablenum{33.9}} \\[0.15em]
\Xhline{.3\arrayrulewidth} \\[-0.9em]
\TAblationNoSolverRANSAC&\TRankFirst{\tablenum{64.3}}&\tablenum{29.5} \\[0.15em]
\TAblationNoSolverScaleOpt&\TRankFirst{\tablenum{64.3}}&\TRankSecond{\tablenum{36.2}} \\[0.15em]
\Xhline{3\arrayrulewidth} \\[-0.9em]
\end{tabular}

        }
        \vspace{.2em}
        \captionof{table}{\textbf{Ablation study.}}
        \label{tab:ablating-components}
      \end{minipage}
      \hfill
      \begin{minipage}[b]{0.53\linewidth}
        \centering
        \includegraphics[width=1.02\textwidth]{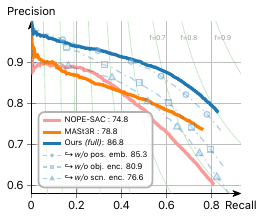}
        \captionof{figure}{\textbf{PR curves.}}
        \label{fig:ablating-pr-curve}
      \end{minipage}
      \vspace{-1em}
    \end{figure}

    \begin{figure*}[htbp]
      \begin{subfigure}[b]{0.33\textwidth}
        \centering
        \includegraphics{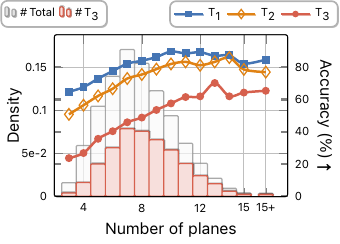}
        \caption{Pose recall.}
        \label{fig:dependence-on-plane-num-a}
      \end{subfigure}%
      \begin{subfigure}[b]{0.33\textwidth}
        \centering
        \includegraphics{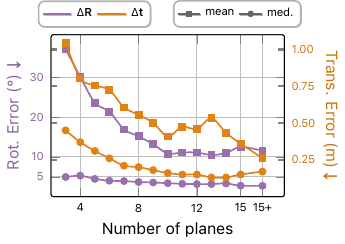}
        \caption{Pose error.}
        \label{fig:dependence-on-plane-num-b}
      \end{subfigure}%
      \begin{subfigure}[b]{0.33\textwidth}
        \centering
        \includegraphics{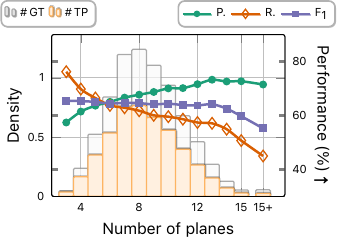}
        \caption{Matching performance.}
        \label{fig:dependence-on-plane-num-c}
      \end{subfigure}%
      \caption{
        \textbf{Impact of plane richness.}
        Results with \postrefine.
        Three thresholds from coarse to fine are referred to as: \FAbbAccCoarse, \FAbbAccMed, and \FAbbAccHigh.
      }
      \label{fig:dependence-on-plane-num}
    \end{figure*}

    \begin{figure*}
      \centering
      \begin{tikzpicture}
        \node[anchor=south west,inner sep=0] (image) at (0,0) {
          \includegraphics{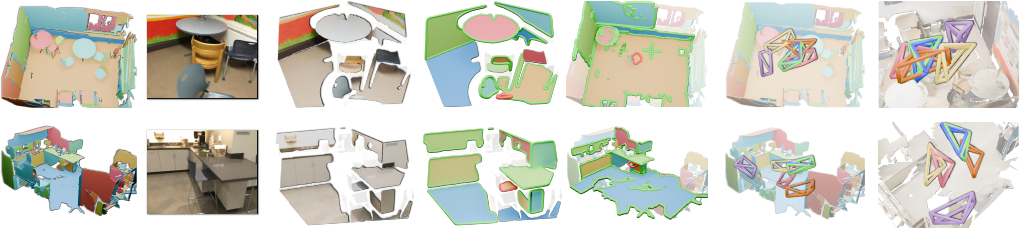}
        };
        {
        \def\subtitley{-0.10}
        \begin{scope}[x={(image.south east)},y={(image.north west)}]
          \node[fill=white, fill opacity=0.7, text opacity=1, inner sep=2pt] at (0.128, \subtitley) {\footnotesize \captiona Input: map \& query};
          \node[fill=white, fill opacity=0.7, text opacity=1, inner sep=2pt] at (0.34, \subtitley) {\footnotesize \captionb Monocular plane recovery};
          \node[fill=white, fill opacity=0.7, text opacity=1, inner sep=2pt] at (0.562, \subtitley) {\footnotesize \captionc Plane correspondences};
          \node[fill=white, fill opacity=0.7, text opacity=1, inner sep=2pt] at (0.861, \subtitley) {\footnotesize \captiond Poses (viewpoint 1 \& 2)};
        \end{scope}
        }
      \end{tikzpicture}
      \caption{
        \textbf{Qualitative examples.}
        \captionc Plane correspondences are color-coded, with true positives outlined in green and false ones in red.
        \captiond Camera poses relocalized by different methods are compared from two viewpoints.
        Legend:
        \qelegend{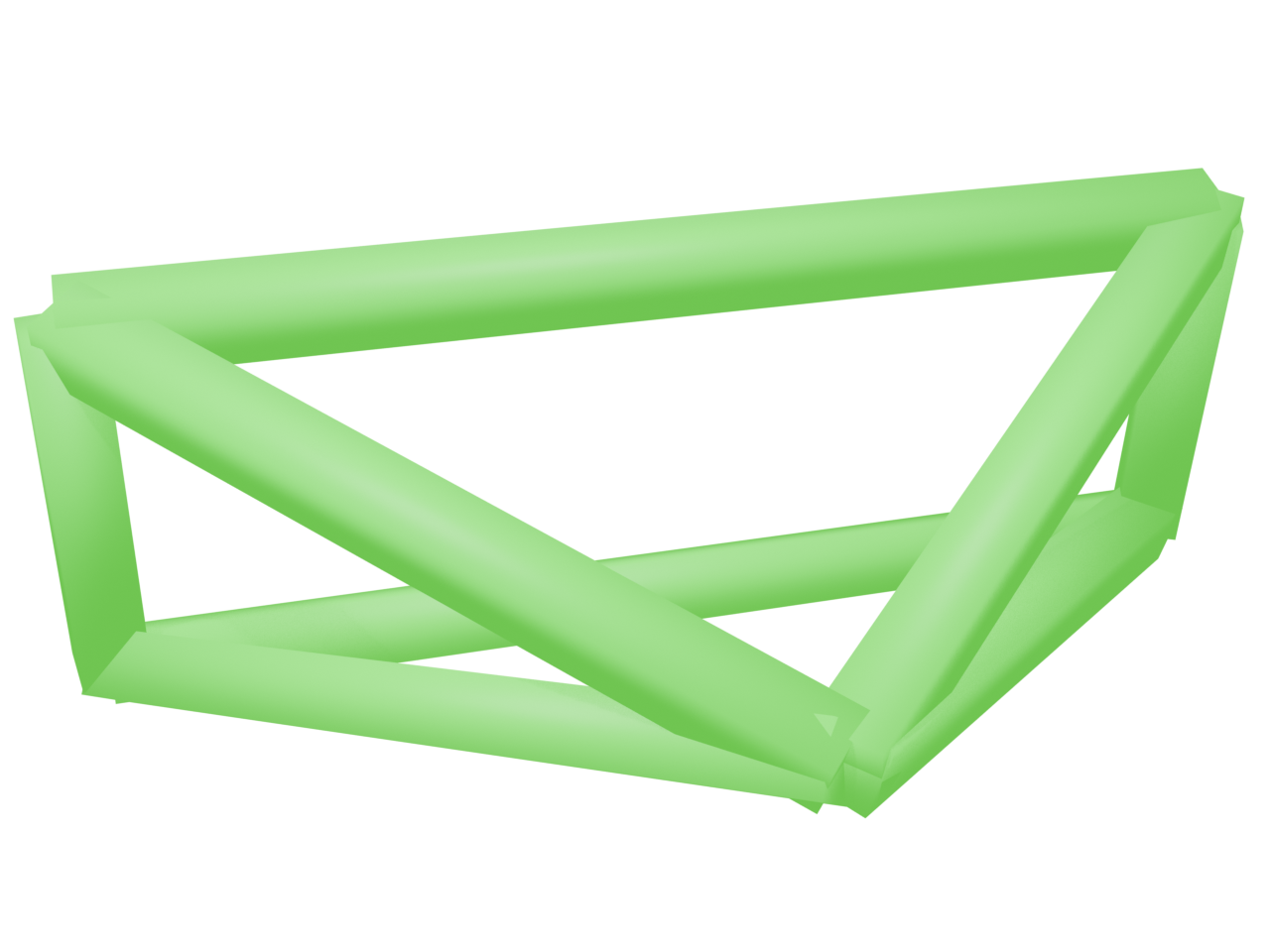}\textcolor{tikzColorMethodGT}{\gtpose},
        \qelegend{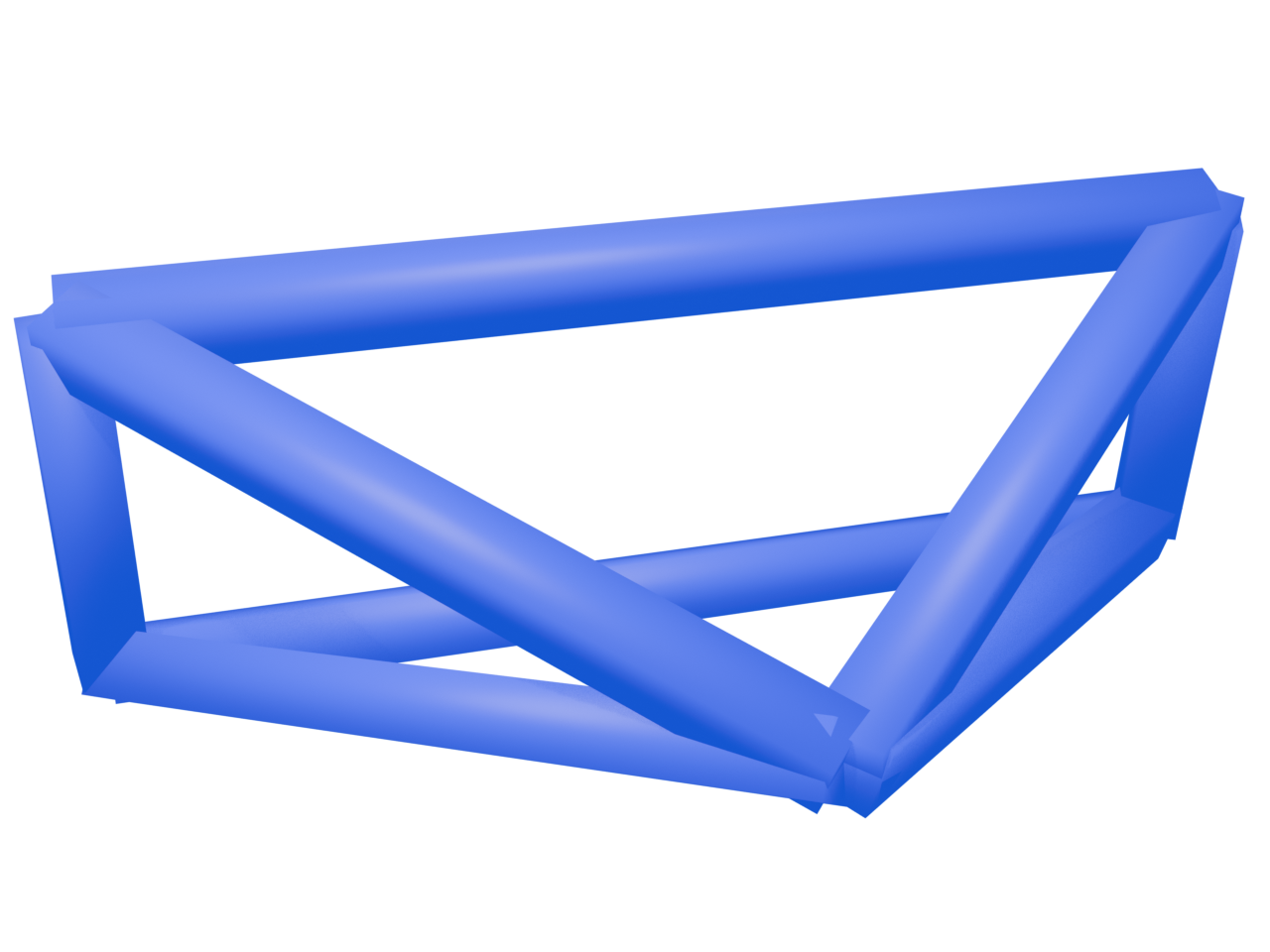}\textcolor{tikzColorMethodOurs}{\methodname\enparen{Ours}},
        \qelegend{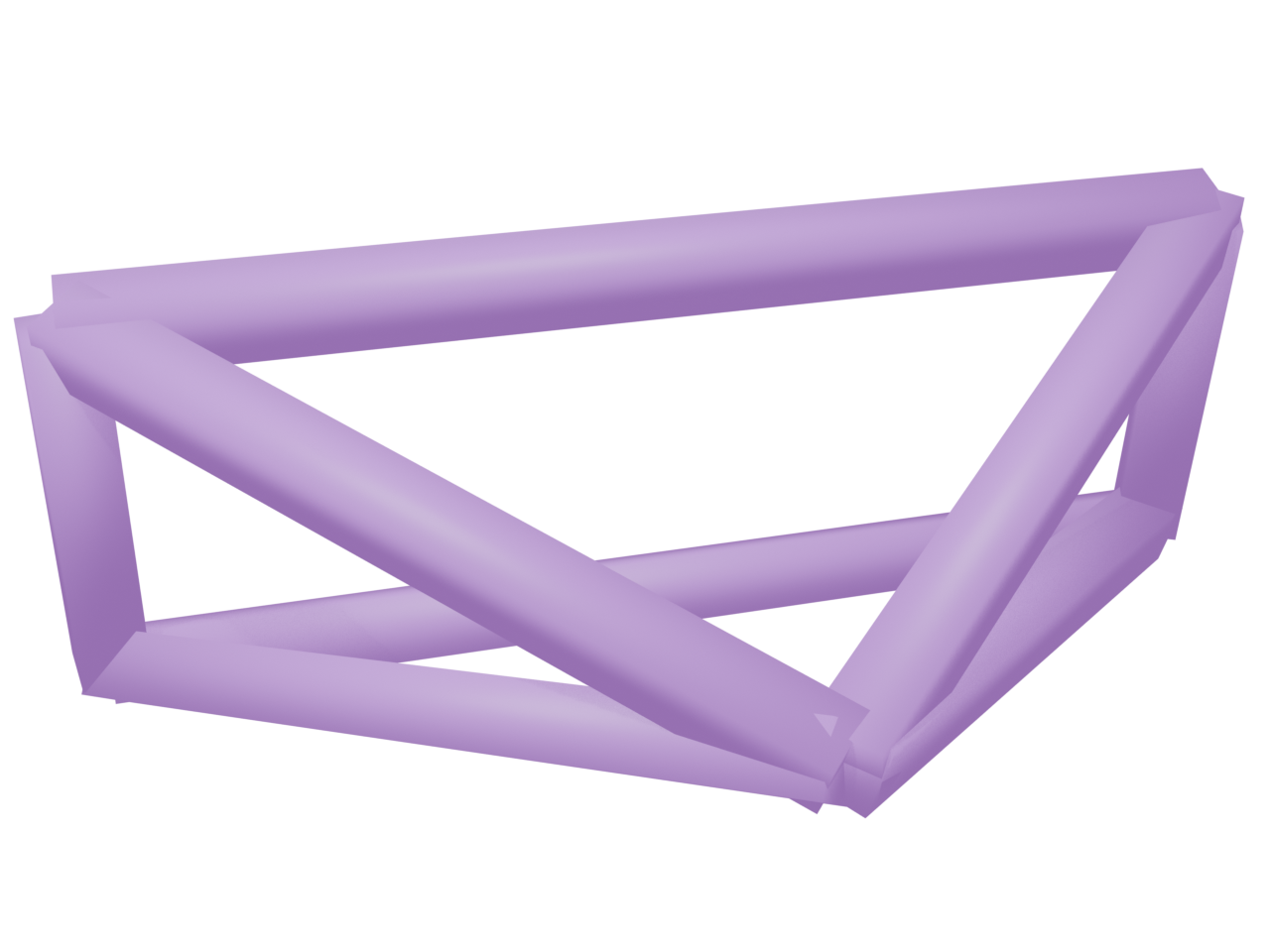}\textcolor{tikzColorMethodGeoTr!
          60!black}{\geotransformer-T},
        \qelegend{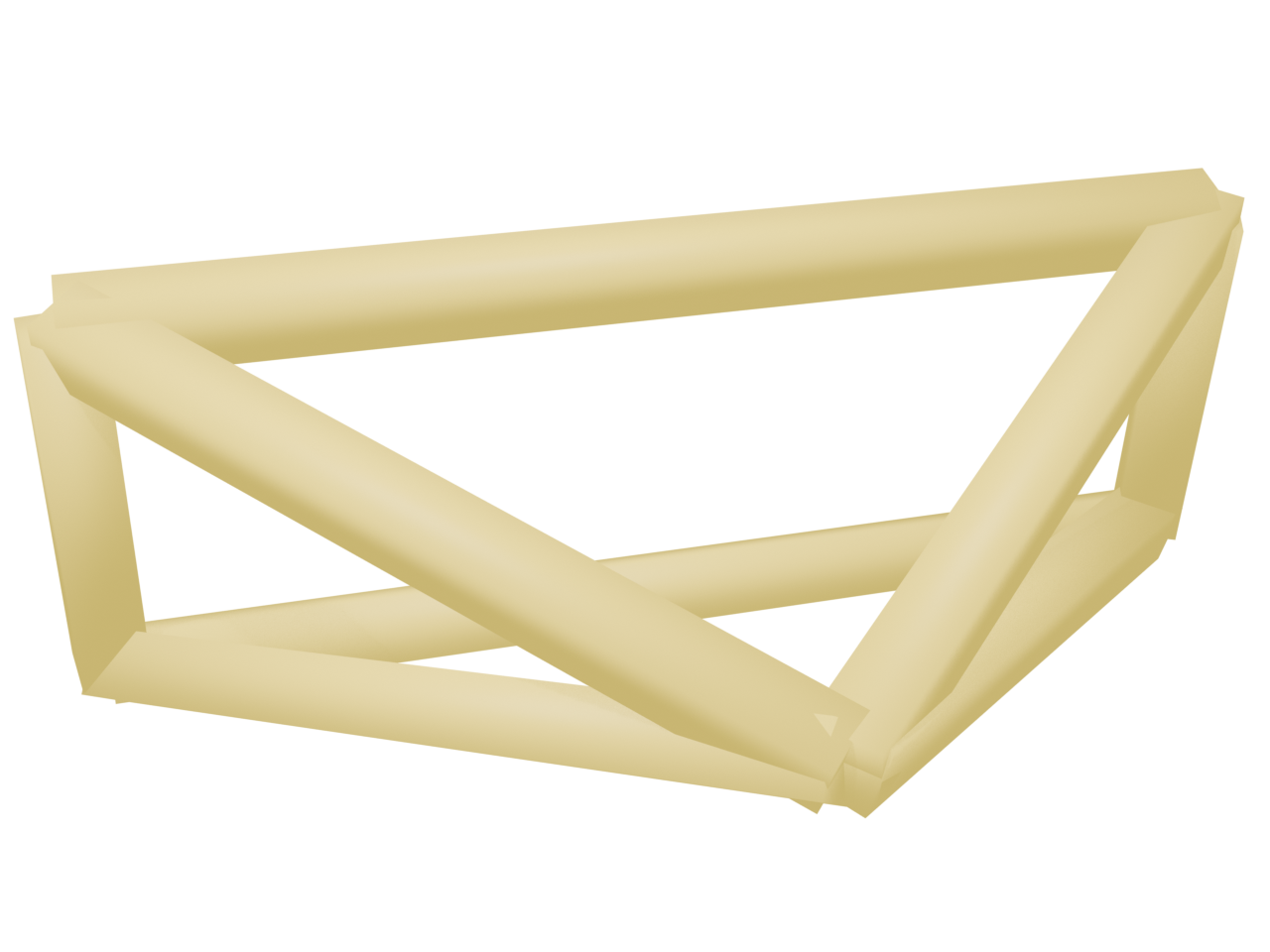}\textcolor{tikzColorMethodCoarse}{\AbbrCoarseInit},
        \qelegend{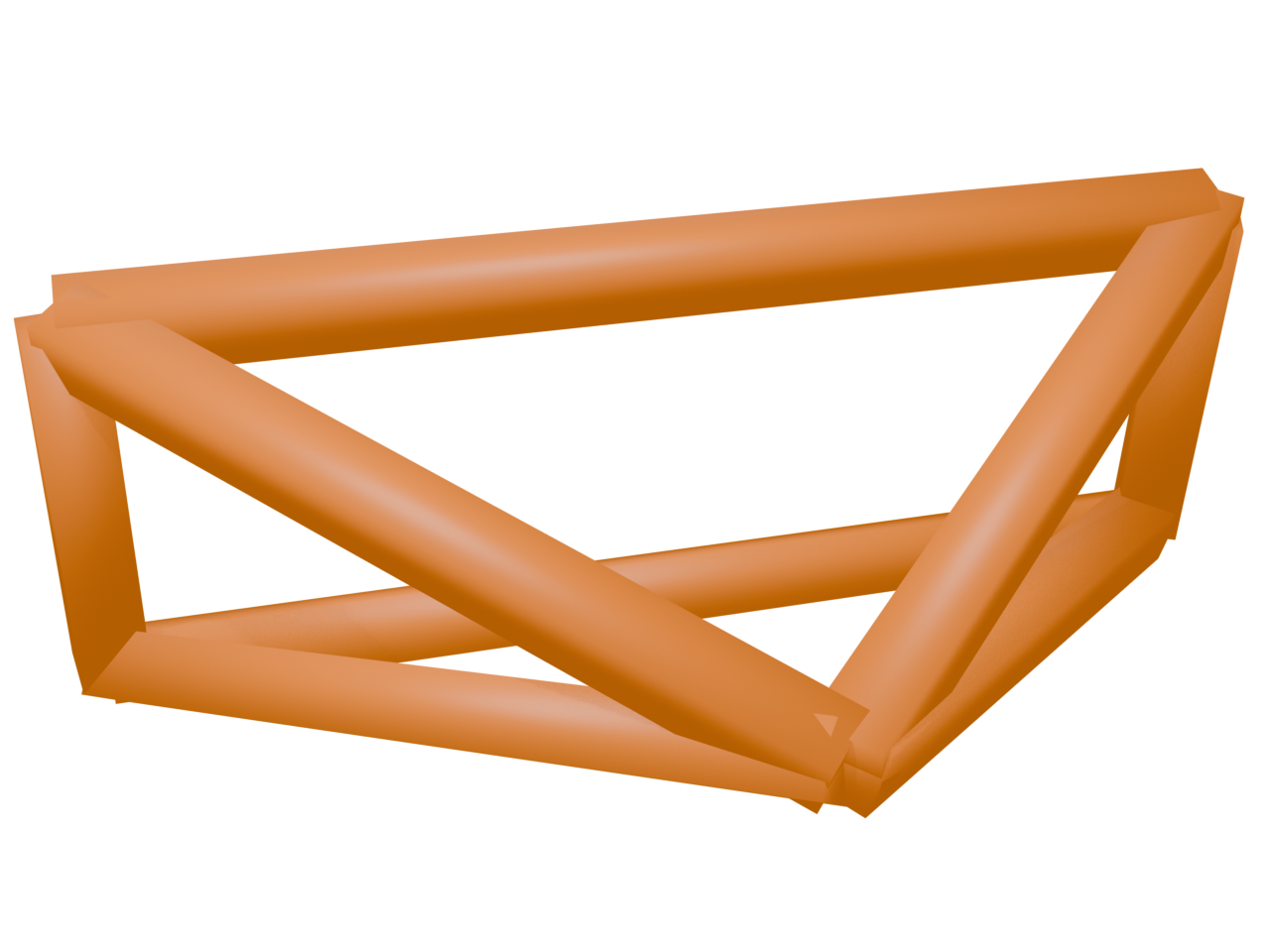}\textcolor{tikzColorMethodMast3r!80!black}{\mastr},
        \qelegend{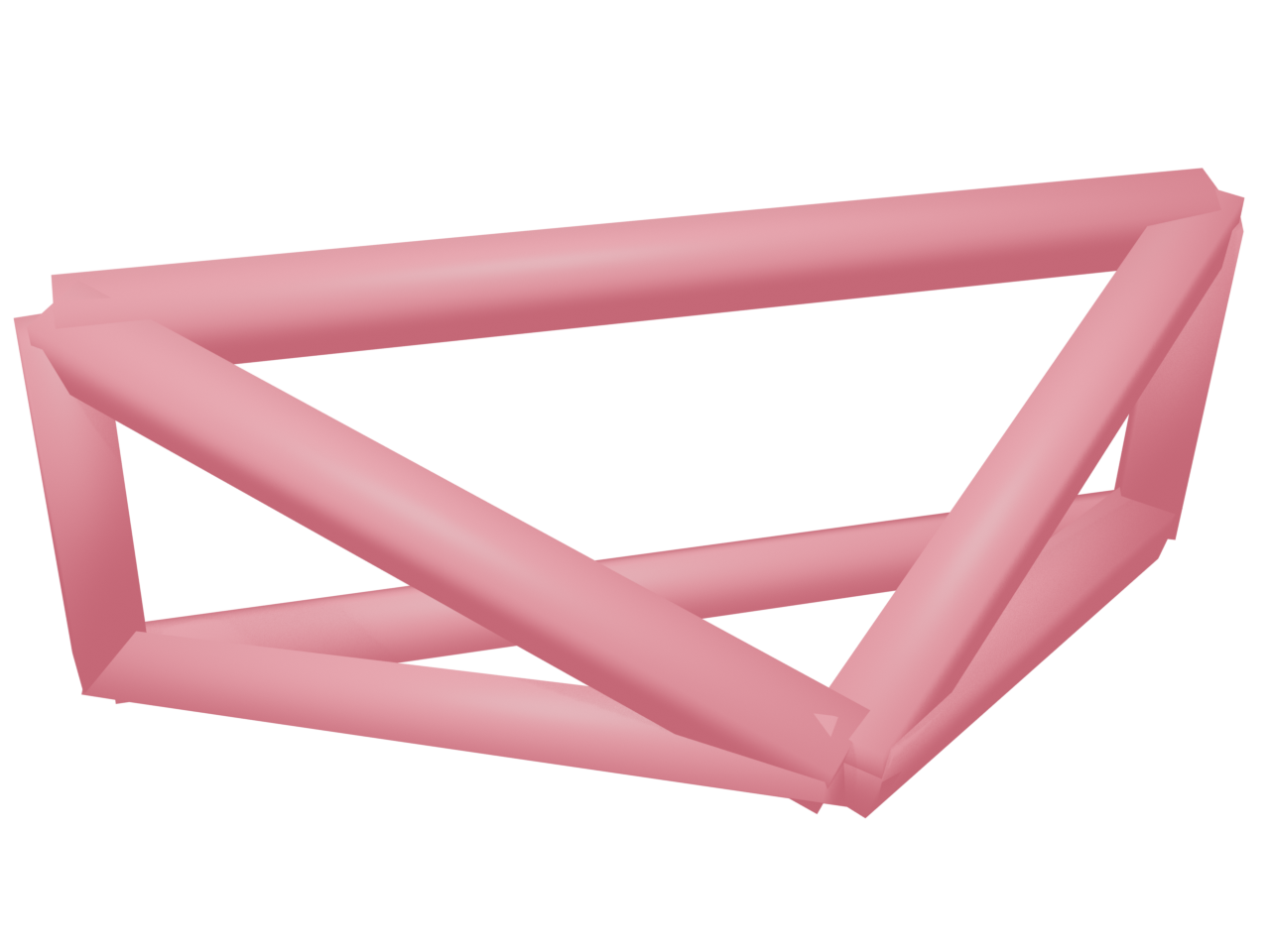}\textcolor{tikzColorMethodNOPESAC!70!black}{\nopesac}.
        See the appendix for additional visualizations on both datasets.
      }
      \label{fig:qualitative-examples}
    \end{figure*}

  \subsection{Analysis}
    \paragraph{Ablating Components.}
      We conduct ablations on \scannet, with results in \cref{tab:ablating-components}.
      Both the scene and object encoders significantly contribute to the map primitive embeddings.
      Meanwhile, the absence of positional embedding incurs a noticeable drop in matching performance.
      \Cref{fig:ablating-pr-curve} displays the matching Precision\textendash Recall~\enparen{PR} curves \wrt IoU$\geq$\num{0.5} and lists the average precision in the legend.
      Estimating poses robustly by filtering outliers with RANSAC is crucial for accuracy, while the joint optimization of the metric scale further improves the results.
      Moreover, as a plug-and-play module, the monocular plane recovery front-end can be instantiated with alternatives, with results detailed in \cref{tab:ablation-2d-detectors}.

      \begin{table}[t]
        \centering
        \caption{
          \textbf{Results~\enparen{\wo \AbbrPoseRefine} with plane recovery alternatives on \scannet.}
          \emph{MoGe-2+RANSAC} is used in our default pipeline.
        }
        \setlength{\tabcolsep}{6pt} %
        \resizebox{\linewidth}{!}{%
          
\begin{tabular}{lcccc}
  \Xhline{3\arrayrulewidth}                                                                                                                                                                                                                                                                                                                              \\[-0.9em]
                                                                     & \multirow{2}{*}{\rotatebox[origin=c]{0}{\makecell[t]{\TAblationMatchFScorePct}}} & \multicolumn{2}{c}{\TAblationPoseErr} & \multirow{2}{*}{\rotatebox[origin=c]{0}{\makecell[t]{\TTimeMilliSecond}}}                                                                              \\[-0.05em]
  \cmidrule(r){3-4}                                                                                                                                                                                                                                                                                                                                      \\[-1.3em]
  \TMethods                                                          &                                                                                  & \TAblationRotErr                      & \TAblationPosErr                                                          &                                                                            \\[-0.05em]
  \Xhline{.3\arrayrulewidth}                                                                                                                                                                                                                                                                                                                             \\[-0.9em]
  \TAblationPlaneTR~\cite{tanPlaneTRStructureGuidedTransformers2021} & \TRankSecond{\tablenum{65.8}}                                                    & \tablenum{6.2}                        & \tablenum{0.42}                                                           & \Tsim\tablenum[round-mode = places, round-precision=1]{52.659}             \\[0.05em]
  \TAblationPlaneRecTR~\cite{shiPlaneRecTRUnifiedQuery2023}          & \tablenum{63.6}                                                                  & \tablenum{4.9}                        & \TRankThird{\tablenum{0.31}}                                              & \Tsim\tablenum[round-mode = places, round-precision=1]{58.033}             \\[0.05em]
  \TAblationZeroPlane~\cite{liuInthewild3DPlane2025}                 & \TRankFirst{\tablenum{66.0}}                                                     & \TRankThird{\tablenum{4.0}}           & \tablenum{0.41}                                                           & \Tsim\tablenum[round-mode = places, round-precision=1]{293.71000000000004} \\[0.05em]
  \TAblationPLANATHREER~\cite{liuPLANA3RZeroshotMetric2025}          & \tablenum{61.4}                                                                  & \TRankFirst{\tablenum{3.7}}           & \TRankSecond{\tablenum{0.28}}                                             & \Tsim\tablenum[round-mode = places, round-precision=1]{2781.82}            \\[0.05em]
  \TAblationOursDetector                                             & \TRankThird{\tablenum{64.3}}                                                     & \TRankSecond{\tablenum{3.9}}          & \TRankFirst{\tablenum{0.27}}                                              & \Tsim\tablenum[round-mode = places, round-precision=1]{59.854}             \\[0.05em]
  \Xhline{.3\arrayrulewidth}                                                                                                                                                                                                                                                                                                                             \\[-0.9em]
  \rowcolor{\TNoRankColor}\TAblationGTDepthSeqRANSAC                 & \tablenum{77.1}                                                                  & \tablenum{0.3}                        & \tablenum{0.03}                                                           & \Tsim\tablenum[round-mode = places, round-precision=1]{48.624}             \\[0.05em]
  \rowcolor{\TNoRankColor}\TAblationGTDepthGTMask                    & \tablenum{88.6}                                                                  & \tablenum{0.0}                        & \tablenum{0.00}                                                           & \Tsim\tablenum[round-mode = places, round-precision=1]{42.803}             \\[0.05em]
  \Xhline{3\arrayrulewidth}                                                                                                                                                                                                                                                                                                                              \\[-0.9em]
\end{tabular}

        }
        \label{tab:ablation-2d-detectors}
        \vspace{-1.5em}
      \end{table}

    \paragraph{Impact of Plane Richness.}
      Intuitively, informative observations facilitate relocalization.
      Here, we investigate how the richness\slash diversity of observed planar primitives affects relocalization performance.
      For simplicity, we approximate the plane richness of each query image by the number of its annotated plane segments.
      Then, we bin the test queries from \scannet into groups according to their ground-truth plane counts.
      Next, as plotted in \cref{fig:dependence-on-plane-num}, we analyze \captiona pose recalls, \captionb pose errors, and \captionc matching performance within each group.
      The results indicate that richer plane observations generally lead to improved relocalization, especially when the plane count is below \num{12}.
      However, as plane richness continues to increase, the performance gain becomes negligible, presumably due to the degraded plane recovery, as planes in richer observations tend to be less salient and harder to recover and match accurately.
      Such attribution is further supported by the drop in matching recall at the tail of the curve~\enparen{see \recalllegend in \cref{fig:dependence-on-plane-num-c}}.

    \paragraph{Qualitative Examples.}
      \Cref{fig:qualitative-examples} presents two cases from \scannet, including intermediate outputs and final poses estimated by our method and compared baselines.

\section{Conclusion}
  \label{sec:conclusion}

  In this paper, we have introduced \emph{\methodname}, a light-weight alternative for camera relocalization that exploits planar primitives in structured environments.
  We adhered to two core principles while designing our method: (1) a simple yet effective matching network, coupled with a plug-and-play monocular plane recovery module that excavates structural cues from query images; (2) a robust pose estimation framework with \postrefine to filter out matching outliers and mitigate imperfections in the monocular front-end.
  This streamlined paradigm supports extensive evaluation across over hundreds of structured indoor scenes, which clearly highlights the strong potential of planar primitives for cross-modal structural associations and pose estimation in the task of 6-DoF camera relocalization.

\section{Acknowledgements}

  This work was supported in part by the Beijing Natural Science Foundation (No.
  L223003), the National Natural Science Foundation of China (No.
  U22B2055, 62273345, 62402495) and the Key R\&D Project in Henan Province (No.
  231111210300).

\appendix

\section*{Appendix}

  The appendix further provides the following supplementary materials in support of the main paper:
  \begin{itemize}
    \item details on dataset preparation~\enparen{\Cref{sec:appendix-dataset-preparation}};
    \item auxiliary settings for baseline methods, including map truncation and the oracle coarse initialization~\enparen{\Cref{sec:appendix-baselines}};
    \item comprehensive implementation details for \methodname, including the network architecture, training scheme, and the pose estimation and post refining process~\enparen{\Cref{sec:appendix-implementation}};
    \item additional analysis and visualizations~\enparen{\Cref{sec:additional-experimental-results}};
    \item discussion of limitations and future work~\enparen{\Cref{sec:appendix-limitations}}.
  \end{itemize}

\section{Dataset Preparation}
  \label{sec:appendix-dataset-preparation}

  In this section, we provide details on the preparation of our experimental datasets.
  Both the data and the preparation code will be made publicly available.

  \paragraph{3D Planar Maps}
    for both the \scannet and \twelvescenes datasets are extracted from their official dense reconstructions using the sequential \seqRANSAC~\cite{fischlerRandomSampleConsensus1981} plane-fitting scripts provided by \cite{liuPlaneRCNN3DPlane2019,xiePlanarReconRealtime3D2022}.
    Specifically, maps from the \scannet are extracted under the guidance of semantic annotations:
    \enparen{1} mesh vertices are first grouped by their semantic instance labels, and plane fitting is performed within each instance that belongs to plane-supporting categories~\enparen{\eg, walls, floors, and tables};
    \enparen{2} vertices identified as inliers supporting a primitive are then projected onto their corresponding plane, while preserving their internal connectivity;
    \enparen{3} adjacent planar primitives within the same category are merged to produce a set of complete and semantically aware planes.
    Lastly, primitives with an area smaller than \qty{0.01}{\meter\squared} are discarded.
    Non-planar vertices and edges connecting distinct primitives are also removed.
    Meanwhile, each primitive is associated with its planar parameters $\planeparameter:=[\planenormal^\top, \planeoffset\,]^\top$, where the plane normal $\planenormal$ is oriented consistently with the original surface normal.

    In contrast, for the \twelvescenes dataset, map primitives are extracted directly using sequential \seqRANSAC without auxiliary semantic annotations and are not further merged, resulting in potentially fragmented and irregular configurations.
    To ensure a level of compactness comparable to that of \scannet, each map primitive in \twelvescenes is further optimized using the Isotropic Explicit Remeshing method implemented in MeshLab~\cite{cignoniMeshLabOpensourceMesh2008}.

\begin{figure}[t]
      \centering
      \includegraphics{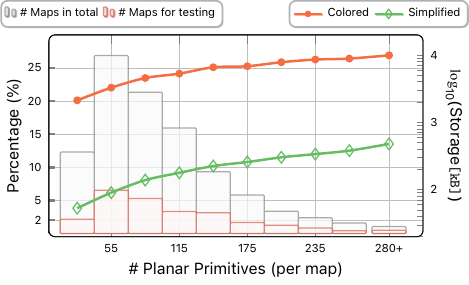}
      \caption{\textbf{Statistics of \num{1513} 3D planar maps constructed from \scannet.}
        Left y-axis: distribution of 3D planar maps grouped by their number of planar primitives.
        Right y-axis: average storage footprint of the colored and simplified map versions in each bin.
      }
      \label{fig:dataset_map_statistics}
    \end{figure}

    To obtain colored maps for baselines that require realistic map appearance in our main experiments~\enparen{\see \cref{tab:scannetv2-pose-results}}, we preserve all plane-supporting vertices along with their original colors.
    Conversely, when visual appearance is not required, the map can be further compressed by retaining only a few key vertices per primitive to represent its spatial extent, using geometry simplification techniques such as the quadric-based edge collapse~\cite{garlandSurfaceSimplificationUsing1997} or cascaded polygon union~\cite{geoscontributorsGEOSComputationalGeometry2025}.
    \Cref{fig:dataset_map_statistics} groups the constructed 3D planar maps from \scannet by the total number of planar primitives.
    For each bin, it reports the proportion of maps~\enparen{left y-axis} and the average storage footprint of both the colored and the simplified map versions~\enparen{right y-axis}.
    The simplified maps occupy an average of \qty{154.3}{\kibi\byte}, which is approximately \num{3.2}\% of the size of their colored counterparts.

  \paragraph{Query Images}
    are sampled from the original RGB-D sequences at regular intervals: every \num{20} and \num{30} frames for the \scannet train and test splits, respectively, and every \num{5} frames for the \twelvescenes test split.
    Following the protocol of \cite{tanNOPESACNeuralOneplane2023}, each sampled frame is then verified using its ground-truth camera pose and depth.
    Frames that fail this consistency check are discarded to ensure precise alignment between the query and the corresponding map.
    Moreover, frames capturing fewer than three map primitives are excluded to guarantee adequate plane observations and geometric constraints for viable pose estimation.

\section{Auxiliary Settings for Baseline Methods}
  \label{sec:appendix-baselines}

\begin{figure}[t]
      \centering
      \includegraphics{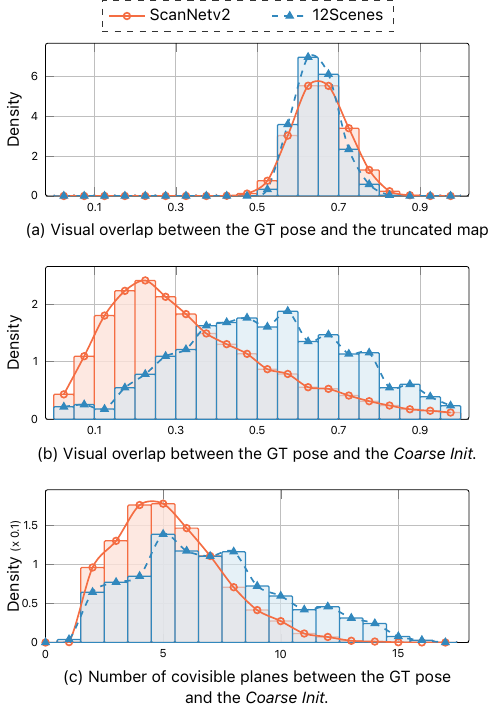}
      \caption{
        \textbf{Visual overlap analysis for the map truncation in \captiona and the oracle coarse initialization strategy in \captionb and \captionc.}
      }
      \label{fig:supplementary-heuristic-analysis}
      \vspace{-.75em}
    \end{figure}

  \paragraph{Map Truncation~\enparen{Map Trunc.}}
    is employed to crop structures far from the ground-truth pose for the image-to-point cloud registration baselines, \ie, \emph{\geotransformer}~\cite{qinGeoTransformerFastRobust2023} and \emph{\ffreereg}~\cite{wangFreeRegImagetoPointCloud2024}, which may struggle to operate stably on the full-scene maps~\enparen{\see \cref{tab:scannetv2-pose-results}}.
    Specifically, given the ground-truth pose and depth map, we define a reference point as the 3D point on the principal ray located at the mean scene depth.
    Then, we retain only the structures within a \qty{3}{\meter}-sized axis-aligned bounding box centered at the reference point.
    As illustrated in \cref{fig:supplementary-heuristic-analysis}\refcaptiona, the strategy yields an average 2D\textendash 3D overlap exceeding \qty{60}{\percent}, as measured by the protocol from \cite{qinGeoTransformerFastRobust2023}.

  \paragraph{Oracle Coarse Initialization~\enparen{Coarse Init.}}
    is used to provide a 6-DoF reference pose for the baselines that rely on map renderings~\enparen{\see \cref{tab:scannetv2-pose-results}}.
    Specifically, we construct a sub-map containing \num{20} primitives
    by first including all visible ones and then supplementing with nearest ones to the reference point defined above.
    To obtain a reference rotation, we first compute the mean normal vector of these primitives.
    The camera's viewing direction is then aligned with the opposite of this average normal vector, towards the map,
    while its up vector is fixed to the global up direction $(0,0,1)^\top$.
    The reference translation is computed by shifting the center of each primitive by 2 meters along its normal direction and then taking the mean of the resulting shifted centers.
    Figures \ref{fig:supplementary-heuristic-analysis}\refcaptionb and \refcaptionc show the distributions of visual overlap~\enparen{as defined by \cite{sarlinLaMARBenchmarkingLocalization2022}} and the number of covisible planes resulting from these heuristic rules, respectively.
    In summary, our oracle initialization heuristics achieve an average visual overlap of \qty{34.7}{\percent}~\enparen{$\sim$\num{5.3} covisible planes} on \scannet and \qty{51.6}{\percent}~\enparen{$\sim$\num{7.0} covisible planes} on \twelvescenes.
    This provides reasonable viewpoints for rendering synthetic images used in matching.

  \paragraph{More Details.}
    For MeshLoc methods, following the protocol of \cite{panekMeshLocMeshbasedVisual2022,panekVisualLocalizationUsing2023}, we render synthetic views using only the model’s base vertex color, without computing any lighting effects, and use PoseLib~\cite{larssonPoseLibMinimalSolvers2020} as the robust pose estimator for all point-matching approaches.
    For all methods, we apply a final clamping step to ensure that the estimated poses lie within the bounds of the scene maps.

\section{Implementation Details for \methodname}
  \label{sec:appendix-implementation}

  \begin{figure*}[htbp]
    \centering
    \includegraphics{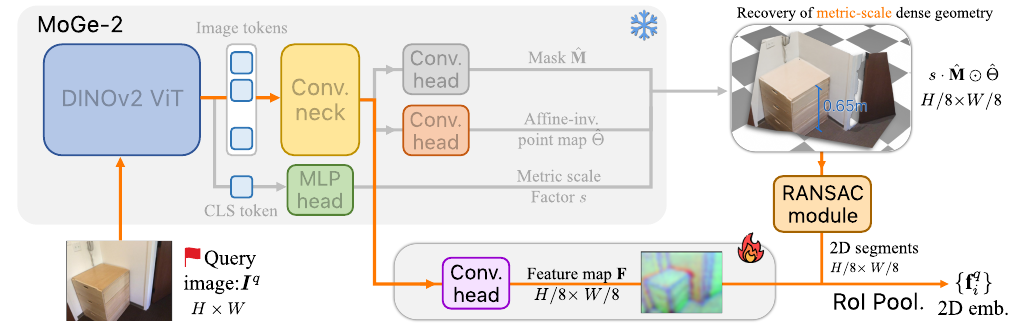}
    \caption{
      \textbf{Architecture of the front-end network on the query side.}
      We retrofit the monocular geometry estimation model \mogevtwo~\cite{wangMoGe2AccurateMonocular2025} with an additional head to encode dense features for 2D plane embedding.
    }
    \label{fig:supplementary-image-encoder-arch}
  \end{figure*}

  \paragraph{2D Plane Embeddings.}
    As shown in \cref{fig:supplementary-image-encoder-arch}, instead of employing an independent image backbone, we augment \mogevtwo~\cite{wangMoGe2AccurateMonocular2025}, the monocular geometry estimation model for plane recovery, with an additional head comprising lightweight convolutional layers to extract dense features for 2D plane embeddings.
    This design allows the model to leverage powerful visual representations learned on large-scale datasets, thereby improving accuracy while maintaining efficiency during both training and inference.
    Further ablation studies of this design choice can be found in \cref{sec:additional-experimental-results}.

    The query images are first resized to a resolution of \num{640} $\times$ \num{480} and fed into \mogevtwo to obtain an estimated metric depth map, which is subsequently downsampled to match the size of the feature map~\enparen{\num{80} $\times$ \num{60}}.
    The \seqRANSAC module~\cite{fischlerRandomSampleConsensus1981,watsonAirPlanesAccuratePlane2024} operates on the downsampled depth map directly, as higher resolution yields little performance improvement but incurs extra computational overhead.
    During plane extraction, a point is considered an inlier of a plane hypothesis if both \enparen{1} its distance residual to the plane is less than \qty{10}{\centi\meter} and \enparen{2} their normal similarity, measured by the dot product, exceeds \num{0.9}.
    The module iteratively extracts planes from the depth map, until either \num{16} primitives have been extracted or the number of inliers falls below \qty{1}{\percent} of the total number of pixels in the depth map.

  \paragraph{3D Plane Embeddings.}
    Both the object and scene encoders are instantiated using PointNet~\cite{qiPointNetDeepLearning2017}, operating on point clouds sampled from the map primitives.
    For each map primitive, we uniformly sample $L$=\num{1024} points, which are then centralized, batched, and fed into the object encoder.
    For the scene encoder, we first retain \num{16} points per primitive to ensure each primitive will be represented.
    Additional points are then randomly sampled from the entire map until the total number of points reaches \num{16} $\times$ \num{1024}=\num{16384}~\enparen{based on the assumption of a maximum of \num{1024} map primitives}.

  \paragraph{Training Scheme.}
    Our model, which consists of the front-end encoders~\enparen{\see\cref{sec:front-end}} and the matching network~\enparen{\see\cref{sec:plane-matching}}, is trained on the \scannet training split by minimizing the loss defined in \cref{equ:loss}.
    We train the model with a batch size of \num{16}, distributed across \num{2} NVIDIA A800 GPUs, for \num{90000} iterations, equivalent to approximately \num{31} epochs.
    The overall training scheme consists of two stages:
    \enparen{1} during the initial \num{45}k iterations, we use the ground-truth primitives augmented with noise to replace the online monocular plane recovery for improved training efficiency;
    \enparen{2} in the remaining \num{45}k iterations, we switch to the full pipeline, enabling online plane recovery.
    For both stages, we employ the AdamW optimizer~\cite{loshchilovDecoupledWeightDecay2019} with an initial learning rate of \num{1e-4}.
    The learning rate is decreased by a factor of \num{0.1} at \num{24}k and \num{36}k iterations using a multi-step scheduler.
    Throughout training, we apply data augmentation to both the query images~\enparen{random resizing and cropping} and the maps~\enparen{random rotation and scaling}.
    The entire training process completes in about \qty{12}{\hour}.

  \paragraph{Estimating Camera Translation}, as introduced in \cref{equ: estimating-camera-translation},
    involves the joint optimization of the metric scale factor $\scale$:
    \begin{equation*}
      \initPos, \scale^* = \argmin_{\Pos, s} \sum_{(i,j)\in \inliercorrespondence} \weightforpos_i(\Pos^\top\planenormal^m_j-\planeoffset^m_j+\scale\planeoffset^q_i)^2.
    \end{equation*}
    The initial translation $\initPos$ and the optimal $\scale^*$ can be solved efficiently by rewriting the above equation as a standard linear least-squares problem:
    \enparen{1} for each correspondence $(i,j)\in \inliercorrespondence$, we construct a linear equation by setting the row vector to $a := [(\planenormal^m_j)^\top, \planeoffset^q_i]$ and the target to $b := \planeoffset^m_j$;
    \enparen{2} stacking all correspondences yields a linear system $\mA\rvx\approx\vb$, with the variable vector $\rvx:=[\initPos^\top, \scale]^\top\in\mathbb{R}^4$, the data matrix $\mA\in\mathbb{R}^{|\inliercorrespondence|\times 4}$, and the target $\vb\in\mathbb{R}^{|\inliercorrespondence|}$;
    \enparen{3} we introduce the diagonal weight matrix $\mW=\diagonal(\sqrt{\weightforpos_i})$ and formulate the final weighted linear least-squares problem as
    \begin{equation}
      \rvx^* = \argmin_{\rvx} \|\mW(\mA\rvx - \vb)\|_2^2,
      \label{equ:weighted-linear-least-squares}
    \end{equation}
    which can be efficiently solved in closed form using SciPy~\cite{virtanenSciPy10Fundamental2020}.
    The degeneracy rate, \ie, the percentage of cases where the number of correspondences with non-parallel normals is less than \qty{3}, is empirically observed to be less than \qty{2}{\percent}.

  \paragraph{Pose Refinement.}
    We optimize for the relative transformation $\refinedpose^*$ alongside the \offsetseedsname $\{\offsetseed_i^*\}$ by minimizing the depth alignment cost $\depthcost$ defined in \cref{equ:depth-alignment-cost}.
    The optimization is performed using the Adam optimizer~\cite{kingmaAdamMethodStochastic2015}, following the practice in \cite{mazurSuperPrimitiveSceneReconstruction2024}.
    Specifically, the optimization variable $\refinedpose$ is converted into a differentiable 6-dimensional vector via LieTorch~\cite{teedTangentSpaceBackpropagation2021} and then optimized for \num{200} iterations with a learning rate of \num[exponent-mode = scientific]{1e-3}.
    The \offsetseedsname $\{\offsetseed_i\}$ are initialized to one and optimized with a learning rate of \num[exponent-mode = scientific]{1e-4}.
    For efficiency, in each iteration we compute $\depthcost$ on a multinomial sample of \num{4096} pixels from the 2D plane segments, rather than capitalizing on every pixel.

\section{Additional Experimental Results}
  \label{sec:additional-experimental-results}
  \begin{figure}[t]
    \centering
    \includegraphics{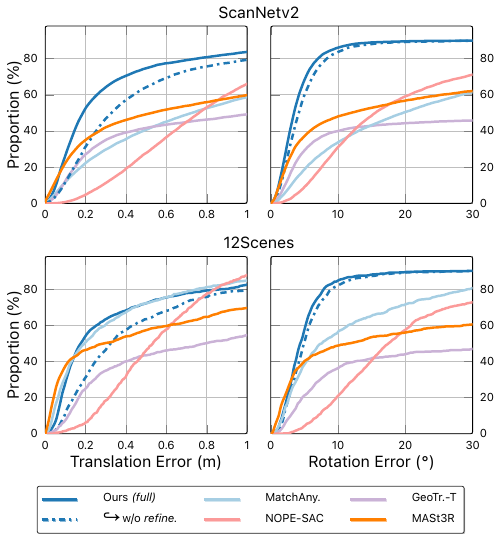}
    \caption{\textbf{Camera relocalization results on \scannet and \twelvescenes datasets.}
      We plot cumulative accuracy curves that show the proportion of correctly localized frames as a function of varying translation and rotation error thresholds.
    }
    \label{fig:cumulative_error}
  \end{figure}

  \paragraph{Cumulative Accuracy Curves.}
    \Cref{fig:cumulative_error} presents the cumulative accuracy curves \wrt translation and rotation errors on both \scannet and \twelvescenes datasets.
    Our method achieves solid performance on both datasets, with particularly favorable results in camera rotation,
    which may benefit from the reliable normal priors predicted by the powerful monocular model.
    In contrast, estimating camera translation is largely affected by depth inaccuracy, a limitation that can be mitigated through the \postrefine procedure introduced in \cref{sec:pose-refinement}.

  \paragraph{Ablating 2D Encoder Variants.}
    We compare different designs of the query-side encoder.
    In addition to our default design depicted in \cref{fig:supplementary-image-encoder-arch}, we also evaluate three alternative configurations:
    \enparen{1} training a dedicated ResNet-50~\cite{heDeepResidualLearning2016} from scratch;
    \enparen{2} employing the official pretrained DINOv2-Vit-L/14~\cite{oquabDINOv2LearningRobust2023} as a frozen backbone;
    \enparen{3} a variant of \enparen{2} augmented with the same learnable convolutional head as in our default design to better adapt the features to our task.
    As reported in \cref{tab:supplementary_ablation_2dEncoders},
    the results suggest that the DINOv2 ViT encoder fine-tuned by \mogevtwo~\cite{wangMoGe2AccurateMonocular2025} possesses enhanced geometric and structural perception, contributing to the slight performance improvement over the official frozen DINOv2 features.
    Moreover, the reuse of visual representations from the pretrained backbone for 2D plane embedding provides good computational efficiency.
    Collectively, these results highlight the effectiveness of our architectural design.

    \begin{table}[t]
      \centering
      \caption{
        \textbf{Ablation study on 2D encoder variants.}
        Experiments are conducted on \scannet with no \postrefine.
      }
      \setlength{\tabcolsep}{4pt}
      \resizebox{\linewidth}{!}{%
        {
            
\begin{tabular}{lccccc}
  \Xhline{3\arrayrulewidth}                                                                                                                                                                                                                                                                                                                                                                    \\[-0.9em]
                                                                   & \multicolumn{3}{c}{\TAblationMatchingPerformance} & \multirow{2}{*}{\rotatebox[origin=c]{0}{\makecell[t]{\TAblationPoseAccHigh}}} & \multirow{2}{*}{\rotatebox[origin=c]{0}{\makecell[t]{\TTimeMilliSecond}}}                                                                                                             \\[-0.05em]
  \cmidrule(r){2-4}                                                                                                                                                                                                                                                                                                                                                                            \\[-1.3em]
  \TMethods                                                        & \TAblationMatchPrecisionPctNoArrow                & \TAblationMatchRecallPctNoArrow                                               & \TAblationMatchFScorePctNoArrow                                           &                               &                                                                           \\[-0.05em]
  \Xhline{.3\arrayrulewidth}                                                                                                                                                                                                                                                                                                                                                                   \\[-0.9em]
  \TAblationEncoderResNet~\cite{heDeepResidualLearning2016}        & \tablenum{62.3}                                   & \tablenum{56.9}                                                               & \tablenum{59.5}                                                           & \tablenum{34.2}               & \Tsim\tablenum[round-mode = places, round-precision=1]{63.86899999999999} \\[0.05em]
  \TAblationEncoderDINONoNeck~\cite{oquabDINOv2LearningRobust2023} & \TRankThird{\tablenum{65.5}}                      & \TRankThird{\tablenum{59.6}}                                                  & \TRankThird{\tablenum{62.4}}                                              & \TRankThird{\tablenum{35.5}}  & \Tsim\tablenum[round-mode = places, round-precision=1]{94.979}            \\[0.05em]
  \TAblationEncoderDINOWithNeck                                    & \TRankSecond{\tablenum{66.3}}                     & \TRankSecond{\tablenum{60.3}}                                                 & \TRankSecond{\tablenum{63.1}}                                             & \TRankSecond{\tablenum{36.2}} & \Tsim\tablenum[round-mode = places, round-precision=1]{97.079}            \\[0.05em]
  \Xhline{.3\arrayrulewidth}                                                                                                                                                                                                                                                                                                                                                                   \\[-0.9em]
  \TAblationEncoderOurs                                            & \TRankFirst{\tablenum{67.6}}                      & \TRankFirst{\tablenum{61.3}}                                                  & \TRankFirst{\tablenum{64.3}}                                              & \TRankFirst{\tablenum{37.1}}  & \Tsim\tablenum[round-mode = places, round-precision=1]{59.854}            \\[0.05em]
  \Xhline{3\arrayrulewidth}                                                                                                                                                                                                                                                                                                                                                                    \\[-0.9em]
\end{tabular}

          }
      }
      \label{tab:supplementary_ablation_2dEncoders}
      \vspace{-1em}
    \end{table}

\begin{figure}
      \centering
      \includegraphics{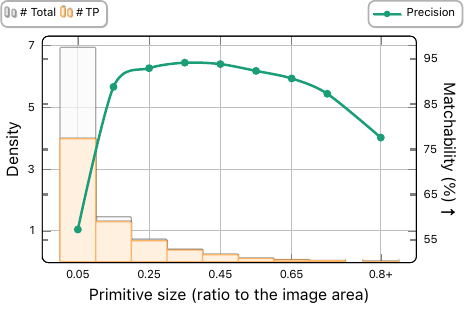}
      \caption{
        \textbf{Impact of the Query Primitive Size on Matching Performance.}
        Left y-axis: the size distribution of all detected query primitives, where size is defined as the percentage of the image area occupied.
        Right y-axis: the matching precision for each size bin.
      }
      \label{fig:supplementary-plane-areas}
      \vspace{-1em}
    \end{figure}

  \paragraph{Impact of the Query Primitive Size.}
    We analyze how plane matching performance varies \wrt the size of primitives observed in query images.
    Specifically, all detected query primitives on \scannet are collected and categorized into bins according to their pixel areas.
    We then compute the matching precision for each bin, defined as the proportion of correctly matched primitives.
    As shown in \cref{fig:supplementary-plane-areas}, our purely geometric plane recovery method tends to over-segment planar regions due to occlusions and noise.
    Consequently, a large number of small plane segments are produced, typically covering less than \qty{10}{\percent} of the image and exhibiting low matching reliability.
    Moreover, the Mutual Nearest Neighbor~\enparen{MNN} matching strategy, which imposes a one-to-one correspondence constraint, also leads to lower matching precision in these small planar primitives.
    Meanwhile, performance drops significantly for extremely large planes that occupy more than \qty{80}{\percent} of the image area.
    As noted in our main paper, this likely stems from the reduced plane richness and the consequent lack of discriminative patterns.
    In contrast, the remaining medium-sized query primitives strike a good balance between salience and discriminativeness, achieving a remarkable matching precision of over \qty{90}{\percent} on average.

  \paragraph{Evaluation on \sevenscenes.}
    To better position the task and the \methodname method within the broader visual localization literature, we additionally include a cross-dataset evaluation on the standard \sevenscenes dataset~\cite{shottonSceneCoordinateRegression2013} to compare against more representative methods and assess the performance gap.
    Specifically, we construct planar maps using the depth scans from the \sevenscenes train split, and evaluate \methodname on the test split.
    The results are reported in \cref{tab:supplementary_s7scenes}.
    While maintaining a compact map representation and exhibiting consistent cross-dataset performance, we acknowledge that \methodname still lags behind state-of-the-art visual localization methods that fully leverage thousands of reference images which provides rich visual cues and pose priors.

    \begin{table*}[htbp]
      \centering
      \caption{
        \textbf{Camera relocalization results on the \sevenscenes dataset.}
        We report median position and rotation errors in centimeters~\enparen{\si{\centi\meter}} and degrees~\enparen{\si{\degree}}, respectively.
        We summarize the map type, map size, the time needed for mapping, and whether mapping and localization rely on visual appearance or pose priors~\enparen{\eg, image retrieval}.
        Results of other methods are taken from the literature.
      }
      \setlength{\tabcolsep}{2pt} %
      \resizebox{\textwidth}{!}{%
        
\begin{tabular}{lccccccccccccc}
  \Xhline{3\arrayrulewidth}                                                                                                                                                                                                                                                                                                                                                                                                                                                                                                                                                                                                                                                                                                                                                                                                                                                                                                                                                                                                                  \\[-0.7em]
  \multirow{2}{*}{\rotatebox[origin=c]{0}{\makecell[t]{\TRebuttalMethods}}} & \multirow{2}{*}{\rotatebox[origin=c]{0}{\makecell[t]{\TRebuttalMapType}}} & \multirow{2}{*}{\rotatebox[origin=c]{0}{\makecell[t]{\TRebuttalMapSize}}} & \multirow{2}{*}{\rotatebox[origin=c]{0}{\makecell[t]{\TRebuttalMappingTime}}} & \multirow{2}{*}{\rotatebox[origin=c]{0}{\makecell[t]{\TRebuttalVisualAppearance}}} & \multirow{2}{*}{\rotatebox[origin=c]{0}{\makecell[t]{\TRebuttalRetrieval}}} & \multirow{2}{*}{\rotatebox[origin=c]{0}{\makecell[t]{Chess}}} & \multirow{2}{*}{\rotatebox[origin=c]{0}{\makecell[t]{Fire}}} & \multirow{2}{*}{\rotatebox[origin=c]{0}{\makecell[t]{Heads}}} & \multirow{2}{*}{\rotatebox[origin=c]{0}{\makecell[t]{Office}}} & \multirow{2}{*}{\rotatebox[origin=c]{0}{\makecell[t]{Pumpkin}}} & \multirow{2}{*}{\rotatebox[origin=c]{0}{\makecell[t]{Kitchen}}} & \multirow{2}{*}{\rotatebox[origin=c]{0}{\makecell[t]{Stairs}}} & \multirow{2}{*}{\rotatebox[origin=c]{0}{\makecell[t]{\textbf{Avg.$\downarrow$} \\ \enparen{\si{\centi\meter}/\si{\degree}}}}} \\
                                                                            &                                                                           &                                                                           &                                                                               &                                                                                    &                                                                             &                                                               &                                                              &                                                               &                                                                &                                                                 &                                                                 &                                                                &                                                                                \\[0.3em]
  \Xhline{.3\arrayrulewidth}                                                                                                                                                                                                                                                                                                                                                                                                                                                                                                                                                                                                                                                                                                                                                                                                                                                                                                                                                                                                                 \\[-0.8em]
  \TRebuttalPoseNet                                                         & \TRebuttalMapTypeNetwork                                                  & \qty{50}{\mega\byte}                                                      & \qtyrange{4}{24}{\hour}                                                       & \Tcheckmark                                                                        &                                                                             & 13/4.5                                                        & 27/11.3                                                      & 17/13.0                                                       & 19/5.6                                                         & 26/4.8                                                          & 23/5.4                                                          & 35/12.4                                                        & 23/8.1                                                                         \\[0.3em]
  \TRebuttalhLoc                                                            & \TRebuttalMapTypeSfMPoints                                                & $\sim$\qty{2}{\giga\byte}                                                 & $\sim$ \qty{1.5}{\hour}                                                       & \Tcheckmark                                                                        & \Tcheckmark                                                                 & 2/0.8                                                         & 2/0.9                                                        & 1/0.8                                                         & 3/0.9                                                          & 5/1.3                                                           & 4/1.4                                                           & 5/1.5                                                          & 3/1.1                                                                          \\[0.3em]
  \TRebuttalGoMatch                                                         & \TRebuttalMapTypeSfMPoints                                                & $\sim$\qty{56}{\mega\byte}                                                & $\sim$ \qty{1.5}{\hour}                                                       &                                                                                    & \Tcheckmark                                                                 & 4/1.6                                                         & 12/3.7                                                       & 5/3.4                                                         & 7/1.8                                                          & 8/5.7                                                           & 14/3.0                                                          & 58/13.1                                                        & 18/4.6                                                                         \\[0.3em]
  \TRebuttalACE                                                             & \TRebuttalMapTypeNetwork                                                  & \qty{5}{\mega\byte}                                                       & \qty{5}{\minute}                                                              & \Tcheckmark                                                                        &                                                                             & 0.6/0.2                                                       & 0.8/0.3                                                      & 0.5/0.3                                                       & 1/0.3                                                          & 1/0.2                                                           & 0.8/0.2                                                         & 3/0.8                                                          & 1/0.3                                                                          \\[0.3em]
  \TRebuttalRelocr                                                          & \TRebuttalMapTypePoseImages                                               & $\sim$\TRebuttalRelocrMapSize                                             & \qty{7}{\second}                                                              & \Tcheckmark                                                                        & \Tcheckmark                                                                 & 3/0.9                                                         & 3/0.8                                                        & 1/1.0                                                         & 4/0.9                                                          & 6/1.1                                                           & 4/1.3                                                           & 7/1.3                                                          & 4/1.0                                                                          \\[0.3em]
  \TRebuttalSTDLoc                                                          & \TRebuttalMapTypeGS                                                       & $\sim$\qty{0.8}{\giga\byte}                                               & $\sim$ \qty{2}{\hour}                                                         & \Tcheckmark                                                                        &                                                                             & 0.5/0.2                                                       & 0.6/0.2                                                      & 0.4/0.3                                                       & 1/0.2                                                          & 1/0.2                                                           & 0.6/0.2                                                         & 1/0.4                                                          & 0.8/0.2                                                                        \\[0.3em]
  \Xhline{.3\arrayrulewidth}                                                                                                                                                                                                                                                                                                                                                                                                                                                                                                                                                                                                                                                                                                                                                                                                                                                                                                                                                                                                                 \\[-0.8em]
  \TRebuttalOurs                                                            & \TRebuttalMapTypePlanes                                                   & \qty{0.4}{\mega\byte}                                                     & \qty{2}{\minute}                                                              &                                                                                    &                                                                             & 26/3.7                                                        & 19/3.8                                                       & 32/3.7                                                        & 19/3.6                                                         & 27/2.6                                                          & 43/3.3                                                          & 61/6.2                                                         & 32/3.9                                                                         \\[0.3em]
  \Xhline{3\arrayrulewidth}                                                                                                                                                                                                                                                                                                                                                                                                                                                                                                                                                                                                                                                                                                                                                                                                                                                                                                                                                                                                                  \\[-1.0em]
\end{tabular}

      }
      \label{tab:supplementary_s7scenes}
    \end{table*}

\begin{figure*}
      \centering
      \begin{tikzpicture}
        \node[anchor=south west,inner sep=0] (image) at (0,0) {
          \includegraphics{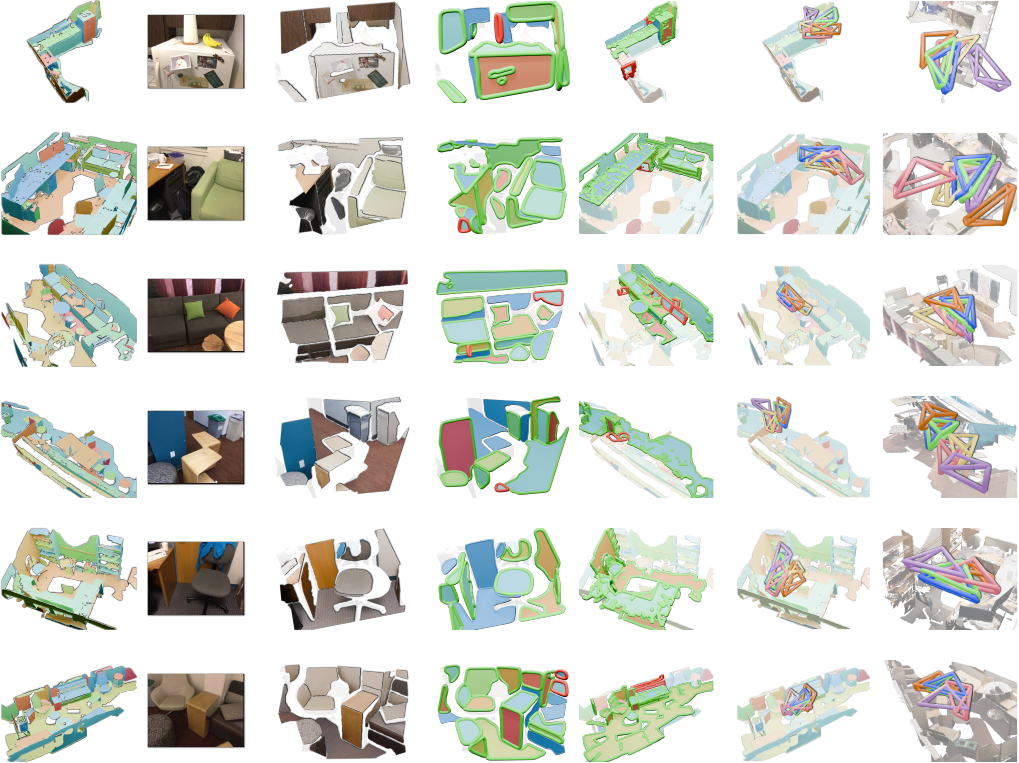}
        };
        {
        \def\subtitley{-0.05}
        \begin{scope}[x={(image.south east)},y={(image.north west)}]
          \node[fill=white, fill opacity=0.7, text opacity=1, inner sep=2pt] at (0.128, \subtitley) {\footnotesize \captiona Input: map \& query};
          \node[fill=white, fill opacity=0.7, text opacity=1, inner sep=2pt] at (0.34, \subtitley) {\footnotesize \captionb Monocular plane recovery};
          \node[fill=white, fill opacity=0.7, text opacity=1, inner sep=2pt] at (0.579, \subtitley) {\footnotesize \captionc Plane correspondences};
          \node[fill=white, fill opacity=0.7, text opacity=1, inner sep=2pt] at (0.871, \subtitley) {\footnotesize \captiond Poses (viewpoint 1 \& 2)};
        \end{scope}
        }
      \end{tikzpicture}
      \caption{
        \textbf{Qualitative examples on \twelvescenes.}
        Correspondences in \captionc are color-coded, with true positives outlined in green and false ones in red.
        Legend for different relocalizers in \captiond:
        \qelegend{gt}\textcolor{tikzColorMethodGT}{\gtpose},
        \qelegend{ours}\textcolor{tikzColorMethodOurs}{\methodname\enparen{Ours}},
        \qelegend{geotransformer}\textcolor{tikzColorMethodGeoTr!
          60!black}{\geotransformer-T},
        \qelegend{init}\textcolor{tikzColorMethodCoarse}{\AbbrCoarseInit},
        \qelegend{mast3r}\textcolor{tikzColorMethodMast3r!80!black}{\mastr},
        \qelegend{nope-sac}\textcolor{tikzColorMethodNOPESAC!70!black}{\nopesac}.
      }
      \label{fig:supplementary-qualitative-examples-12scenes}
    \end{figure*}

\begin{figure*}
      \centering
      \begin{tikzpicture}
        \node[anchor=south west,inner sep=0] (image) at (0,0) {
          \includegraphics{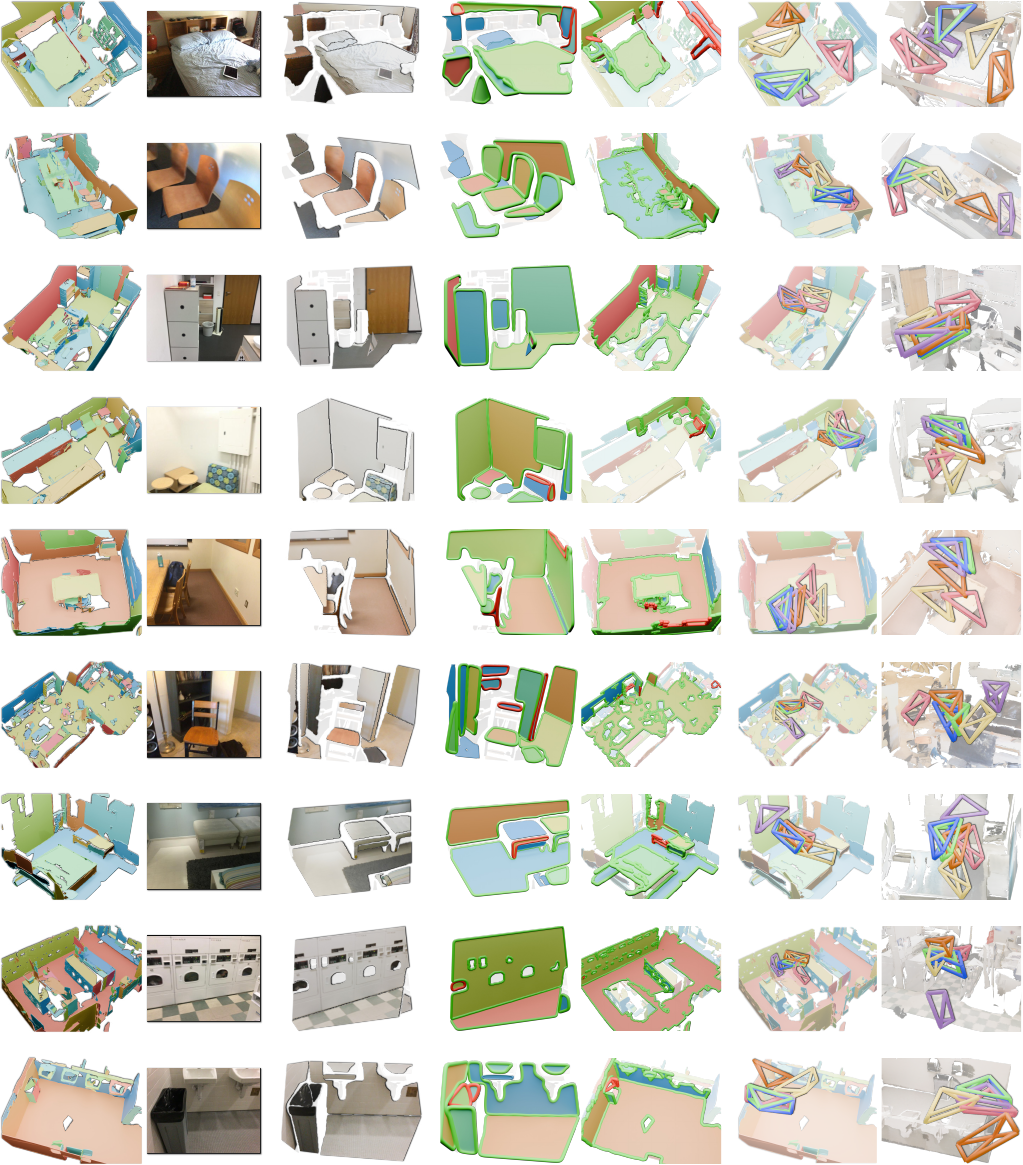}
        };
        {
        \def\subtitley{-0.02}
        \begin{scope}[x={(image.south east)},y={(image.north west)}]
          \node[fill=white, fill opacity=0.7, text opacity=1, inner sep=2pt] at (0.128, \subtitley) {\footnotesize \captiona Input: map \& query};
          \node[fill=white, fill opacity=0.7, text opacity=1, inner sep=2pt] at (0.34, \subtitley) {\footnotesize \captionb Monocular plane recovery};
          \node[fill=white, fill opacity=0.7, text opacity=1, inner sep=2pt] at (0.579, \subtitley) {\footnotesize \captionc Plane correspondences};
          \node[fill=white, fill opacity=0.7, text opacity=1, inner sep=2pt] at (0.871, \subtitley) {\footnotesize \captiond Poses (viewpoint 1 \& 2)};
        \end{scope}
        }
      \end{tikzpicture}
      \caption{
        \textbf{More qualitative examples on \scannet.}
        Correspondences in \captionc are color-coded, with true positives outlined in green and false ones in red.
        Legend for different relocalizers in \captiond:
        \qelegend{gt}\textcolor{tikzColorMethodGT}{\gtpose},
        \qelegend{ours}\textcolor{tikzColorMethodOurs}{\methodname\enparen{Ours}},
        \qelegend{geotransformer}\textcolor{tikzColorMethodGeoTr!
          60!black}{\geotransformer-T},
        \qelegend{init}\textcolor{tikzColorMethodCoarse}{\AbbrCoarseInit},
        \qelegend{mast3r}\textcolor{tikzColorMethodMast3r!80!black}{\mastr},
        \qelegend{nope-sac}\textcolor{tikzColorMethodNOPESAC!70!black}{\nopesac}.
      }
      \label{fig:supplementary-qualitative-examples-scannet}
    \end{figure*}

  \paragraph{Additional Qualitative Results.}
    More visualizations of intermediate outputs and relocalization results on \twelvescenes and \scannet are shown in \cref{fig:supplementary-qualitative-examples-12scenes} and \cref{fig:supplementary-qualitative-examples-scannet}, respectively.

  \section{Limitations and Future Work}
  \label{sec:appendix-limitations}

  \paragraph{Limitations}
    A key bottleneck of our method lies in the monocular plane recovery module, given its critical role in providing 2D plane proposals and geometric priors for subsequent matching and pose estimation.
    Despite significant progress in this area,
    unreliable predictions from this module under challenging scenarios can still lead to catastrophic failures, even with our robust pose estimation and refinement pipeline designed to mitigate errors.

\begin{figure}[!b]
      \centering
      \includegraphics{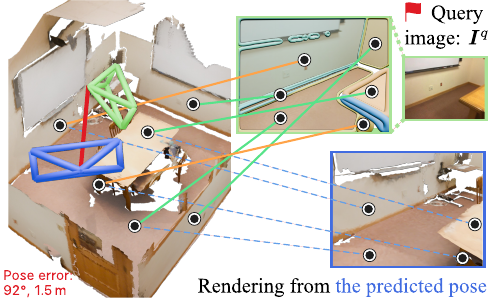}
      \caption{
        \textbf{A representative case where \methodname underperforms.}
        Despite four out of six primitives being correctly matched~\enparen{colored in \textcolor{matchinlier!
            90!black}{green}}, the pose estimation framework fails to reject matching outliers~\enparen{colored in \textcolor{matchoutlier!90!black}{orange}} due to the perfectly repeated pattern~\enparen{compare the query image with the colored rendering from the predicted pose}.
      }
      \label{fig:failure-cases-similar-structures}
    \end{figure}

    Another issue arises in environments with a limited level of detail or exhibiting highly repetitive structures, or when the query image captures only weak structural hints~\enparen{\see \cref{fig:failure-cases-similar-structures}}.
    This limitation is also indicated by \cref{tab:ablation-2d-detectors} in the main paper: even when provided with ground-truth monocular plane recoveries, \methodname may still fail to establish enough correct matches in certain cases.

\begin{figure}[htbp]
      \centering
      \includegraphics{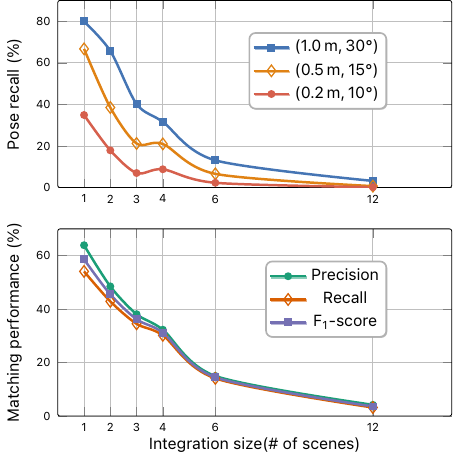}
      \caption{
        \textbf{Relocalization results on Integrated Rooms.}
        Following \cite{brachmannExpertSampleConsensus2019}, we arrange scenes in \twelvescenes inside a 2D grid with a cell size of \qty{5}{\meter} and integrate varying numbers of adjacent scenes to form larger maps.
        As the integration size increases, \methodname's performance degrades due to increased ambiguities and the scene encoder's limited capacity for larger maps.
      }
      \label{fig:supplementary-integrated-rooms}
    \end{figure}

    Furthermore, in large multi-room scenarios~\enparen{\see \cref{fig:supplementary-integrated-rooms}}, \methodname's performance is constrained by the increasing structural ambiguities and the fixed point budget that the scene encoder consumes.
    Increasing the point budget or processing subdivided regions in parallel yield limited performance improvements, but at the cost of substantial memory and computational overhead.

    Finally, \methodname is currently better suited to indoor settings and is not trained or validated in outdoor environments, where the plane distribution may differ significantly.

  \paragraph{Future Work.}
    Although \methodname demonstrates strong performance in cross-modal 2D\textendash 3D matching and enables a plane-centric paradigm for room-level 6-DoF camera relocalization, scaling it up to larger and more complex scenes demands improved scene understanding and structural disambiguation.
    This could be addressed by enhancing structural feature encoding, incorporating plane semantics, and adopting a coarse-to-fine strategy.
    Moreover, exploring an end-to-end approach that jointly tackles structural matching and pose estimation could further improve robustness and accuracy.
    Lastly, extending the method to sequential inputs offers another promising direction for practical use.

{
  \small
  \bibliographystyle{ieeenat_fullname}
  \bibliography{src/reference-cr}

@inproceedings{abeNormalLocVisualLocalization2025,
  title = {{{NormalLoc}}: {{Visual}} Localization on Textureless {{3D}} Models Using Surface Normals},
  shorttitle = {{{NormalLoc}}},
  booktitle = {{{ICCV}}},
  author = {Abe, Jiro and Nakano, Gaku and Ogura, Kazumine},
  year = 2025,
  pages = {25421--25430}
}

@inproceedings{agarwalaPlaneFormersSparseView2022,
  title = {{{PlaneFormers}}: {{From}} Sparse View Planes to {{3D}} Reconstruction},
  shorttitle = {{{PlaneFormers}}},
  booktitle = {{{ECCV}}},
  author = {Agarwala, Samir and Jin, Linyi and Rockwell, Chris and Fouhey, David F.},
  year = 2022,
  volume = {13663},
  pages = {192--209},
  publisher = {Springer},
  url = {https://doi.org/10.1007/978-3-031-20062-5_12}
}

@inproceedings{anMinCDPnPLearning2D3D2025,
  title = {{{MinCD-PnP}}: {{Learning 2D-3D}} Correspondences with Approximate Blind {{PnP}}},
  shorttitle = {{{MinCD-PnP}}},
  booktitle = {{{ICCV}}},
  author = {An, Pei and Yang, Jiaqi and Peng, Muyao and Yang, You and Liu, Qiong and Wu, Xiaolin and Nan, Liangliang},
  year = 2025,
  pages = {26519--26528}
}

@inproceedings{arandjelovicNetVLADCNNArchitecture2016,
  title = {{{NetVLAD}}: {{CNN}} Architecture for Weakly Supervised Place Recognition},
  shorttitle = {{{NetVLAD}}},
  booktitle = {{{CVPR}}},
  author = {Arandjelovi{\'c}, Relja and Gronat, Petr and Torii, Akihiko and Pajdla, Tomas and Sivic, Josef},
  year = 2016,
  eprint = {1511.07247},
  publisher = {arXiv},
  url = {http://arxiv.org/abs/1511.07247},
  urldate = {2024-11-30},
  archiveprefix = {arXiv}
}

@manual{arcoreFundamentalConceptsEnvironmental2025,
  title = {Fundamental {{Concepts}}: {{Environmental Understanding}}},
  shorttitle = {{{ARCore}}},
  author = {ARCore},
  year = 2025,
  booktitle = {Google for Developer: Augmented Reality Essentials},
  url = {https://developers.google.com/ar/develop/fundamentals},
  urldate = {2025-10-23}
}

@manual{arkitPlacingContentDetected2025,
  title = {Placing Content on Detected Planes},
  shorttitle = {{{ARKit}}},
  author = {ARKit},
  year = 2025,
  booktitle = {Apple Developer Documentation},
  url = {https://developer.apple.com/documentation/visionos/placing-content-on-detected-planes},
  urldate = {2025-10-23}
}

@article{bavleSituationalGraphsRobot2022,
  title = {Situational Graphs for Robot Navigation in Structured Indoor Environments},
  author = {Bavle, Hriday and {Sanchez-Lopez}, Jose Luis and Shaheer, Muhammad and Civera, Javier and Voos, Holger},
  year = 2022,
  journal = {IEEE Robotics Autom. Lett.},
  volume = {7},
  number = {4},
  pages = {9107--9114}
}

@inproceedings{bochkovskiy2025depth,
  title = {Depth pro: {{Sharp}} Monocular Metric Depth in Less than a Second},
  shorttitle = {Depth Pro},
  booktitle = {{{ICLR}}},
  author = {Bochkovskiy, Alexey and Delaunoy, Ama{\"e}l and Germain, Hugo and Santos, Marcel and Zhou, Yichao and Richter, Stephan and Koltun, Vladlen},
  year = 2025,
  url = {https://openreview.net/forum?id=aueXfY0Clv}
}

@inproceedings{brachmannAcceleratedCoordinateEncoding2023,
  title = {Accelerated Coordinate Encoding: {{Learning}} to Relocalize in Minutes Using {{RGB}} and Poses},
  shorttitle = {{{ACE}}},
  booktitle = {{{CVPR}}},
  author = {Brachmann, Eric and Cavallari, Tommaso and Prisacariu, Victor Adrian},
  year = 2023,
  pages = {5044--5053},
  publisher = {IEEE}
}

@inproceedings{brachmannDSACDifferentiableRANSAC2016,
  title = {{{DSAC}} --- Differentiable {{RANSAC}} for Camera Localization},
  booktitle = {{{CVPR}}},
  author = {Brachmann, Eric and Krull, Alexander and Nowozin, Sebastian and Shotton, Jamie and Michel, Frank and Gumhold, Stefan and Rother, Carsten},
  year = 2016,
  pages = {2492--2500},
  url = {https://api.semanticscholar.org/CorpusID:4001530}
}

@inproceedings{brachmannExpertSampleConsensus2019,
  title = {Expert {{Sample Consensus Applied}} to {{Camera Re-Localization}}},
  booktitle = {{{ICCV}}},
  author = {Brachmann, Eric and Rother, Carsten},
  year = 2019,
  pages = {7524--7533},
  publisher = {IEEE}
}

@inproceedings{brejchaLandscapeARLargeScale2020,
  title = {{{LandscapeAR}}: {{Large Scale Outdoor Augmented Reality}} by {{Matching Photographs}} with {{Terrain Models Using Learned Descriptors}}},
  shorttitle = {{{LandscapeAR}}},
  booktitle = {{{ECCV}}},
  author = {Brejcha, Jan and Luk{\'a}{\v c}, Michal and {Hold-Geoffroy}, Yannick and Wang, Oliver and {\v C}ad{\'i}k, Martin},
  year = 2020,
  volume = {12374},
  pages = {295--312},
  publisher = {Springer International Publishing},
  url = {https://link.springer.com/10.1007/978-3-030-58526-6_18},
  urldate = {2025-11-02}
}

@inproceedings{campbellSolvingBlindPerspectivenpoint2020,
  title = {Solving the Blind Perspective-n-Point Problem End-to-End with Robust Differentiable Geometric Optimization},
  shorttitle = {{{BPnPNet}}},
  booktitle = {{{ECCV}}},
  author = {Campbell, Dylan and Liu, Liu and Gould, Stephen},
  year = 2020,
  pages = {244--261},
  publisher = {Springer}
}

@inproceedings{camposecoHybridSceneCompression2019a,
  title = {Hybrid Scene Compression for Visual Localization},
  booktitle = {{{CVPR}}},
  author = {Camposeco, Federico and Cohen, Andrea and Pollefeys, Marc and Sattler, Torsten},
  year = 2019,
  pages = {7653--7662}
}

@inproceedings{chenF$^3$LocFusionFiltering2024,
  title = {F$^\text{3}${{Loc}}: {{Fusion}} and {{Filtering}} for {{Floorplan Localization}}},
  shorttitle = {F$^\text{3}${{Loc}}},
  booktitle = {{{CVPR}}},
  author = {Chen, Changan and Wang, Rui and Vogel, Christoph and Pollefeys, Marc},
  year = 2024,
  eprint = {2403.03370},
  primaryclass = {cs},
  publisher = {arXiv},
  url = {http://arxiv.org/abs/2403.03370},
  urldate = {2025-03-02},
  archiveprefix = {arXiv}
}

@inproceedings{chengMaskedattentionMaskTransformer2022,
  title = {Masked-Attention {{Mask Transformer}} for {{Universal Image Segmentation}}},
  shorttitle = {{{Mask2Former}}},
  booktitle = {{{CVPR}}},
  author = {Cheng, Bowen and Misra, Ishan and Schwing, Alexander G. and Kirillov, Alexander and Girdhar, Rohit},
  year = 2022,
  eprint = {2112.01527},
  primaryclass = {cs},
  publisher = {arXiv},
  url = {http://arxiv.org/abs/2112.01527},
  urldate = {2025-03-10},
  archiveprefix = {arXiv}
}

@inproceedings{chenPlanarNeRFOnlineLearning2025,
  title = {{{PlanarNeRF}}: {{Online Learning}} of {{Planar Primitives}} with {{Neural Radiance Fields}}},
  shorttitle = {{{PlanarNeRF}}},
  booktitle = {{{ICRA}}},
  author = {Chen, Zheng and Yan, Qingan and Zhan, Huangying and Cai, Changjiang and Xu, Xiangyu and Huang, Yuzhong and Wang, Weihan and Feng, Ziyue and Liu, Lantao and Xu, Yi},
  year = 2025,
  publisher = {arXiv},
  urldate = {2024-02-20}
}

@inproceedings{chenRefinementAbsolutePose2024,
  title = {Refinement for {{Absolute Pose Regression}} with {{Neural Feature Synthesis}}},
  shorttitle = {{{NeFeS}}},
  booktitle = {{{CVPR}}},
  author = {Chen, Shuai and Bhalgat, Yash and Li, Xing Hui and Bian, Jia Wang and Li, Ke Jie and Wang, Zirui and Prisacariu, Victor Adrian},
  year = 2024,
  eprint = {2303.10087},
  primaryclass = {cs},
  pages = {20987--20996},
  publisher = {IEEE},
  url = {http://arxiv.org/abs/2303.10087},
  urldate = {2023-12-24},
  archiveprefix = {arXiv}
}

@inproceedings{chenSimpleFrameworkContrastive2020,
  title = {A Simple Framework for Contrastive Learning of Visual Representations},
  shorttitle = {{{SimCLR}}},
  booktitle = {{{ICML}}},
  author = {Chen, Ting and Kornblith, Simon and Norouzi, Mohammad and Hinton, Geoffrey},
  year = 2020,
  pages = {1597--1607},
  publisher = {PmLR}
}

@inproceedings{chopraLearningSimilarityMetric2005,
  title = {Learning a Similarity Metric Discriminatively, with Application to Face Verification},
  shorttitle = {{{DrLIM}}},
  booktitle = {{{CVPR}}},
  author = {Chopra, Sumit and Hadsell, Raia and LeCun, Yann},
  year = 2005,
  volume = {1},
  pages = {539--546},
  url = {https://api.semanticscholar.org/CorpusID:5555257}
}

@inproceedings{cignoniMeshLabOpensourceMesh2008,
  title = {{{MeshLab}}: An Open-Source Mesh Processing Tool},
  shorttitle = {{{MeshLab}}},
  booktitle = {Eurographics Italian Chapter Conference},
  author = {Cignoni, Paolo and Callieri, Marco and Corsini, Massimiliano and Dellepiane, Matteo and Ganovelli, Fabio and Ranzuglia, Guido},
  year = 2008,
  publisher = {The Eurographics Association}
}

@inproceedings{cruzZillowIndoorDataset2021,
  title = {Zillow Indoor Dataset: {{Annotated}} Floor Plans with 360{$^\circ$} Panoramas and {{3D}} Room Layouts},
  booktitle = {{{CVPR}}},
  author = {Cruz, Steve and Hutchcroft, Will and Li, Yuguang and Khosravan, Naji and Boyadzhiev, Ivaylo and Kang, Sing Bing},
  year = 2021,
  pages = {2133--2143}
}

@inproceedings{daiScanNetRichlyannotated3D2017,
  title = {{{ScanNet}}: {{Richly-annotated 3D}} Reconstructions of Indoor Scenes},
  shorttitle = {{{ScanNetV2}}},
  booktitle = {{{CVPR}}},
  author = {Dai, Angela and Chang, Angel X. and Savva, Manolis and Halber, Maciej and Funkhouser, Thomas and Nie{\ss}ner, Matthias},
  year = 2017,
  pages = {5828--5839}
}

@inproceedings{detoneSuperPointSelfsupervisedInterest2018a,
  title = {{{SuperPoint}}: {{Self-supervised}} Interest Point Detection and Description},
  shorttitle = {{{SuperPoint}}},
  booktitle = {{{CVPRW}}},
  author = {DeTone, Daniel and Malisiewicz, Tomasz and Rabinovich, Andrew},
  year = 2018,
  pages = {224--236},
  publisher = {Computer Vision Foundation / IEEE Computer Society},
  url = {http://openaccess.thecvf.com/content_cvpr_2018_workshops/w9/html/DeTone_SuperPoint_Self-Supervised_Interest_CVPR_2018_paper.html},
  bibsource = {dblp computer science bibliography, https://dblp.org}
}

@inproceedings{dongReloc3rLargeScaleTraining2025,
  title = {Reloc3r: {{Large-Scale Training}} of {{Relative Camera Pose Regression}} for {{Generalizable}}, {{Fast}}, and {{Accurate Visual Localization}}},
  shorttitle = {Reloc3r},
  booktitle = {{{CVPR}}},
  author = {Dong, Siyan and Wang, Shuzhe and Liu, Shaohui and Cai, Lulu and Fan, Qingnan and Kannala, Juho and Yang, Yanchao},
  year = 2025,
  eprint = {2412.08376},
  primaryclass = {cs},
  publisher = {arXiv},
  url = {http://arxiv.org/abs/2412.08376},
  urldate = {2025-04-10},
  archiveprefix = {arXiv}
}

@inproceedings{dongVisualLocalizationFewShot2022,
  title = {Visual {{Localization}} via {{Few-Shot Scene Region Classification}}},
  booktitle = {{{3DV}}},
  author = {Dong, Siyan and Wang, Shuzhe and Zhuang, Yixin and Kannala, Juho and Pollefeys, Marc and Chen, Baoquan},
  year = 2022,
  eprint = {2208.06933},
  primaryclass = {cs},
  publisher = {arXiv},
  url = {http://arxiv.org/abs/2208.06933},
  urldate = {2025-11-06},
  archiveprefix = {arXiv}
}

@inproceedings{dosovitskiyImageWorth16x162021,
  title = {An {{Image}} Is {{Worth}} 16x16 {{Words}}: {{Transformers}} for {{Image Recognition}} at {{Scale}}},
  shorttitle = {{{ViT}}},
  booktitle = {{{ICLR}}},
  author = {Dosovitskiy, Alexey and Beyer, Lucas and Kolesnikov, Alexander and Weissenborn, Dirk and Zhai, Xiaohua and Unterthiner, Thomas and Dehghani, Mostafa and Minderer, Matthias and Heigold, Georg and Gelly, Sylvain and Uszkoreit, Jakob and Houlsby, Neil},
  year = 2021,
  eprint = {2010.11929},
  primaryclass = {cs},
  publisher = {arXiv},
  url = {http://arxiv.org/abs/2010.11929},
  urldate = {2023-05-06},
  archiveprefix = {arXiv}
}

@article{fischlerRandomSampleConsensus1981,
  title = {Random Sample Consensus: A Paradigm for Model Fitting with Applications to Image Analysis and Automated Cartography},
  shorttitle = {{{RANSAC}}},
  author = {Fischler, Martin A. and Bolles, Robert C.},
  year = 1981,
  journal = {Communications of The Acm},
  volume = {24},
  number = {6},
  pages = {381--395},
  publisher = {Association for Computing Machinery},
  url = {https://doi.org/10.1145/358669.358692},
  issue_date = {June 1981}
}

@article{gaoCompleteSolutionClassification2003,
  title = {Complete Solution Classification for the Perspective-Three-Point Problem},
  author = {Gao, Xiaoshan and Hou, Xiaorong and Tang, Jianliang and Cheng, Hangfei},
  year = 2003,
  journal = {IEEE Trans. Pattern Anal. Mach. Intell.},
  volume = {25},
  number = {8},
  pages = {930--943}
}

@inproceedings{gardSPVLocSemanticPanoramic2024,
  title = {{{SPVLoc}}: {{Semantic Panoramic Viewport Matching}} for {{6D Camera Localization}} in {{Unseen Environments}}},
  shorttitle = {{{SPVLoc}}},
  booktitle = {{{ECCV}}},
  author = {Gard, Niklas and Hilsmann, Anna and Eisert, Peter},
  year = 2024,
  volume = {15131},
  pages = {398--415},
  publisher = {Springer Nature Switzerland},
  url = {https://link.springer.com/10.1007/978-3-031-73464-9_24},
  urldate = {2025-10-31}
}

@inproceedings{garlandSurfaceSimplificationUsing1997,
  title = {Surface Simplification Using Quadric Error Metrics},
  booktitle = {{{SIGGRAPH}}},
  author = {Garland, Michael and Heckbert, Paul S.},
  year = 1997,
  pages = {209--216},
  publisher = {ACM Press/Addison-Wesley Publishing Co.},
  url = {https://doi.org/10.1145/258734.258849}
}

@misc{geoscontributorsGEOSComputationalGeometry2025,
  title = {{{GEOS}} Computational Geometry Library},
  author = {{GEOS contributors}},
  year = 2025,
  url = {https://libgeos.org/},
  howpublished = {Open Source Geospatial Foundation}
}

@inproceedings{graderSuperchargingFloorplanLocalization2025a,
  title = {Supercharging Floorplan Localization with Semantic Rays},
  shorttitle = {{{SemRayLoc}}},
  booktitle = {{{ICCV}}},
  author = {Grader, Yuval and {Averbuch-Elor}, Hadar},
  year = 2025,
  eprint = {2507.09291},
  primaryclass = {cs.CV},
  url = {https://arxiv.org/abs/2507.09291},
  archiveprefix = {arXiv}
}

@inproceedings{hartleyProjectiveGeometryTransformations2003,
  title = {Projective {{Geometry}} and {{Transformations}} of {{3D}}},
  booktitle = {Multiple View Geometry in Computer Vision, 2nd},
  author = {Hartley, Richard and Zisserman, Andrew},
  year = 2003,
  pages = {65--86},
  publisher = {Cambridge university press}
}

@inproceedings{heAlphaTabletsGenericPlane2024,
  title = {{{AlphaTablets}}: {{A Generic Plane Representation}} for {{3D Planar Reconstruction}} from {{Monocular Videos}}},
  shorttitle = {{{AlphaTablets}}},
  booktitle = {{{NeurIPS}}},
  author = {He, Yuze and Zhao, Wang and Liu, Shaohui and Hu, Yubin and Bai, Yushi and Wen, Yu-Hui and Liu, Yong-Jin},
  year = 2024,
  eprint = {2411.19950},
  primaryclass = {cs},
  publisher = {arXiv},
  url = {http://arxiv.org/abs/2411.19950},
  urldate = {2025-10-27},
  archiveprefix = {arXiv}
}

@inproceedings{heDeepResidualLearning2016,
  title = {Deep Residual Learning for Image Recognition},
  shorttitle = {{{ResNet50}}},
  booktitle = {{{CVPR}}},
  author = {He, Kaiming and Zhang, Xiangyu and Ren, Shaoqing and Sun, Jian},
  year = 2016,
  pages = {770--778}
}

@inproceedings{heMaskRCNN2017,
  title = {Mask R-{{CNN}}},
  booktitle = {{{ICCV}}},
  author = {He, Kaiming and Gkioxari, Georgia and Doll{\'a}r, Piotr and Girshick, Ross},
  year = 2017,
  pages = {2980--2988},
  url = {https://arxiv.org/abs/1703.06870}
}

@misc{heMatchAnythingUniversalCrossModality2025,
  title = {{{MatchAnything}}: {{Universal Cross-Modality Image Matching}} with {{Large-Scale Pre-Training}}},
  shorttitle = {{{MatchAnything}}},
  author = {He, Xingyi and Yu, Hao and Peng, Sida and Tan, Dongli and Shen, Zehong and Bao, Hujun and Zhou, Xiaowei},
  year = 2025,
  number = {arXiv:2501.07556},
  eprint = {2501.07556},
  primaryclass = {cs},
  publisher = {arXiv},
  url = {http://arxiv.org/abs/2501.07556},
  urldate = {2025-01-14},
  archiveprefix = {arXiv}
}

@article{hornClosedformSolutionAbsolute1987,
  title = {Closed-Form Solution of Absolute Orientation Using Unit Quaternions},
  author = {Horn, Berthold KP},
  year = 1987,
  journal = {Journal of the optical society of America A},
  volume = {4},
  number = {4},
  pages = {629--642},
  publisher = {Optical Society of America}
}

@inproceedings{howard-jenkinsLalalocGlobalFloor2022,
  title = {Lalaloc++: {{Global}} Floor Plan Comprehension for Layout Localisation in Unvisited Environments},
  shorttitle = {Lalaloc++},
  booktitle = {{{ECCV}}},
  author = {{Howard-Jenkins}, Henry and Prisacariu, Victor Adrian},
  year = 2022,
  pages = {693--709},
  publisher = {Springer}
}

@inproceedings{howard-jenkinsLalalocLatentLayout2021,
  title = {Lalaloc: {{Latent}} Layout Localisation in Dynamic, Unvisited Environments},
  shorttitle = {Lalaloc},
  booktitle = {{{ICCV}}},
  author = {{Howard-Jenkins}, Henry and {Ruiz-Sarmiento}, Jose-Raul and Prisacariu, Victor Adrian},
  year = 2021,
  pages = {10107--10116}
}

@inproceedings{hrubyEfficientSolutionPointline2024,
  title = {Efficient Solution of Point-Line Absolute Pose},
  booktitle = {{{CVPR}}},
  author = {Hruby, Petr and Duff, Timothy and Pollefeys, Marc},
  year = 2024,
  pages = {21316--21325},
  url = {https://api.semanticscholar.org/CorpusID:269362112}
}

@inproceedings{huangSparseDenseCamera2025,
  title = {From Sparse to Dense: {{Camera}} Relocalization with Scene-Specific Detector from {{Feature Gaussian Splatting}}},
  shorttitle = {{{STDLoc}}},
  booktitle = {{{CVPR}}},
  author = {Huang, Zhiwei and Yu, Hailin and Shentu, Yichun and Yuan, Jin and Zhang, Guofeng},
  year = 2025
}

@misc{humenbergerRobustImageRetrievalbased2022,
  title = {Robust {{Image Retrieval-based Visual Localization}} Using {{Kapture}}},
  author = {Humenberger, Martin and Cabon, Yohann and Guerin, Nicolas and Morat, Julien and Leroy, Vincent and Revaud, J{\'e}r{\^o}me and Rerole, Philippe and Pion, No{\'e} and de Souza, Cesar and Csurka, Gabriela},
  year = 2022,
  number = {arXiv:2007.13867},
  eprint = {2007.13867},
  primaryclass = {cs},
  publisher = {arXiv},
  url = {http://arxiv.org/abs/2007.13867},
  urldate = {2026-03-19},
  archiveprefix = {arXiv}
}

@inproceedings{jiangRSCoReRevisitingScene2025,
  title = {R-{{SCoRe}}: {{Revisiting Scene Coordinate Regression}} for {{Robust Large-Scale Visual Localization}}},
  shorttitle = {R-{{SCoRe}}},
  booktitle = {{{CVPR}}},
  author = {Jiang, Xudong and Wang, Fangjinhua and Galliani, Silvano and Vogel, Christoph and Pollefeys, Marc},
  year = 2025,
  eprint = {2501.01421},
  primaryclass = {cs},
  publisher = {arXiv},
  url = {http://arxiv.org/abs/2501.01421},
  urldate = {2025-01-05},
  archiveprefix = {arXiv}
}

@inproceedings{jinPlanarSurfaceReconstruction2021,
  title = {Planar Surface Reconstruction from Sparse Views},
  shorttitle = {{{SparsePlane}}},
  booktitle = {{{ICCV}}},
  author = {Jin, Linyi and Qian, Shengyi and Owens, Andrew and Fouhey, David F.},
  year = 2021,
  pages = {12971--12980},
  publisher = {IEEE},
  url = {https://doi.org/10.1109/ICCV48922.2021.01275}
}

@inproceedings{juelinzhuLoDlocVisualLocalization2024,
  title = {{{LoD-loc}}: {{Visual}} Localization Using {{LoD 3D}} Map with Neural Wireframe Alignment},
  shorttitle = {{{LoD-Loc}}},
  booktitle = {{{NeurIPS}}},
  author = {Juelin Zhu, Shen Yan, Long Wang and Zhang, Maojun},
  year = 2024
}

@article{kabschSolutionBestRotation1976,
  title = {A Solution for the Best Rotation to Relate Two Sets of Vectors},
  shorttitle = {Kabsch},
  author = {Kabsch, Wolfgang},
  year = 1976,
  journal = {Foundations of Crystallography},
  volume = {32},
  number = {5},
  pages = {922--923},
  publisher = {International Union of Crystallography}
}

@inproceedings{kendallGeometricLossFunctions2017,
  title = {Geometric {{Loss Functions}} for {{Camera Pose Regression}} with {{Deep Learning}}},
  booktitle = {2017 {{IEEE Conference}} on {{Computer Vision}} and {{Pattern Recognition}}},
  author = {Kendall, Alex and Cipolla, Roberto},
  year = 2017,
  pages = {6555--6564},
  publisher = {IEEE}
}

@inproceedings{keRepurposingDiffusionbasedImage2024,
  title = {Repurposing Diffusion-Based Image Generators for Monocular Depth Estimation},
  shorttitle = {Marigold},
  booktitle = {{{CVPR}}},
  author = {Ke, Bingxin and Obukhov, Anton and Huang, Shengyu and Metzger, Nando and Daudt, Rodrigo Caye and Schindler, Konrad},
  year = 2024,
  pages = {9492--9502}
}

@inproceedings{khoslaRELOCATESimpleTrainingfree2025,
  title = {{{RELOCATE}}: A Simple Training-Free Baseline for Visual Query Localization Using Region-Based Representations},
  shorttitle = {{{RELOCATE}}},
  booktitle = {{{CVPR}}},
  author = {Khosla, Savya and V, Sethuraman T and Schwing, Alexander and Hoiem, Derek},
  year = 2025,
  pages = {3697--3706}
}

@inproceedings{khoslaSupervisedContrastiveLearning2020,
  title = {Supervised Contrastive Learning},
  booktitle = {{{NeurIPS}}},
  author = {Khosla, Prannay and Teterwak, Piotr and Wang, Chen and Sarna, Aaron and Tian, Yonglong and Isola, Phillip and Maschinot, Aaron and Liu, Ce and Krishnan, Dilip},
  year = 2020,
  volume = {33},
  pages = {18661--18673}
}

@inproceedings{kingmaAdamMethodStochastic2015,
  title = {Adam: {{A Method}} for {{Stochastic Optimization}}},
  shorttitle = {Adam},
  booktitle = {{{ICLR}}},
  author = {Kingma, Diederik P. and Ba, Jimmy},
  year = 2015,
  eprint = {1412.6980},
  primaryclass = {cs},
  publisher = {arXiv},
  url = {http://arxiv.org/abs/1412.6980},
  urldate = {2025-11-18},
  archiveprefix = {arXiv}
}

@misc{larssonPoseLibMinimalSolvers2020,
  title = {{{PoseLib}} - Minimal Solvers for Camera Pose Estimation},
  author = {Larsson, Viktor and {contributors}},
  year = 2020,
  url = {https://github.com/vlarsson/PoseLib}
}

@inproceedings{leroyGroundingImageMatching2024,
  title = {Grounding {{Image Matching}} in {{3D}} with {{MASt3R}}},
  shorttitle = {{{MASt3R}}},
  booktitle = {{{ECCV}}},
  author = {Leroy, Vincent and Cabon, Yohann and Revaud, J{\'e}r{\^o}me},
  year = 2024,
  eprint = {2406.09756},
  publisher = {arXiv},
  url = {http://arxiv.org/abs/2406.09756},
  urldate = {2024-10-15},
  archiveprefix = {arXiv}
}

@inproceedings{li2D3DMATR2D3DMatching2023,
  title = {{{2D3D-MATR}}: {{2D-3D Matching Transformer}} for {{Detection-free Registration}} between {{Images}} and {{Point Clouds}}},
  shorttitle = {{{2D3D-MATR}}},
  booktitle = {{{ICCV}}},
  author = {Li, Minhao and Qin, Zheng and Gao, Zhirui and Yi, Renjiao and Zhu, Chenyang and Guo, Yulan and Xu, Kai},
  year = 2023,
  pages = {14082--14092},
  publisher = {IEEE},
  url = {https://ieeexplore.ieee.org/document/10378633/},
  urldate = {2025-03-01},
  copyright = {https://doi.org/10.15223/policy-029}
}

@inproceedings{liActLocLearningLocalize2025,
  title = {{{ActLoc}}: {{Learning}} to {{Localize}} on the {{Move}} via {{Active Viewpoint Selection}}},
  shorttitle = {{{ActLoc}}},
  booktitle = {{{CoRL}}},
  author = {Li, Jiajie and Sun, Boyang and Giammarino, Luca Di and Blum, Hermann and Pollefeys, Marc},
  year = 2025,
  eprint = {2508.20981},
  primaryclass = {cs},
  publisher = {arXiv},
  url = {http://arxiv.org/abs/2508.20981},
  urldate = {2025-10-31},
  archiveprefix = {arXiv}
}

@inproceedings{liLearnableFourierFeatures2021,
  title = {Learnable Fourier Features for Multi-Dimensional Spatial Positional Encoding},
  booktitle = {{{NeurIPS}}},
  author = {Li, Yang and Si, Si and Li, Gang and Hsieh, Cho-Jui and Bengio, Samy},
  year = 2021,
  publisher = {Curran Associates Inc.},
  articleno = {1210}
}

@inproceedings{liMatchingAnythingSegmenting2024,
  title = {Matching Anything by Segmenting Anything},
  shorttitle = {{{MASA}}},
  booktitle = {{{CVPR}}},
  author = {Li, Siyuan and Ke, Lei and Danelljan, Martin and Piccinelli, Luigi and Seg{\`u}, Mattia and Gool, Luc Van and Yu, Fisher},
  year = 2024,
  pages = {18963--18973},
  publisher = {IEEE},
  url = {https://doi.org/10.1109/CVPR52733.2024.01794},
  bibsource = {dblp computer science bibliography, https://dblp.org}
}

@inproceedings{linBARFBundleadjustingNeural2021,
  title = {{{BARF}}: {{Bundle-adjusting}} Neural Radiance Fields},
  shorttitle = {{{BARF}}},
  booktitle = {{{ICCV}}},
  author = {Lin, Chen Hsuan and Ma, Wei Chiu and Torralba, Antonio and Lucey, Simon},
  year = 2021,
  pages = {5721--5731},
  publisher = {IEEE}
}

@inproceedings{lindenbergerLightGlueLocalFeature2023,
  title = {{{LightGlue}}: {{Local}} Feature Matching at Light Speed},
  shorttitle = {{{LightGlue}}},
  booktitle = {{{ICCV}}},
  author = {Lindenberger, Philipp and Sarlin, Paul-Edouard and Pollefeys, Marc},
  year = 2023,
  pages = {17581--17592},
  publisher = {IEEE},
  url = {https://doi.org/10.1109/ICCV51070.2023.01616},
  bibsource = {dblp computer science bibliography, https://dblp.org}
}

@inproceedings{liu3DLineMapping2023,
  title = {{{3D}} Line Mapping Revisited},
  shorttitle = {{{LIMAP}}},
  booktitle = {{{CVPR}}},
  author = {Liu, Shaohui and Yu, Yifan and Pautrat, R{\'e}mi and Pollefeys, Marc and Larsson, Viktor},
  year = 2023,
  pages = {21445--21455},
  publisher = {IEEE}
}

@inproceedings{liuBWFormerBuildingWireframe2025,
  title = {{{BWFormer}}: {{Building}} Wireframe Reconstruction from Airborne {{LiDAR}} Point Cloud with Transformer},
  shorttitle = {{{BWFormer}}},
  booktitle = {{{CVPR}}},
  author = {Liu, Yuzhou and Zhu, Lingjie and Ye, Hanqiao and Huang, Shangfeng and Gao, Xiang and Zheng, Xianwei and Shen, Shuhan},
  year = 2025,
  pages = {22215--22224}
}

@inproceedings{liuEfficientGlobal2d3d2017,
  title = {Efficient Global 2d-3d Matching for Camera Localization in a Large-Scale 3d Map},
  booktitle = {{{ICCV}}},
  author = {Liu, Liu and Li, Hongdong and Dai, Yuchao},
  year = 2017,
  pages = {2372--2381}
}

@inproceedings{liuGSCPREfficientCamera2025,
  title = {{{GS-CPR}}: {{Efficient}} Camera Pose Refinement via {{3D}} Gaussian Splatting},
  shorttitle = {{{GS-CPR}}},
  booktitle = {{{ICLR}}},
  author = {Liu, Changkun and Chen, Shuai and Bhalgat, Yash Sanjay and HU, Siyan and Cheng, Ming and Wang, Zirui and Prisacariu, Victor Adrian and Braud, Tristan},
  year = 2025,
  url = {https://openreview.net/forum?id=mP7uV59iJM}
}

@inproceedings{liuInthewild3DPlane2025,
  title = {Towards {{In-the-wild 3D Plane Reconstruction}} from a {{Single Image}}},
  shorttitle = {Zeroplane},
  booktitle = {{{CVPR}}},
  author = {Liu, Jiachen and Yu, Rui and Chen, Sili and Huang, Sharon X. and Guo, Hengkai},
  year = 2025,
  eprint = {2506.02493},
  primaryclass = {cs},
  publisher = {arXiv},
  url = {http://arxiv.org/abs/2506.02493},
  urldate = {2025-06-19},
  archiveprefix = {arXiv}
}

@article{liuLightweightStructuredLine2024,
  title = {Lightweight Structured Line Map Based Visual Localization},
  shorttitle = {{{LSLM}}\_{{VLoc}}},
  author = {Liu, Hongmin and Cao, Chengyang and Ye, Hanqiao and Cui, Hainan and Gao, Wei and Wang, Xing and Shen, Shuhan},
  year = 2024,
  journal = {IEEE Robotics and Automation Letters},
  volume = {9},
  number = {6},
  pages = {5182--5189},
  copyright = {All rights reserved}
}

@inproceedings{liuPLANA3RZeroshotMetric2025,
  title = {{{PLANA3R}}: {{Zero-shot Metric Planar 3D Reconstruction}} via {{Feed-Forward Planar Splatting}}},
  shorttitle = {{{PLANA3R}}},
  booktitle = {{{NeurIPS}}},
  author = {Liu, Changkun and Tan, Bin and Ke, Zeran and Zhang, Shangzhan and Liu, Jiachen and Qian, Ming and Xue, Nan and Shen, Yujun and Braud, Tristan},
  year = 2025,
  eprint = {2510.18714},
  primaryclass = {cs},
  publisher = {arXiv},
  url = {http://arxiv.org/abs/2510.18714},
  urldate = {2025-10-22},
  archiveprefix = {arXiv}
}

@inproceedings{liuPlaneMVS3DPlane2022,
  title = {{{PlaneMVS}}: {{3D}} Plane Reconstruction from Multi-View Stereo},
  shorttitle = {{{PlaneMVS}}},
  booktitle = {{{CVPR}}},
  author = {Liu, Jiacheng and Ji, Pan and Bansal, Nitin and Cai, Changjiang and Yan, Qingan and Huang, Xiaolei and Xu, Yi},
  year = 2022,
  pages = {8655--8665},
  url = {https://api.semanticscholar.org/CorpusID:247619023}
}

@inproceedings{liuPlaneNetPiecewisePlanar2018,
  title = {{{PlaneNet}}: {{Piece-wise}} Planar Reconstruction from a Single {{RGB}} Image},
  shorttitle = {{{PlaneNet}}},
  booktitle = {{{CVPR}}},
  author = {Liu, Chen and Yang, Jimei and Ceylan, Duygu and Yumer, Ersin and Furukawa, Yasutaka},
  year = 2018,
  pages = {2579--2588}
}

@inproceedings{liuPlaneRCNN3DPlane2019,
  title = {{{PlaneRCNN}}: {{3D}} Plane Detection and Reconstruction from a Single Image},
  shorttitle = {{{PlaneRCNN}}},
  booktitle = {{{CVPR}}},
  author = {Liu, Chen and Kim, Kihwan and Gu, Jinwei and Furukawa, Yasutaka and Kautz, Jan},
  year = 2019,
  pages = {4445--4454},
  publisher = {IEEE},
  url = {https://ieeexplore.ieee.org/document/8953257/},
  urldate = {2024-03-03}
}

@inproceedings{liuPolyRoomRoomawareTransformer2024,
  title = {{{PolyRoom}}: {{Room-aware}} Transformer for Floorplan Reconstruction},
  shorttitle = {{{PolyRoom}}},
  booktitle = {{{ECCV}}},
  author = {Liu, Yuzhou and Zhu, Lingjie and Ma, Xiaodong and Ye, Hanqiao and Gao, Xiang and Zheng, Xianwei and Shen, Shuhan},
  year = 2024,
  volume = {15108},
  pages = {322--339},
  publisher = {Springer},
  url = {https://doi.org/10.1007/978-3-031-72973-7_19},
  bibsource = {dblp computer science bibliography, https://dblp.org}
}

@inproceedings{loshchilovDecoupledWeightDecay2019,
  title = {Decoupled Weight Decay Regularization},
  shorttitle = {{{AdamW}}},
  booktitle = {{{ICLR}}},
  author = {Loshchilov, Ilya and Hutter, Frank},
  year = 2019,
  eprint = {1711.05101},
  primaryclass = {cs.LG},
  url = {https://arxiv.org/abs/1711.05101},
  archiveprefix = {arXiv}
}

@inproceedings{mazurSuperPrimitiveSceneReconstruction2024,
  title = {{{SuperPrimitive}}: {{Scene Reconstruction}} at a {{Primitive Level}}},
  shorttitle = {{{SuperPrimitive}}},
  booktitle = {{{CVPR}}},
  author = {Mazur, Kirill and Bae, Gwangbin and Davison, Andrew J.},
  year = 2024,
  pages = {4979--4989},
  url = {https://arxiv.org/abs/2312.05889v1},
  urldate = {2024-03-07}
}

@inproceedings{miaoSceneGraphLocCrossModalCoarse2024,
  title = {{{SceneGraphLoc}}: {{Cross-Modal Coarse Visual Localization}} on {{3D Scene Graphs}}},
  shorttitle = {{{SceneGraphLoc}}},
  booktitle = {{{ECCV}}},
  author = {Miao, Yang and Engelmann, Francis and Vysotska, Olga and Tombari, Federico and Pollefeys, Marc and Bar{\'a}th, D{\'a}niel B{\'e}la},
  year = 2024,
  publisher = {arXiv},
  url = {https://arxiv.org/abs/2404.00469},
  urldate = {2025-02-28},
  copyright = {arXiv.org perpetual, non-exclusive license}
}

@article{monszpartRAPterRebuildingManmade2015a,
  title = {{{RAPter}}: Rebuilding Man-Made Scenes with Regular Arrangements of Planes},
  shorttitle = {{{RAPter}}},
  author = {Monszpart, Aron and Mellado, Nicolas and Brostow, Gabriel J. and Mitra, Niloy J.},
  year = 2015,
  journal = {ACM Trans. Graph.},
  volume = {34},
  number = {4},
  publisher = {Association for Computing Machinery},
  url = {https://doi.org/10.1145/2766995},
  articleno = {103},
  issue_date = {August 2015}
}

@inproceedings{moreauCROSSFIRECameraRelocalization2023,
  title = {{{CROSSFIRE}}: {{Camera}} Relocalization on Self-Supervised Features from an Implicit Representation},
  shorttitle = {{{CROSSFIRE}}},
  booktitle = {{{ICCV}}},
  author = {Moreau, Arthur and Piasco, Nathan and Bennehar, Moussab and Tsishkou, Dzmitry and Stanciulescu, Bogdan and {de La Fortelle}, Arnaud},
  year = 2023,
  pages = {252--262},
  publisher = {IEEE}
}

@inproceedings{muDiff$^2$I2PDifferentiableImagetoPoint2025,
  title = {Diff$^\text{2}${{I2P}}: {{Differentiable Image-to-Point Cloud Registration}} with {{Diffusion Prior}}},
  shorttitle = {Diff\$\textasciicircum 2\${{I2P}}},
  booktitle = {{{ICCV}}},
  author = {Mu, Juncheng and Ren, Chengwei and Zhang, Weixiang and Pan, Liang and Zhang, Xiao-Ping and Gao, Yue},
  year = 2025,
  eprint = {2507.06651},
  primaryclass = {cs},
  publisher = {arXiv},
  url = {http://arxiv.org/abs/2507.06651},
  urldate = {2025-07-13},
  archiveprefix = {arXiv}
}

@inproceedings{nanPolyFitPolygonalSurface2017,
  title = {{{PolyFit}}: {{Polygonal}} Surface Reconstruction from Point Clouds},
  booktitle = {{{ICCV}}},
  author = {Nan, Liangliang and Wonka, Peter},
  year = 2017,
  pages = {2372--2380},
  publisher = {IEEE Computer Society},
  url = {https://doi.org/10.1109/ICCV.2017.258}
}

@article{oquabDINOv2LearningRobust2023,
  title = {{{DINOv2}}: {{Learning Robust Visual Features}} without {{Supervision}}},
  shorttitle = {{{DINOv2}}},
  author = {Oquab, Maxime and Darcet, Timoth{\'e}e and Moutakanni, Th{\'e}o and Vo, Huy and Szafraniec, Marc and Khalidov, Vasil and Fernandez, Pierre and Haziza, Daniel and Massa, Francisco and {El-Nouby}, Alaaeldin and Assran, Mahmoud and Ballas, Nicolas and Galuba, Wojciech and Howes, Russell and Huang, Po-Yao and Li, Shang-Wen and Misra, Ishan and Rabbat, Michael and Sharma, Vasu and Synnaeve, Gabriel and Xu, Hu and Jegou, Herv{\'e} and Mairal, Julien and Labatut, Patrick and Joulin, Armand and Bojanowski, Piotr},
  year = 2024,
  journal = {TMLR},
  eprint = {2304.07193},
  primaryclass = {cs},
  pages = {2835--8856},
  url = {https://openreview.net/forum?id=a68SUt6zFt},
  urldate = {2024-01-30},
  archiveprefix = {arXiv}
}

@inproceedings{panekMeshLocMeshbasedVisual2022,
  title = {{{MeshLoc}}: {{Mesh-based}} Visual Localization},
  shorttitle = {{{MeshLoc}}},
  booktitle = {{{ECCV}}},
  author = {Panek, Vojtech and Kukelova, Zuzana and Sattler, Torsten},
  year = 2022
}

@inproceedings{panekVisualLocalizationUsing2023,
  title = {Visual {{Localization}} Using {{Imperfect 3D Models}} from the {{Internet}}},
  shorttitle = {Cadloc},
  booktitle = {{{ICCV}}},
  author = {Panek, Vojtech and Kukelova, Zuzana and Sattler, Torsten},
  year = 2023,
  eprint = {2304.05947},
  primaryclass = {cs},
  publisher = {arXiv},
  url = {http://arxiv.org/abs/2304.05947},
  urldate = {2025-05-28},
  archiveprefix = {arXiv}
}

@inproceedings{pautratGlueStickRobustImage2023,
  title = {{{GlueStick}}: {{Robust Image Matching}} by {{Sticking Points}} and {{Lines Together}}},
  shorttitle = {{{GlueStick}}},
  booktitle = {{{ICCV}}},
  author = {Pautrat, R{\'e}mi and Su{\'a}rez, Iago and Yu, Yifan and Pollefeys, Marc and Larsson, Viktor},
  year = 2023,
  eprint = {2304.02008},
  primaryclass = {cs},
  publisher = {arXiv},
  url = {http://arxiv.org/abs/2304.02008},
  urldate = {2025-05-31},
  archiveprefix = {arXiv}
}

@inproceedings{pietrantoniGaussianSplattingFeature2025,
  title = {Gaussian {{Splatting Feature Fields}} for {{Privacy-Preserving Visual Localization}}},
  booktitle = {{{CVPR}}},
  author = {Pietrantoni, Maxime and Csurka, Gabriela and Sattler, Torsten},
  year = 2025,
  eprint = {2507.23569},
  primaryclass = {cs},
  publisher = {arXiv},
  url = {http://arxiv.org/abs/2507.23569},
  urldate = {2025-11-06},
  archiveprefix = {arXiv}
}

@article{qinGeoTransformerFastRobust2023,
  title = {{{GeoTransformer}}: {{Fast}} and {{Robust Point Cloud Registration}} with {{Geometric Transformer}}},
  shorttitle = {{{GeoTransformer}}},
  author = {Qin, Zheng and Yu, Hao and Wang, Changjian and Guo, Yulan and Peng, Yuxing and Ilic, Slobodan and Hu, Dewen and Xu, Kai},
  year = 2023,
  journal = {IEEE Trans. Pattern Anal. Mach. Intell.},
  volume = {45},
  number = {8},
  eprint = {2308.03768},
  primaryclass = {cs},
  pages = {9806--9821},
  url = {http://arxiv.org/abs/2308.03768},
  urldate = {2025-06-02},
  archiveprefix = {arXiv}
}

@inproceedings{qiPointNetDeepLearning2017,
  title = {{{PointNet}}: {{Deep Learning}} on {{Point Sets}} for {{3D Classification}} and {{Segmentation}}},
  shorttitle = {{{PointNet}}},
  booktitle = {{{CVPR}}},
  author = {Qi, Charles R. and Su, Hao and Mo, Kaichun and Guibas, Leonidas J.},
  year = 2017,
  eprint = {1612.00593},
  primaryclass = {cs},
  publisher = {arXiv},
  url = {http://arxiv.org/abs/1612.00593},
  urldate = {2023-05-06},
  archiveprefix = {arXiv}
}

@inproceedings{radfordLearningTransferableVisual2021,
  title = {Learning {{Transferable Visual Models From Natural Language Supervision}}},
  shorttitle = {{{CLIP}}},
  booktitle = {{{ICML}}},
  author = {Radford, Alec and Kim, Jong Wook and Hallacy, Chris and Ramesh, Aditya and Goh, Gabriel and Agarwal, Sandhini and Sastry, Girish and Askell, Amanda and Mishkin, Pamela and Clark, Jack and Krueger, Gretchen and Sutskever, Ilya},
  year = 2021,
  eprint = {2103.00020},
  primaryclass = {cs},
  publisher = {arXiv},
  url = {http://arxiv.org/abs/2103.00020},
  urldate = {2023-05-28},
  archiveprefix = {arXiv}
}

@inproceedings{ramalingamPoseEstimationUsing2011,
  title = {Pose Estimation Using Both Points and Lines for Geo-Localization},
  booktitle = {{{ICRA}}},
  author = {Ramalingam, Srikumar and Bouaziz, Sofien and Sturm, Peter F.},
  year = 2011,
  pages = {4716--4723},
  url = {https://api.semanticscholar.org/CorpusID:2167641}
}

@article{raposoPiecewiseplanarStereoScanSequential2017,
  title = {Piecewise-Planar {{StereoScan}}: {{Sequential}} Structure and Motion Using Plane Primitives},
  author = {Raposo, Carolina and Antunes, Michel and Barreto, Jo{\~a}o P.},
  year = 2017,
  journal = {IEEE Trans. Pattern Anal. Mach. Intell.},
  volume = {40},
  number = {8},
  pages = {1918--1931}
}

@inproceedings{raposoPlanebasedOdometryUsing2013,
  title = {Plane-Based Odometry Using an {{RGB-d}} Camera.},
  booktitle = {{{BMVC}}},
  author = {Raposo, Carolina and Louren{\c c}o, Miguel and Antunes, Michel and Barreto, Joao Pedro},
  year = 2013,
  volume = {2},
  pages = {6}
}

@inproceedings{sarkarCrossOver3DScene2025,
  title = {{{CrossOver}}: {{3D}} Scene Cross-Modal Alignment},
  shorttitle = {{{CrossOver}}},
  booktitle = {{{CVPR}}},
  author = {Sarkar, Sayan Deb and Miksik, Ondrej and Pollefeys, Marc and Barath, Daniel and Armeni, Iro},
  year = 2025
}

@inproceedings{sarkarSGAligner3DScene2023,
  title = {{{SGAligner}}: {{3D Scene Alignment}} with {{Scene Graphs}}},
  shorttitle = {{{SGAligner}}},
  booktitle = {2023 {{IEEE}}/{{CVF International Conference}} on {{Computer Vision}} ({{ICCV}})},
  author = {Sarkar, Sayan Deb and Miksik, Ondrej and Pollefeys, Marc and Barath, Daniel and Armeni, Iro},
  year = 2023,
  pages = {21870--21880},
  publisher = {IEEE},
  url = {https://ieeexplore.ieee.org/document/10378487/},
  urldate = {2025-02-28},
  copyright = {https://doi.org/10.15223/policy-029}
}

@inproceedings{sarlinCoarseFineRobust2019,
  title = {From {{Coarse}} to {{Fine}}: {{Robust Hierarchical Localization}} at {{Large Scale}}},
  shorttitle = {{{hLoc}}},
  booktitle = {{{CVPR}}},
  author = {Sarlin, Paul-Edouard and Cadena, Cesar and Siegwart, Roland and Dymczyk, Marcin},
  year = 2019,
  pages = {12708--12717},
  publisher = {IEEE}
}

@inproceedings{sarlinLaMARBenchmarkingLocalization2022,
  title = {{{LaMAR}}: {{Benchmarking Localization}} and~{{Mapping}} for~{{Augmented Reality}}},
  shorttitle = {{{LaMAR}}},
  booktitle = {{{ECCV}}},
  author = {Sarlin, Paul-Edouard and Dusmanu, Mihai and Sch{\"o}nberger, Johannes L. and Speciale, Pablo and Gruber, Lukas and Larsson, Viktor and Miksik, Ondrej and Pollefeys, Marc},
  year = 2022,
  pages = {686--704},
  publisher = {Springer Nature Switzerland}
}

@inproceedings{sarlinSuperGlueLearningFeature2020,
  title = {{{SuperGlue}}: {{Learning Feature Matching With Graph Neural Networks}}},
  shorttitle = {{{SuperGlue}}},
  booktitle = {{{CVPR}}},
  author = {Sarlin, Paul-Edouard and DeTone, Daniel and Malisiewicz, Tomasz and Rabinovich, Andrew},
  year = 2020,
  pages = {4937--4946},
  publisher = {IEEE},
  url = {https://ieeexplore.ieee.org/document/9157489/},
  urldate = {2023-08-28}
}

@article{sattlerEfficientEffectivePrioritized2016,
  title = {Efficient \& Effective Prioritized Matching for Large-Scale Image-Based Localization},
  author = {Sattler, Torsten and Leibe, Bastian and Kobbelt, Leif},
  year = 2016,
  journal = {IEEE Trans. Pattern Anal. Mach. Intell.},
  volume = {39},
  number = {9},
  pages = {1744--1756},
  publisher = {IEEE}
}

@inproceedings{sattlerImprovingImagebasedLocalization2012,
  title = {Improving Image-Based Localization by Active Correspondence Search},
  booktitle = {{{ECCV}}},
  author = {Sattler, Torsten and Leibe, Bastian and Kobbelt, Leif},
  year = 2012,
  pages = {752--765},
  publisher = {Springer}
}

@inproceedings{schonbergerSemanticVisualLocalization2018,
  title = {Semantic {{Visual Localization}}},
  booktitle = {{{CVPR}}},
  author = {Sch{\"o}nberger, Johannes L. and Pollefeys, Marc and Geiger, Andreas and Sattler, Torsten},
  year = 2018,
  eprint = {1712.05773},
  primaryclass = {cs},
  publisher = {arXiv},
  url = {http://arxiv.org/abs/1712.05773},
  urldate = {2025-11-06},
  archiveprefix = {arXiv}
}

@inproceedings{schonbergerStructurefrommotionRevisited2016,
  title = {Structure-from-Motion Revisited},
  shorttitle = {{{COLMAP}}},
  booktitle = {{{CVPR}}},
  author = {Sch{\"o}nberger, Johannes Lutz and Frahm, Jan-Michael},
  year = 2016,
  pages = {4104--4113}
}

@inproceedings{shiPlaneMatchPatchCoplanarity2018,
  title = {{{PlaneMatch}}: {{Patch Coplanarity Prediction}} for {{Robust RGB-D Reconstruction}}},
  shorttitle = {{{PlaneMatch}}},
  booktitle = {{{ECCV}}},
  author = {Shi, Yifei and Xu, Kai and Niessner, Matthias and Rusinkiewicz, Szymon and Funkhouser, Thomas},
  year = 2018,
  eprint = {1803.08407},
  primaryclass = {cs},
  publisher = {arXiv},
  url = {http://arxiv.org/abs/1803.08407},
  urldate = {2025-08-01},
  archiveprefix = {arXiv}
}

@inproceedings{shiPlaneRecTRUnifiedQuery2023,
  title = {{{PlaneRecTR}}: {{Unified Query Learning}} for {{3D Plane Recovery}} from a {{Single View}}},
  shorttitle = {{{PlaneRecTR}}},
  booktitle = {{{ICCV}}},
  author = {Shi, Jingjia and Zhi, Shuaifeng and Xu, Kai},
  year = 2023,
  pages = {9377--9386},
  urldate = {2024-05-06}
}

@article{shiPlaneRecTRUnifiedQuery2025,
  title = {{{PlaneRecTR}}++: {{Unified Query Learning}} for {{Joint 3D Planar Reconstruction}} and {{Pose Estimation}}},
  shorttitle = {{{PlaneRecTR}}++},
  author = {Shi, Jingjia and Zhi, Shuaifeng and Xu, Kai},
  year = 2025,
  journal = {IEEE Trans. Pattern Anal. Mach. Intell.},
  eprint = {2307.13756},
  primaryclass = {cs},
  url = {http://arxiv.org/abs/2307.13756},
  urldate = {2025-03-01},
  archiveprefix = {arXiv}
}

@inproceedings{shlapentokh-rothmanRegionBasedRepresentationsRevisited2024,
  title = {Region-{{Based Representations Revisited}}},
  booktitle = {{{CVPR}}},
  author = {{Shlapentokh-Rothman}, Michal and Blume, Ansel and Xiao, Yao and Wu, Yuqun and V, Sethuraman T. and Tao, Heyi and Lee, Jae Yong and Torres, Wilfredo and Wang, Yu-Xiong and Hoiem, Derek},
  year = 2024,
  eprint = {2402.02352},
  primaryclass = {cs},
  publisher = {arXiv},
  url = {http://arxiv.org/abs/2402.02352},
  urldate = {2025-09-19},
  archiveprefix = {arXiv}
}

@inproceedings{shottonSceneCoordinateRegression2013,
  title = {Scene Coordinate Regression Forests for Camera Relocalization in {{RGB-d}} Images},
  booktitle = {{{CVPR}}},
  author = {Shotton, Jamie and Glocker, Ben and Zach, Christopher and Izadi, Shahram and Criminisi, Antonio and Fitzgibbon, Andrew},
  year = 2013,
  pages = {2930--2937}
}

@article{sommerPlanesCornersMultiPurpose2020,
  title = {From {{Planes}} to {{Corners}}: {{Multi-Purpose Primitive Detection}} in {{Unorganized 3D Point Clouds}}},
  shorttitle = {From {{Planes}} to {{Corners}}},
  author = {Sommer, Christiane and Sun, Yumin and Guibas, Leonidas and Cremers, Daniel and Birdal, Tolga},
  year = 2020,
  journal = {IEEE Robot. Autom. Lett.},
  volume = {5},
  number = {2},
  pages = {1764--1771},
  url = {https://ieeexplore.ieee.org/document/8972549/},
  urldate = {2025-11-06},
  copyright = {https://ieeexplore.ieee.org/Xplorehelp/downloads/license-information/IEEE.html}
}

@inproceedings{sunLoFTRDetectorFreeLocal2021,
  title = {{{LoFTR}}: {{Detector-Free Local Feature Matching}} with {{Transformers}}},
  shorttitle = {{{LoFTR}}},
  booktitle = {{{CVPR}}},
  author = {Sun, Jiaming and Shen, Zehong and Wang, Yuang and Bao, Hujun and Zhou, Xiaowei},
  year = 2021,
  pages = {8918--8927},
  publisher = {IEEE},
  url = {https://ieeexplore.ieee.org/document/9578008/},
  urldate = {2023-08-28}
}

@article{suRoFormerEnhancedTransformer2023,
  title = {{{RoFormer}}: {{Enhanced Transformer}} with {{Rotary Position Embedding}}},
  shorttitle = {{{RoFormer}}},
  author = {Su, Jianlin and Lu, Yu and Pan, Shengfeng and Murtadha, Ahmed and Wen, Bo and Liu, Yunfeng},
  year = 2024,
  journal = {Neurocomput.},
  eprint = {2104.09864},
  primaryclass = {cs},
  publisher = {arXiv},
  url = {http://arxiv.org/abs/2104.09864},
  urldate = {2025-10-16},
  archiveprefix = {arXiv}
}

@inproceedings{tairaInLocIndoorVisual2018,
  title = {{{InLoc}}: {{Indoor Visual Localization}} with {{Dense Matching}} and {{View Synthesis}}},
  shorttitle = {{{InLoc}}},
  booktitle = {{{CVPR}}},
  author = {Taira, Hajime and Okutomi, Masatoshi and Sattler, Torsten and Cimpoi, Mircea and Pollefeys, Marc and Sivic, Josef and Pajdla, Tomas and Torii, Akihiko},
  year = 2018,
  eprint = {1803.10368},
  primaryclass = {cs},
  publisher = {arXiv},
  url = {http://arxiv.org/abs/1803.10368},
  urldate = {2025-11-06},
  archiveprefix = {arXiv}
}

@article{tanNOPESACNeuralOneplane2023,
  title = {{{NOPE-SAC}}: {{Neural}} One-Plane {{RANSAC}} for Sparse-View Planar {{3D}} Reconstruction},
  shorttitle = {{{NOPE-SAC}}},
  author = {Tan, Bin and Xue, Nan and Wu, Tianfu and Xia, Gui-Song},
  year = 2023,
  journal = {IEEE Trans. Pattern Anal. Mach. Intell.},
  volume = {45},
  number = {12},
  pages = {15233--15248},
  url = {http://dx.doi.org/10.1109/TPAMI.2023.3314745}
}

@inproceedings{tanPlanarSplattingAccuratePlanar2025,
  title = {{{PlanarSplatting}}: {{Accurate Planar Surface Reconstruction}} in 3 {{Minutes}}},
  shorttitle = {{{PlanarSplatting}}},
  booktitle = {{{CVPR}}},
  author = {Tan, Bin and Yu, Rui and Shen, Yujun and Xue, Nan},
  year = 2025,
  eprint = {2412.03451},
  primaryclass = {cs},
  publisher = {arXiv},
  url = {http://arxiv.org/abs/2412.03451},
  urldate = {2024-12-11},
  archiveprefix = {arXiv}
}

@inproceedings{tanPlaneTRStructureGuidedTransformers2021,
  title = {{{PlaneTR}}: {{Structure-Guided Transformers}} for {{3D Plane Recovery}}},
  shorttitle = {{{PlaneTR}}},
  booktitle = {{{ICCV}}},
  author = {Tan, Bin and Xue, Nan and Bai, Song and Wu, Tianfu and Xia, Gui-Song},
  year = 2021,
  pages = {4186--4195},
  url = {http://arxiv.org/abs/2107.13108},
  urldate = {2024-05-01}
}

@inproceedings{teedTangentSpaceBackpropagation2021,
  title = {Tangent Space Backpropagation for {{3D}} Transformation Groups},
  shorttitle = {{{LieTorch}}},
  booktitle = {{{CVPR}}},
  author = {Teed, Zachary and Deng, Jia},
  year = 2021
}

@inproceedings{torii247Place2015,
  title = {24/7 Place Recognition by View Synthesis},
  booktitle = {{{CVPR}}},
  author = {Torii, Akihiko and Arandjelovic, Relja and Sivic, Josef and Okutomi, Masatoshi and Pajdla, Tomas},
  year = 2015,
  pages = {1808--1817}
}

@inproceedings{valentinLearningNavigateEnergy2016,
  title = {Learning to Navigate the Energy Landscape},
  shorttitle = {12scenes},
  booktitle = {{{3DV}}},
  author = {Valentin, Julien and Dai, Angela and Nie{\ss}ner, Matthias and Kohli, Pushmeet and Torr, Philip and Izadi, Shahram and Keskin, Cem},
  year = 2016,
  pages = {323--332},
  publisher = {IEEE}
}

@misc{vandenoordRepresentationLearningContrastive2019,
  title = {Representation Learning with Contrastive Predictive Coding},
  shorttitle = {{{InfoNCE}}},
  author = {{van den Oord}, Aaron and Li, Yazhe and Vinyals, Oriol},
  year = 2019,
  eprint = {1807.03748},
  primaryclass = {cs.LG},
  url = {https://arxiv.org/abs/1807.03748},
  archiveprefix = {arXiv}
}

@inproceedings{vaswaniAttentionAllYou2017,
  title = {Attention Is All You Need},
  shorttitle = {Transformer},
  booktitle = {Advances in Neural Information Processing Systems},
  author = {Vaswani, Ashish and Shazeer, Noam and Parmar, Niki and Uszkoreit, Jakob and Jones, Llion and Gomez, Aidan N and Kaiser, {\L}ukasz and Polosukhin, Illia},
  year = 2017,
  volume = {30},
  publisher = {Curran Associates, Inc.},
  url = {https://proceedings.neurips.cc/paper_files/paper/2017/file/3f5ee243547dee91fbd053c1c4a845aa-Paper.pdf}
}

@article{virtanenSciPy10Fundamental2020,
  title = {{{SciPy}} 1.0: Fundamental Algorithms for Scientific Computing in {{Python}}},
  shorttitle = {{{SciPy}} 1.0},
  author = {Virtanen, Pauli and Gommers, Ralf and Oliphant, Travis E. and Haberland, Matt and Reddy, Tyler and Cournapeau, David and Burovski, Evgeni and Peterson, Pearu and Weckesser, Warren and Bright, Jonathan and {van der Walt}, St{\'e}fan J. and Brett, Matthew and Wilson, Joshua and Millman, K. Jarrod and Mayorov, Nikolay and Nelson, Andrew R. J. and Jones, Eric and Kern, Robert and Larson, Eric and Carey, C. J. and Polat, {\.I}lhan and Feng, Yu and Moore, Eric W. and VanderPlas, Jake and Laxalde, Denis and Perktold, Josef and Cimrman, Robert and Henriksen, Ian and Quintero, E. A. and Harris, Charles R. and Archibald, Anne M. and Ribeiro, Ant{\^o}nio H. and Pedregosa, Fabian and {van Mulbregt}, Paul},
  year = 2020,
  journal = {Nature Methods},
  volume = {17},
  number = {3},
  pages = {261--272},
  publisher = {Nature Publishing Group},
  url = {https://www.nature.com/articles/s41592-019-0686-2},
  urldate = {2025-11-18},
  copyright = {2020 The Author(s)}
}

@inproceedings{wang3DGaussianSplatting2025,
  title = {{{3D}} Gaussian Splatting Based Scene-Independent Relocalization with Unidirectional and Bidirectional Feature Fusion},
  booktitle = {{{NeurIPS}}},
  author = {Wang, Junyi and Wang, Yuze and Duan, Wantong and Wang, Meng and Qi, Yue},
  year = 2025,
  url = {https://openreview.net/forum?id=ewgZItWaHh}
}

@inproceedings{wangDGCGNNLeveragingGeometry2024,
  title = {{{DGC-GNN}}: {{Leveraging Geometry}} and {{Color Cues}} for {{Visual Descriptor-Free 2D-3D Matching}}},
  shorttitle = {{{DGC-GNN}}},
  booktitle = {{{CVPR}}},
  author = {Wang, Shuzhe and Kannala, Juho and Barath, Daniel},
  year = 2024,
  eprint = {2306.12547},
  primaryclass = {cs},
  publisher = {arXiv},
  url = {http://arxiv.org/abs/2306.12547},
  urldate = {2025-08-06},
  archiveprefix = {arXiv}
}

@inproceedings{wangDUSt3RGeometric3D2024,
  title = {{{DUSt3R}}: {{Geometric 3D Vision Made Easy}}},
  shorttitle = {{{DUSt3R}}},
  booktitle = {{{CVPR}}},
  author = {Wang, Shuzhe and Leroy, Vincent and Cabon, Yohann and Chidlovskii, Boris and Revaud, Jerome},
  year = 2024,
  eprint = {2312.14132},
  primaryclass = {cs},
  pages = {20697--20709},
  publisher = {arXiv},
  url = {http://arxiv.org/abs/2312.14132},
  urldate = {2023-12-24},
  archiveprefix = {arXiv}
}

@inproceedings{wangFreeRegImagetoPointCloud2024,
  title = {{{FreeReg}}: {{Image-to-Point Cloud Registration Leveraging Pretrained Diffusion Models}} and {{Monocular Depth Estimators}}},
  shorttitle = {{{FreeReg}}},
  booktitle = {{{ICLR}}},
  author = {Wang, Haiping and Liu, Yuan and Wang, Bing and Sun, Yujing and Dong, Zhen and Wang, Wenping and Yang, Bisheng},
  year = 2024,
  eprint = {2310.03420},
  primaryclass = {cs},
  url = {http://arxiv.org/abs/2310.03420},
  urldate = {2025-03-01},
  archiveprefix = {arXiv}
}

@inproceedings{wangMoGe2AccurateMonocular2025,
  title = {{{MoGe-2}}: {{Accurate Monocular Geometry}} with {{Metric Scale}} and {{Sharp Details}}},
  shorttitle = {{{MoGe-2}}},
  booktitle = {{{NeurIPS}}},
  author = {Wang, Ruicheng and Xu, Sicheng and Dong, Yue and Deng, Yu and Xiang, Jianfeng and Lv, Zelong and Sun, Guangzhong and Tong, Xin and Yang, Jiaolong},
  year = 2025,
  eprint = {2507.02546},
  primaryclass = {cs},
  publisher = {arXiv},
  url = {http://arxiv.org/abs/2507.02546},
  urldate = {2025-07-07},
  archiveprefix = {arXiv}
}

@inproceedings{wangMoGeUnlockingAccurate2025,
  title = {{{MoGe}}: {{Unlocking Accurate Monocular Geometry Estimation}} for {{Open-Domain Images}} with {{Optimal Training Supervision}}},
  shorttitle = {{{MoGe}}},
  booktitle = {{{CVPR}}},
  author = {Wang, Ruicheng and Xu, Sicheng and Dai, Cassie and Xiang, Jianfeng and Deng, Yu and Tong, Xin and Yang, Jiaolong},
  year = 2025,
  eprint = {2410.19115},
  primaryclass = {cs},
  publisher = {arXiv},
  url = {http://arxiv.org/abs/2410.19115},
  urldate = {2024-12-11},
  archiveprefix = {arXiv}
}

@inproceedings{wangNeRFIBVSVisualServo2023,
  title = {{{NeRF-IBVS}}: {{Visual}} Servo Based on {{NeRF}} for Visual Localization and Navigation},
  shorttitle = {{{NeRF-IBVS}}},
  booktitle = {{{NeurIPS}}},
  author = {Wang, Yuanze and Yan, Yichao and Shi, Dianxi and Zhu, Wenhan and Xia, Jianqiang and Jeff, Tan and Jin, Songchang and Gao, Ke and Li, Xiaobo and Yang, Xiaokang},
  year = 2023,
  publisher = {OpenReview. net},
  url = {https://openreview.net/forum?id=9pLaDXX8m3}
}

@inproceedings{watsonAirPlanesAccuratePlane2024,
  title = {{{AirPlanes}}: {{Accurate}} Plane Estimation via {{3D-consistent}} Embeddings},
  shorttitle = {{{AirPlanes}}},
  booktitle = {{{CVPR}}},
  author = {Watson, Jamie and Aleotti, Filippo and Sayed, Mohamed and Qureshi, Zawar and Mac Aodha, Oisin and Brostow, Gabriel and Firman, Michael and Vicente, Sara},
  year = 2024,
  pages = {5270--5280}
}

@article{wietrzykowskiPlaneLocProbabilisticGlobal2019,
  title = {{{PlaneLoc}}: {{Probabilistic}} Global Localization in 3-{{D}} Using Local Planar Features},
  shorttitle = {{{PlaneLoc}}},
  author = {Wietrzykowski, Jan and Skrzypczy{\'n}ski, Piotr},
  year = 2019,
  journal = {Robotics and Autonomous Systems},
  volume = {113},
  pages = {160--173},
  url = {https://www.sciencedirect.com/science/article/pii/S0921889018303701},
  urldate = {2025-03-02}
}

@misc{wuDetectron22019,
  title = {Detectron2},
  shorttitle = {Detectron2},
  author = {Wu, Yuxin and Kirillov, Alexander and Massa, Francisco and Lo, Wan-Yen and Girshick, Ross},
  year = 2019,
  url = {https://github.com/facebookresearch/detectron2}
}

@inproceedings{xiePlanarReconRealtime3D2022,
  title = {{{PlanarRecon}}: {{Realtime 3D Plane Detection}} and {{Reconstruction}} from {{Posed Monocular Videos}}},
  shorttitle = {{{PlanarRecon}}},
  booktitle = {{{CVPR}}},
  author = {Xie, Yiming and Gadelha, Matheus and Yang, Fengting and Zhou, Xiaowei and Jiang, Huaizu},
  year = 2022,
  pages = {6209--6218},
  url = {https://ieeexplore.ieee.org/document/9880246/},
  urldate = {2024-02-24}
}

@inproceedings{yangRecovering3DPlanes2018,
  title = {Recovering {{3D}} Planes from a Single Image via Convolutional Neural Networks},
  shorttitle = {{{PlaneRecover}}},
  booktitle = {{{ECCV}}},
  author = {Yang, Fengting and Zhou, Zihan},
  year = 2018,
  pages = {87--103}
}

@inproceedings{ye2025neuralplane,
  title = {{{NeuralPlane}}: {{Structured 3D}} Reconstruction in Planar Primitives with Neural Fields},
  shorttitle = {{{NeuralPlane}}},
  booktitle = {{{ICLR}}},
  author = {Ye, Hanqiao and Liu, Yuzhou and Liu, Yangdong and Shen, Shuhan},
  year = 2025,
  url = {https://openreview.net/forum?id=5UKrnKuspb}
}

@inproceedings{yen-chenINeRFInvertingNeural2021,
  title = {{{iNeRF}}: {{Inverting Neural Radiance Fields}} for {{Pose Estimation}}},
  shorttitle = {{{INeRF}}},
  booktitle = {2021 {{IEEE}}/{{RSJ International Conference}} on {{Intelligent Robots}} and {{Systems}}},
  author = {{Yen-Chen}, Lin and Florence, Pete and Barron, Jonathan T. and Rodriguez, Alberto and Isola, Phillip and Lin, Tsung-Yi},
  year = 2021,
  eprint = {2012.05877},
  primaryclass = {cs},
  pages = {1323--1330},
  publisher = {IEEE},
  url = {http://arxiv.org/abs/2012.05877},
  urldate = {2023-06-06},
  archiveprefix = {arXiv}
}

@inproceedings{yuFindingGoodConfigurations2022,
  title = {Finding Good Configurations of Planar Primitives in Unorganized Point Clouds},
  booktitle = {{{CVPR}}},
  author = {Yu, Mulin and Lafarge, Florent},
  year = 2022,
  pages = {6357--6366},
  publisher = {IEEE},
  url = {https://doi.org/10.1109/CVPR52688.2022.00626}
}

@inproceedings{yuSingleimagePiecewisePlanar2019,
  title = {Single-Image Piece-Wise Planar {{3D}} Reconstruction via Associative Embedding},
  shorttitle = {{{PlaneAE}}},
  booktitle = {{{CVPR}}},
  author = {Yu, Zehao and Zheng, Jia and Lian, Dongze and Zhou, Zihan and Gao, Shenghua},
  year = 2019,
  pages = {1029--1037},
  publisher = {Computer Vision Foundation / IEEE},
  bibsource = {dblp computer science bibliography, https://dblp.org}
}

@article{zhaiSplatLoc3DGaussian2024a,
  title = {{{SplatLoc}}: {{3D}} Gaussian Splatting-Based Visual Localization for Augmented Reality},
  shorttitle = {{{SplatLoc}}},
  author = {Zhai, Hongjia and Zhang, Xiyu and Zhao, Boming and Li, Hai and He, Yijia and Cui, Zhaopeng and Bao, Hujun and Zhang, Guofeng},
  year = 2024,
  journal = {IEEE Trans. Vis. Comput. Graph.},
  volume = {31},
  number = {5},
  pages = {3591--3601},
  url = {https://api.semanticscholar.org/CorpusID:272827161}
}

@inproceedings{zhangA2GNNAngleAnnularGNN2025,
  title = {A2-{{GNN}}: {{Angle-Annular GNN}} for {{Visual Descriptor-free Camera Relocalization}}},
  shorttitle = {A2-{{GNN}}},
  booktitle = {{{3DV}}},
  author = {Zhang, Yejun and Wang, Shuzhe and Kannala, Juho},
  year = 2025,
  eprint = {2502.20036},
  primaryclass = {cs},
  publisher = {arXiv},
  url = {http://arxiv.org/abs/2502.20036},
  urldate = {2025-05-24},
  archiveprefix = {arXiv}
}

@article{zhangCornerVINSAccurateLocalization2025,
  title = {{{CornerVINS}}: {{Accurate}} Localization and Layout Mapping for Structural Environments Leveraging Hierarchical Geometric Representations},
  shorttitle = {{{CornerVINS}}},
  author = {Zhang, Yidi and Tang, Fulin and Wu, Yihong},
  year = 2025,
  journal = {IEEE Trans. Robot.},
  volume = {41},
  pages = {3500--3517}
}

@inproceedings{zhangMultiviewSceneGraph2024,
  title = {Multiview {{Scene Graph}}},
  shorttitle = {{{MSG}}},
  booktitle = {{{NeurIPS}}},
  author = {Zhang, Juexiao and Zhu, Gao and Li, Sihang and Liu, Xinhao and Song, Haorui and Tang, Xinran and Feng, Chen},
  year = 2024,
  eprint = {2410.11187},
  primaryclass = {cs},
  publisher = {arXiv},
  url = {http://arxiv.org/abs/2410.11187},
  urldate = {2025-02-18},
  archiveprefix = {arXiv}
}

@inproceedings{zhaoPNeRFLocVisualLocalization2024,
  title = {{{PNeRFLoc}}: {{Visual Localization}} with {{Point-based Neural Radiance Fields}}},
  shorttitle = {{{PNeRFLoc}}},
  booktitle = {{{AAAI}}},
  author = {Zhao, Boming and Yang, Luwei and Mao, Mao and Bao, Hujun and Cui, Zhaopeng},
  year = 2024,
  pages = {7450--7459},
  publisher = {AAAI},
  url = {https://zju3dv.github.io/PNeRFLoc/},
  urldate = {2023-12-20}
}

@inproceedings{zhengStructured3DLargePhotorealistic2020,
  title = {{{Structured3D}}: {{A Large Photo-realistic Dataset}} for {{Structured 3D Modeling}}},
  shorttitle = {{{Structured3D}}},
  booktitle = {{{ECCV}}},
  author = {Zheng, Jia and Zhang, Junfei and Li, Jing and Tang, Rui and Gao, Shenghua and Zhou, Zihan},
  year = 2020,
  eprint = {1908.00222},
  primaryclass = {cs},
  publisher = {arXiv},
  url = {http://arxiv.org/abs/1908.00222},
  urldate = {2025-07-21},
  archiveprefix = {arXiv}
}

@inproceedings{zhouGeometryEnoughMatching2022,
  title = {Is Geometry Enough {{For Matching In Visual}} Localization?},
  shorttitle = {{{GoMatch}}},
  booktitle = {{{ECCV}}},
  author = {Zhou, Qun Jie and Agostinho, S{\'e}rgio and O{\v s}ep, Aljo{\v s}a and {Leal-Taix{\'e}}, Laura},
  year = 2022,
  pages = {407--425},
  publisher = {Springer},
  url = {https://doi.org/10.1007/978-3-031-20080-9_24}
}

@inproceedings{zhouNeRFectMatchExploring2024,
  title = {The {{NeRFect Match}}: {{Exploring NeRF Features}} for {{Visual Localization}}},
  shorttitle = {The {{NeRFect Match}}},
  booktitle = {{{ECCV}}},
  author = {Zhou, Qunjie and Maximov, Maxim and Litany, Or and {Leal-Taix{\'e}}, Laura},
  year = 2024,
  eprint = {2403.09577},
  primaryclass = {cs},
  publisher = {arXiv},
  url = {http://arxiv.org/abs/2403.09577},
  urldate = {2025-11-06},
  archiveprefix = {arXiv}
}

@inproceedings{zhuLoDlocV2Aerial2025,
  title = {{{LoD-loc}} v2: {{Aerial}} Visual Localization over Low Level-of-Detail City Models Using Explicit Silhouette Alignment},
  shorttitle = {{{LoD-loc}} V2},
  booktitle = {{{ICCV}}},
  author = {Zhu, Juelin and Peng, Shuaibang and Wang, Long and Tan, Hanlin and Liu, Yu and Zhang, Maojun and Yan, Shen},
  year = 2025
}
}
\end{document}